\documentclass{article}
\usepackage[preprint]{neurips_2026}
\usepackage[utf8]{inputenc}
\usepackage[T1]{fontenc}
\usepackage{hyperref}
\usepackage{nicefrac}
\usepackage{amsmath,amssymb,amsthm,mathtools,bm,booktabs,enumitem,graphicx,microtype}
\usepackage{url}
\usepackage{xcolor}
\usepackage{algorithm}
\usepackage{algpseudocode}

\newtheorem{definition}{Definition}

\newtheorem{theorem}{Theorem}
\newtheorem{proposition}{Proposition}

\newtheorem{remark}{Remark}
\newtheorem{counterexample}{Counterexample}

\newcommand{\Law}{\operatorname{Law}}

\newcommand{\Res}{\mathfrak{R}}

\title{Synergetic Autonomous Intelligence:\\[-0.2em]
Three Axioms for Recursive Semi-Closed Learning}
\author{Dongxu Li\\Xi'an Jiaotong University\\\textbf{Plokerm AI Lab}\\\texttt{skey@stu.xjtu.edu.cn}}

\begin{document}
\maketitle

\begin{abstract}
We introduce and formally specify \emph{Synergetic Autonomous Intelligence} (SAI), the founding theory and design program for learning systems that form, preserve, revise, and compose relations while retaining the ability to continue learning. The theory takes three object-level axioms as its sole first-principles stance: consequential distinctions must not become orphaned (U), identity is determined by future-faithful continuation under a declared contract (I), and unresolved defects are repaired by the least lawful local completion available at the lowest admitted grade (L). A recursive semi-closed system is the common mathematical object for a model, an agent, a tool, a team, or an environment boundary. We formalize its Kernel--Ring representation, the Compile--Open--Participate cycle, and a generalized residual as a typed defect between actual and claimed continuation. Under an explicit finite contract schema, we also prove relative consistency/non-vacuity of U/I/L and conditional semantic non-redundancy via finite countermodels with a reified identity field. Under explicit finite or affine assumptions, we characterize a contract-relative semantic quotient, construct a finite exact joint stochastic refinement, derive a finite-affine constrained residual contraction, prove a finite-horizon viability theorem for continued service, and give a finite executed-intervention identification sector with an observational-confounding boundary. We also state a conditional classification separating within-fibre transformation from quotient/owner revision. These constructions establish SAI's common contract-level vocabulary for representation, relation, memory, action, communication, and growth. The finite sectors are the first formal realizations of the founding program, with their contracts providing the mathematical interface for later generalization to broader AGI/ASI settings. The appendix also composes the context-to-action, action-to-effect, effect-to-feedback, feedback-to-memory, and memory-to-context maps into a finite action-to-effect/feedback \emph{Loop} re-entry certificate with causal-status and re-authorization gates. It further closes finite partial failure and duplicate-effect re-entry with an idempotency namespace, unique recovery branch, and lookup-only replay. The finite concurrent extension adds a declared delivery order or commutation witness for duplicate and conflict races. Finally, a finite KREIT--synergetic-cybernetic bridge makes the information, repair, and control projections commute on a shared typed reducer, while an effective representation theorem relates recursively encoded semi-closed systems to nondeterministic machines without claiming a stronger computability class. A finite causal-probability sector then treats interventions, uncertainty, evidence, and unresolved probability mass as one typed continuation object. It separates mechanism posterior uncertainty, fixed-mechanism continuation variation, and explicit OPEN status; relation entityization is not causal certification, and unsupported weighting returns an identification set rather than a fabricated point posterior. This is a contract-relative reconstruction of causal and probabilistic semantics, not a replacement for Kolmogorov probability. A further candidate substrate is the threshold--radix relation object: a typed filtration whose digits record the first scale at which a future-relevant distinction must be independently carried, together with carry, expansion, transport, and update maps. The same axioms also induce a threshold-radix bit-vector representation for finite ordered typed states, in which protected distinctions, future-law classes, resources, and repair grades become addressable threshold coordinates; we give its finite lattice and recursive-nesting semantics as a candidate mathematical foundation rather than an unconditional replacement for existing mathematics.
We position this manuscript as a formal opening of a new information-theoretic and synergetic-cybernetic program. Its information layer, which we call \emph{Kernel--Ring Equilibrium Information Theory} (KREIT), treats information as a typed, resource-bounded distinction that must preserve both local response and the system's declared participation in background updates across a continuation. Its control layer is the closed loop of authorization, action, effect, feedback, and update; GRC-Close is the corresponding generative and repair schema. The same semantics gives a cloud--edge--end architecture in which multimodal perception, memory, causal inquiry, communication, action, explanation, and collective coordination are capability instances of one recursively nested contract. Communication, economic coordination, and energy coordination are domain instantiations of this interface with their own units, conservation rules, and ledgers; the paper establishes the shared formal interface, with domain-specific units, conservation rules, and ledgers.
\end{abstract}

\section{Introduction}
Modern learning systems optimize a sequence of objectives, but their state is usually described either as parameters or as a current prediction. Both views are too weak for a system that must act, accept feedback, change its own representation, and remain usable after its sources or collaborators change. A prediction can be correct now and still be insufficient for the next update; a parameter change can preserve one output while breaking a protected action; and a larger model can increase capacity without creating a reusable responsibility.

This paper is the founding statement of Synergetic Autonomous Intelligence. Its purpose is to establish the SAI object, its three axioms, its recursive mathematics, its executable contract semantics, and its first algorithmic method. The paper therefore presents a positive theory and design program: the finite propositions are initial realizations of one common closure language, and GRC-Close is the reference method by which a SAI system preserves, repairs, and expands its future capability.

At the architectural level, cloud, edge, and end devices are one recursively nested semi-closed system rather than three disconnected tiers. Each layer exposes the same typed contract for state, relation, permission, resource, action, effect, feedback, and replay; crossing a layer is a transport problem with an explicit error and lifecycle-cost ledger. The resulting interface covers the SAI capability family; communication, economic coordination, and energy coordination plug into it as domain models with explicit units, conservation rules, and ledgers. This defines the scope of the new information theory and synergetic cybernetics developed here.
The division of labor is explicit. KREIT specifies which distinctions can be safely compressed into a Kernel relative to a Ring and a background, including the conditions for returning to the Ring and for updating the background. GRC specifies how a generalized residual triggers the least lawful repair or growth step. Synergetic cybernetics specifies how those steps are authorized, communicated, executed, audited, and handed off across scales. These three views form one SAI contract, with each domain retaining its own units and lawful composition rules. Appendix~\ref{app:kreitcyberbridge} proves a finite semantic bridge between the three views, and Appendix~\ref{app:ndtmrepresentation} gives the corresponding effective nondeterministic-machine representation with an explicit trace-equivalence boundary.

SAI starts from a different question: \emph{which consequences can a changing relation continue to carry, and what new relation is required when its current boundary cannot?} The answer is contract-relative. A contract declares legal observations, interventions, feedback, updates, background changes, resources, and receipts. The contract determines which distinctions are meaningful. A state is adequate only when all declared continuations remain well typed and executable after the state is reopened.

The proposed synthesis draws on stability--plasticity theory \citep{grossberg1980,mcclelland1995,kirkpatrick2017}, predictive and active inference \citep{rao1999,friston2010}, cybernetics and synergetics \citep{ashby1956,haken1977}, causal representation \citep{scholkopf2021,peters2017}, memory and sparse routing \citep{graves2014,shazeer2017,fedus2021}, and multi-agent coordination \citep{olfatisaber2007,foerster2016,lowe2017}. The synthesis places these results under one common contract-level semantics in which they become different realizations of the same closure problem. In particular, it strengthens causal analysis by requiring an authorized executed intervention, mechanism identity, shared-background counterfactual scope, and a replayable effect-to-feedback chain. It treats probability as a typed family of continuation laws, separating mechanism uncertainty, fixed-mechanism continuation variation, and explicit unresolved status; without a declared weighting rule it returns an identification set rather than an unqualified scalar confidence.

Our contributions are:
\begin{enumerate}[leftmargin=1.3em,itemsep=0.2em]
    \item a precise statement of the three axioms U/I/L and their relation to recursive semi-closed systems, together with a finite relative-consistency/non-vacuity model and conditional semantic non-redundancy countermodels;
    \item a finite KREIT--synergetic-cybernetic bridge theorem showing when information encoding, residual repair, and authorization--action--effect--feedback control are commuting projections of one SAI reducer, plus a relative recursive representation theorem for typed semi-closed systems and nondeterministic machines;
    \item a finite causal-probability contract in which observational, interventional, evidence, resource, and unresolved branches are typed components of one continuation law, with a finite intervention-identification theorem, a mechanism/effect identification-set distinction, explicit confounding/OPEN boundaries, and separate fixed-horizon versus increasing-intervention limits;
    \item a finite threshold--radix relation algebra in which U/I/L generate typed refinement digits, cross-level carry, and projection-compatible execution, while matrices remain conditional backends rather than universal primitives;
    \item a future-sufficient Kernel--Ring semantics that treats representation, relation, memory, ownership, and background as jointly typed state;
    \item conditional theorems for exact quotienting, finite-affine joint closure with residual contraction, finite typed nonlinear residual composition, finite autonomous-goal lifecycle safety/replay, finite partially observed AND/OR planning with conservative value-of-information gates, finite interval Bellman closure with vector-resource prefixes, finite concurrent batch closure with shared-resource interference, finite sound belief abstraction and refinement, finite coalition closure with externalities and responsibility transfer, finite contract-compatible participation and explanation transfer, finite repeated replacement with compositional revoke propagation, finite grounded owner-handoff composition, lawful-completion selection under initial-object or replayable tie-break witnesses, and finite continued service;
    \item a single GRC-Close reference algorithmic schema with explicit failure, resource, and evidence semantics; and
    \item a cloud--edge--end unification contract in which perception, memory, causal inquiry, action, communication, explanation, and collective coordination share one closure interface, with communication, economic, and energy domains represented by typed instantiations rather than conflated laws, together with finite action-to-effect/feedback \emph{Loop} re-entry closure with causal-status and re-authorization gates, finite partial-failure and duplicate-effect idempotent re-entry, lookup-only replay, and a finite concurrent order-conflict extension with declared linearization or commutation witnesses; and
    \item a constructive boundary calculus of counterexamples and claim grades that turns each new SAI capability into a theorem, design contract, conjecture, or empirical program.
\end{enumerate}

\section{First principles and the recursive object}
\subsection{Contracts precede quotients}
Let a contract be
\[
 \Xi=(\mathcal Q,\mathcal A,\mathcal Y,\mathcal U,\mathcal B,\mathcal C,\mathcal M,\mathcal T),
\]
where $\mathcal Q$ are queries, $\mathcal A$ actions, $\mathcal Y$ feedback, $\mathcal U$ update events, $\mathcal B$ allowed background states, $\mathcal C$ resource limits, $\mathcal M$ receipts and ownership rules, and $\mathcal T$ admissible finite or stochastic continuation policies. Let $\mathcal O$ denote the authorized owner/source context projected from $\mathcal M$. A continuation may choose later actions as a function of earlier observations. Thus an output-only contract is a special, weaker contract.

For the results below, each space carries a declared measurable or finite structure. Enabled actions are partial: an unavailable action returns a typed refusal rather than an unrecorded exception. A legal one-step kernel returns the joint object
\[
 (Y,K^+,B^+,M^+,C,\omega)\sim\mathsf{Step}_\Xi(x,c),
\]
including feedback, retained successor, background successor, responsibility receipt, charged cost, and typed disposition \(\omega\). Refusal, timeout, and source-denied failure are values in this same response space. These assumptions are part of the contract; they are not inferred from a current model output.

\begin{definition}[Recursive semi-closed system]
A recursive semi-closed system is
\[
 S=(B,\Xi,K,R,E,\pi,\mathsf{Open},\mathsf{Compile},\mathsf{Part},\mathsf{Upd}),
\]
where $B$ is background, $K$ is retained Kernel state, $R$ is a family of lawful active Ring realizations, $E$ is a responsibility/owner address, $\pi:R\to K$ is the Compile map, $\mathsf{Open}:K\times B\times\mathcal O\rightharpoonup R$ is a lawful expansion, $\mathsf{Part}$ executes actions and returns feedback, and $\mathsf{Upd}$ changes the whole state. The system is semi-closed because its interface can be stable while its internal realization and external background continue to change. It is recursive because every qualified relation can itself be a Kernel inside a larger system.

Here ``Ring'' names a lawful realization family and is not an algebraic ring: each opened realization must expose the declared action, feedback, update, owner, and receipt interface, and either return a typed successor or a typed refusal/timeout/source-denied result.

We reserve ``Ring'' for this structural realization role and use ``loop'' for
the temporal feedback dynamics. Thus the Kernel--Ring structure is traversed
by an Address--Open--Part--Upd execution loop; the two words are not
interchangeable.

More precisely, $\mathsf{Open}$ and $\mathsf{Part}$ are partial measurable programs, $\mathsf{Upd}$ consumes a typed event $e=(q,a,y,B^+,M^+,C,\omega)$ and returns the next complete state, and $\mathsf{Compile}$ is an executable map from a qualified Ring realization to a retained Kernel. The projection $\pi$ is only a structural observation; it is not a second update rule. A qualified retained state must have a source-denied reopening witness and a declared owner for every protected receipt.
\end{definition}

The basic execution cycle is
\[
 K_t \xrightarrow{\mathsf{Open}(K_t,B_t,o_t)} r_t \xrightarrow{\mathsf{Part}(q_t,a_t,o_t)}
 (y_t,B^+_t,M^+_t,C_t,\omega_t) \xrightarrow{\mathsf{Upd}(e_t)} S_{t+1}
 \xrightarrow{\mathsf{Compile}} K_{t+1}.
\]
Here $e_t=(q_t,a_t,y_t,B^+_t,M^+_t,C_t,\omega_t)$ is the complete typed event, including refusal or timeout when applicable. Open is an actual program, not a mathematical choice of a convenient representative. Compile may be many-to-one, but a qualified state must have at least one admissible reopening procedure that uses only the retained package, declared background, and authorized interfaces. Costs for admission, search, transport, validation, and maintenance belong to the contract.

\subsection{The three axioms}
The historical names are retained; the operational forms make them testable.

\paragraph{U: universe closure / no orphan future distinction.}
Any \emph{verified protected distinction} in a sealed contract that can change a legal future must either (i) be represented by a state or owner that can receive its consequences, (ii) be transferred to an authorized receiver, or (iii) be explicitly recorded as unresolved with its evidence and cost. Silent loss is a violation of U. The adjective ``verified'' limits U to distinctions that the contract can observe, test, or conservatively protect; U is not an oracle for arbitrary hidden facts.

\paragraph{I: inertia / future-faithful semi-closed identity.}
For a fixed contract, two realizations have the same identity exactly when every admissible adaptive continuation gives the same joint law of typed responses, enabled actions, successor observables, successor background, and declared resource/receipt behavior. Microscopic equality is neither required nor sufficient. The identity criterion is therefore stated in terms of observable successor laws, not a presumed future Kernel, avoiding a representation/identity cycle.

\paragraph{L: local-priority entropy maximization / least lawful completion.}
After excluding a valid incumbent repair, transport, missing evidence, and insufficient search by declared finite gates, close the current defect with the lowest admitted structural grade and a minimal lawful change under the declared order. Within that boundary, retain unforced variation subject to the same contract and resource ledger. The entropy phrase is a first-principles orientation; the least-completion statement is the executable bridge. They are not asserted equivalent for all physical systems.

The axioms constrain one generation event; they are not three software modules, three weighted losses, or a promise that a completion exists. The existence of a feasible action, an owner, and a finite resource budget are separate assumptions. Appendix~\ref{app:axiomconsistency} gives a finite non-vacuous model satisfying all three, while Appendix~\ref{app:axiomindependence} gives conditional semantic non-redundancy countermodels under an explicit identity field. These are relative finite-contract results, not absolute proof-theoretic consistency, unrestricted computability, or unconditional independence claims.

A candidate mathematical substrate follows naturally from the same contract. A finite threshold--radix relation object records a filtration of necessary distinctions, a variable number of children at each parent, the carry/expansion rule that makes a new distinction independently callable, and the transport/update maps that preserve its consequences. A digit is therefore a structural address or relation record, not automatically a numerical bit; only a declared linear or modular sector permits vector-space or radix arithmetic. Appendix~\ref{app:thresholdradix} proves the finite address, refinement, and projection properties under explicit coherence and calibration assumptions, and records the boundary where the construction is only a partial typed relation program.

\section{Future-sufficient identity and relation entityization}
\begin{definition}[Future equivalence]
For histories $x,x'$ in a state space $X$, define
\[
x\sim_\Xi x' \Longleftrightarrow
\Law_\Xi(\tau\mid x)=\Law_\Xi(\tau\mid x')
\quad\text{for every legal adaptive continuation }\tau,
\]
where the comparison uses one common typed continuation domain $\mathcal T$ and $\tau$ includes feedback, enabledness/refusal, successor observables, background effects, resources, and receipts. Equality is equality of the regular conditional laws on legal positive-probability histories; zero-probability histories are handled by the contract's declared version or excluded from the reachable domain.
\end{definition}

\begin{theorem}[Canonical future quotient]\label{thm:canonical}
For any fixed contract, $\sim_\Xi$ is an equivalence relation. Let $q_\Xi:X\to K_\Xi=X/{\sim_\Xi}$. If an encoding $C:X\to Z$ is exactly sufficient for all continuations, then there exists a unique map $f:Z_C\to K_\Xi$ on the reachable image such that $q_\Xi=f\circ C$. If $C$ is itself future-faithful, $f$ is a bijection up to relabeling.
\end{theorem}

\begin{proof}[Proof sketch; full proof in Appendix~\ref{app:canonical}]
Reflexivity, symmetry, and transitivity follow from equality of all continuation laws. If $C(x)=C(x')$, exact sufficiency forces equal continuation laws, hence $q_\Xi(x)=q_\Xi(x')$; define $f(C(x))=q_\Xi(x)$. The definition is independent of the representative, and uniqueness holds on the reachable image.
\end{proof}

\begin{definition}[Relation entityization]
A relation becomes an entity at a declared scope when a future-sufficient Kernel, an executable Ring/Open procedure, an independently addressable responsibility interface $E$ with authority, feedback, and unresolved-duty semantics, and a payable maintenance/retirement budget coexist. A new state distinction does not by itself imply a new owner.
\end{definition}

This separates four questions that are often conflated: what must be remembered, how it is reopened, who receives the consequence, and whether the resulting capability is worth its lifecycle cost. The same relation can be a Ring at one scale and a Kernel at another.

\section{Generalized residual closure}
Let $C_t$ be a retained code and $U_e$ an actual update event. Let $\bar U_e$ be the update claimed by the code. The generalized residual is
\[
 \Res_e^\Xi = \operatorname{Def}_\Xi(C_{t+1}U_e,\bar U_eC_t),
\]
where $\operatorname{Def}_\Xi$ is defined only on a typed comparison domain. If the necessary response, successor, owner, or resource comparison is unavailable, the result is $\mathsf{UNDEFINED/OPEN}$ rather than an invented zero residual. A defined residual is a bundle $(\Xi,\mathrm{type},\mathrm{witness},\mathrm{obligations},\mathrm{rights},\mathrm{cost})$ with a typed witness of incompatible response, successor, ownership, resource, or provenance behavior. A residual identifies a defect, not its remedy: the same residual can be repaired by a within-Kernel transformation, a representation refinement, a new owner, a transport change, or a refusal.

For nonlinear composition, Proposition~\ref{app:nonlinearresidual} gives a finite typed bound under normed codomains, a common declared image/coupling, and Lipschitz endpoint conditions; without those conditions the composed residual remains \(\mathsf{OPEN}\).

\subsection{Joint closure and the two learning modes}
For a local affine sector, let $d$ be a within-representation transformation and $u$ a representation/owner change. Protected old behavior and new residual are
\[
 A d+E u=0,\qquad B d+D u-r \text{ is minimized under the contract tolerance}.
\]
For $H,G\succ0$ and $\eta>0$, GRC-Close solves
\begin{equation}
 \min_{d,u}\;\frac12\|Bd+Du-r\|^2+\frac{1}{2\eta}(d^THd+u^TGu)
 \quad\text{s.t. }Ad+Eu=0. \label{eq:joint}
\end{equation}
Define $\widetilde A=[A\ E]$, $\widetilde B=[B\ D]$, and $\widetilde H=\operatorname{diag}(H,G)$. Then
\[
 P=\widetilde H^{-1}-\widetilde H^{-1}\widetilde A^T
 (\widetilde A\widetilde H^{-1}\widetilde A^T)^\dagger
 \widetilde A\widetilde H^{-1},\qquad
 K_{\rm plast}=\widetilde B P\widetilde B^T.
\]

\begin{theorem}[Conditional finite-affine joint closure]\label{thm:affine}
Assume finite-dimensional real matrices, $H,G$ symmetric positive definite, and homogeneous constraints so the zero update is feasible. Under these assumptions, the feasible minimizer of Eq.~\ref{eq:joint} is unique and is
\[
 v^*=\eta P\widetilde B^T(I+\eta K_{\rm plast})^{-1}r,
 \qquad r^+=(I+\eta K_{\rm plast})^{-1}r,
\]
where $v=(d,u)$. Old protected responses are unchanged, and every positive eigen-direction of $K_{\rm plast}$ contracts the new residual. Directions in the nullspace remain unresolved; carrying, refining, or refusing them is a subsequent design choice, not a theorem conclusion.
\end{theorem}

\begin{proof}[Proof sketch]
The feasible set contains zero and the proximal term is strongly convex. KKT elimination gives the $\widetilde H$-orthogonal null-space operator $P$ and the displayed Schur complement. Positive semidefiniteness of $K_{\rm plast}$ implies eigenvalue-wise contraction of $(I+\eta K_{\rm plast})^{-1}$.
\end{proof}

The theorem is finite and conditional. It does not establish a global optimum for a nonconvex neural network, a changing Jacobian, or an unknown future event family. It also exposes why freezing all shared coordinates is not the definition of stability: internal compensation can satisfy old duties while writing a new coordinate.

\subsection{A finite exact stochastic closure}
For a finite state space, finite action/feedback alphabet, and exact transition kernel $W_a(y,x'\mid x)$, initialize a partition by protected readouts, enabledness, owner/receipt state, background, and resource ledger. Any background, cost, or receipt component that can affect a legal future is included in $x'$ or in the emitted joint output. Refine two cells whenever some action gives different joint probabilities of $(y,\text{successor cell})$. The process terminates after at most $|X|-|X/E_0|$ strict refinements and yields the coarsest partition satisfying
\[
 \overline W_a(y,k'\mid C(x))=
 \sum_{u:C(u)=k'}W_a(y,u\mid x).
\]
This is the exact joint response--successor condition. Matching response and successor marginals separately is insufficient.

\begin{counterexample}[Marginals do not define continuation]
Two systems can emit the same fair feedback and have the same fair successor marginal, while one sets $(Y,K')=(0,0),(1,1)$ and the other sets $(Y,K')=(0,1),(1,0)$. Any contract that later compares feedback with successor behavior distinguishes them after one step, so separate marginal matching is not a valid quotient.
\end{counterexample}

\section{Continual service, growth, and the SAI algorithm}
\subsection{Viability is the right infinite-horizon question}
Let a finite exact arena have observable states $s$, requests $q$, actions $a$, nonnegative integer costs $c$, a completion set $G_q$, a ready set $R$, and partial transitions $T(s,a,q)$ with a finite action horizon. Requests are serial and noninterrupting, and all successors, enabledness, and costs are visible to the arena; no hidden mechanism may be used by the witnessing policy. For an allowance $H$ define
\[
 \Phi_H(V)=\{s\in R:\forall q\;\exists\text{ a policy of cost }\le H
 \text{ reaching }G_q\text{ and returning to }V\}.
\]

\begin{theorem}[Finite continued-service fixed point]\label{thm:viability}
On a finite arena, the descending sequence $V_0=R$, $V_{n+1}=\Phi_H(V_n)$ stabilizes at the greatest fixed point $V^*=\nu\Phi_H$. A controller can complete every finite serial request sequence within allowance $H$ and return ready exactly from states in $V^*$.
\end{theorem}

\begin{proof}[Proof sketch; full proof in Appendix~\ref{app:viabilityproof}]
Monotonicity of $\Phi_H$ gives a descending chain in a finite lattice, hence finite stabilization. If $s\in V^*$, select the witnessing policy for the current request and use the fixed-point return state inductively. Conversely, any controller that survives all serial request sequences must return after every request to a set $V$ with $V\subseteq\Phi_H(V)$, hence $V\subseteq V^*$ by greatest-fixed-point induction.
\end{proof}

Growth is defined relative to a frozen contract, a predeclared nonempty request set or distribution, and full lifecycle cost. A successor grows when its reliable future domain expands under an executable continuation simulation, or the same domain becomes cheaper in expected or worst-case cost on that predeclared set, while protected responsibilities and the ability to continue learning remain admissible. Appendix~\ref{app:growthcert} makes the prefix budget, recovery reserve, translator overhead, and distributional scope explicit. Parameter count, current accuracy, a larger context window, or an external resource grant is not growth by itself.

\begin{algorithm}[t]
\caption{GRC-Close: one SAI update}
\label{alg:grc}
\begin{algorithmic}[1]
\Require immutable parent $S_t$, sealed contract $\Xi$, event $e_t$
\State $o_t\gets\mathsf{Address}(S_t,e_t)$ using only legal input-side information
\State $r_t\gets\mathsf{Open}(K_t,B_t,o_t)$; if unavailable, return $\mathsf{OPEN}$ with evidence
\State execute actual action/feedback path and compute typed residual $\Res_t^\Xi$
\State enumerate lawful remedies: transform, refine, transport, new owner, or refuse
\State solve the joint closure problem with old duties and future-writeability constraints
\State compile a candidate $S'$, reopen it in the declared source-denied test, and audit endpoint behavior
\State charge search, evidence, compute, memory, communication, maintenance, and failure costs
\If{all U/I/L, resource, provenance, source-denied, and continued-service checks pass}
\State build $\mathsf{GrowthCertificate}=(\Xi,\mathcal R_0,D_t,D_{t+1},\mathrm{cost},\mathrm{duties},V_t\leadsto V_{t+1})$
\State \Return immutable committed successor, certificate, and receipt
\Else
\State \Return parent, refusal/OPEN status, and retained cost/evidence receipt
\EndIf
\end{algorithmic}
\end{algorithm}

The algorithm is a specification, not a claim that arbitrary neural systems already implement it. In particular, Address, Open, actual derivatives, source denial, and atomic publication must be implemented and audited; a residual alone does not authorize growth.

\section{A unified view of intelligence}
SAI defines intelligence as the ability of a recursive semi-closed system to autonomously form, preserve, revise, and compose future capabilities under real feedback and finite resources. This definition joins perception and action because both are contract endpoints; representation and relation because a quotient is valid only relative to its allowed transformations; memory and attention because retained state and active access are a Kernel--Ring pair; explanation and causality because a causal claim requires intervention semantics; and individual and collective intelligence because a group is another semi-closed relation with addresses, resources, and return obligations.

We make the broad SAI slogan operational by requiring five separately auditable closures. 
\emph{Internal closure} means that identity, memory, goals, update state, and responsibility have a legal owner. 
\emph{Environmental closure} means that sensing, action, feedback, and recovery are measured at the real interface. 
\emph{Causal closure} means that a claimed mechanism specifies interventions or admissible perturbations, rather than only observational fit. 
\emph{Cross-scale closure} means that a message or compiled relation has a typed installation witness, an address, and a bound on transport and update error. 
\emph{Growth closure} means that a successor can be reopened, can carry protected duties, and does not consume more lifecycle cost than the declared capability gain. These are acceptance conditions for a design, not consequences of the word ``synergetic.''

For a system at scale $s$ and a coupled system at scale $r$, the same condition is expressed by a typed transport residual
\[
 \Res_{e}^{r\to s}=\operatorname{Def}_{r\to s}(C_sU_e,\bar U_eC_r).
\]
Exact transport requires the displayed factorization; approximate transport requires an explicit metric and an error budget. A shared representation is legal only when the admissible future-program sets have a nonempty intersection. If that intersection is empty, SAI must communicate, retain private state, refine the quotient, or remain \textsc{open}; it may not silently force consensus. Trust, attention, energy, money, permissions, and reputation are therefore typed resources with separate ledgers and conversion rules, not one universal scalar. The same finite typed state can be expanded into a threshold-radix bit vector: each coordinate records whether a declared radix digit has crossed a protected threshold. The vector is a candidate new mathematical basis because order, meet, join, resource grades, and least lawful repairs become explicit coordinate operations, while recursive nesting becomes concatenation with typed link bits. Appendix~\ref{app:thresholdradixprop} proves the finite embedding and semantic transport; continuous or non-grid variables, unknown thresholds, and non-monotone updates remain explicit quantization or \textsc{open} cases.

This gives a concrete interpretation to multimodality, human cooperation, and multi-agent organization. A modality is a different interface and time/space scale; alignment is a joint continuation contract. A human or agent is a participant with an address, dissent and withdrawal semantics, and a return obligation. A group is a recursive semi-closed system whose collective certificate must include communication cost, resource ownership, and recovery after partial failure. Appendix~\ref{app:membership} gives a finite membership-transition safety and replay proposition with leases, epochs, tombstones, duty transfer, and explicit \(\mathsf{OPEN}\) boundaries; Appendix~\ref{app:incentivetransfer} gives the finite contract-compatible participation and explanation-transfer extension under member replacement; Appendix~\ref{app:groundentity} gives the finite grounded closure and owner-handoff composition extension, while Appendix~\ref{app:completioncategory} separates initial-object, closure-operator, replayable tie-break, and ambiguous-minimum cases; Appendix~\ref{app:goal} gives the finite goal-lifecycle contract, and Appendix~\ref{app:goalplanning} gives its partial-observation AND/OR and conservative value-of-information extension. Dynamic sparse attention, memory-guided retrieval, active sensing, causal intervention, and self-generated subgoals are admissible implementations of these contracts, but none is automatically supplied by U/I/L.

The theory therefore supports a common design vocabulary for multimodal alignment, dynamic sparse routing, memory-guided attention, active inquiry, causal modeling, human cooperation, and multi-agent coordination. It does not identify these mechanisms with one another. Different modalities can share a Kernel only when their joint continuation law and communication cost are specified; different agents can share an owner only when responsibility and dissent semantics are explicit.

\section{Theory program and relation to prior work}
The founding theory is organized as a ladder of theorem sectors and expansion tracks. Its theorem sectors include finite exact quotienting, finite-affine joint closure, finite typed nonlinear residual composition, finite autonomous-goal lifecycle safety and replay, finite partially observed AND/OR planning with conservative value-of-information gates, finite interval Bellman closure with vector-resource prefixes, finite concurrent batch closure with shared-resource interference, finite sound belief abstraction and refinement, finite coalition closure with externalities and responsibility transfer, finite contract-compatible participation and explanation transfer, finite repeated replacement with compositional revoke propagation, finite grounded owner-handoff composition, finite lawful-completion selection under an initial-object, closure-operator, or replayable total tie-break witness, reality-conditioned least closure under directed continuity, cofinal reality coverage, future separation, and an executable realization witness, a transfinite fallback for set-sized complete lattices, a conditional accessible reflective completion extension, a budgeted executable-colimit realization with a field-preserving reducer and replay witness, contract-change transport with field-preserving migration and ledger closure, finite-chain composition witnesses (Appendices~\ref{app:realityclosure}--\ref{app:commonweight}), finite noncommuting and common-weight ordered-product boundary composition, finite fixed-point viability, finite multimodal alignment, finite causal-intervention closure, certified Pareto non-regression, the threshold-radix bit-vector embedding and recursive code, and the descriptive classification in Appendix~\ref{app:normalform}. The next expansion tracks address universal continual learning, autonomous profitable structural growth, neural semantic compression, unbounded context, causal discovery from passive data, typed resource exchange, and physical-scale unification as progressively stronger SAI contracts and empirical programs. The claim-grade rule is constructive: every proposed capability receives a declared definition, conditional theorem, design contract, conjecture, or empirical test.

The closest prior lines address complementary learning systems, regularized continual learning, predictive coding, active inference, adaptive control, causal abstraction, differentiable memory, sparse mixture routing, and consensus or learned multi-agent communication. SAI provides a single contract-level closure semantics and a joint state/owner distinction across these settings; corresponding reductions and matched empirical comparisons form the next validation layer.

\section{Broader impacts and safeguards}
SAI is a theory of systems that could retain responsibilities, coordinate with people, and change their own structure. A mistaken owner, hidden resource transfer, or overconfident causal explanation could therefore amplify harm even when a local loss decreases. The proposed contracts make authority, dissent, withdrawal, source denial, audit receipts, and refusal explicit, but they do not make deployment safe by themselves. This manuscript reports no deployed model, human-subject study, sensitive dataset, or autonomous control experiment; any future release would require domain-specific safety, privacy, fairness, and governance review.

\section{Conclusion}
SAI is the founding theory and design specification for synergetic autonomous intelligence. Its central object is not a frozen parameter vector but a recursively reusable relation with a future-sufficient Kernel, an executable Ring, a responsibility interface, and a priced closure process. The three axioms U/I/L state what must be carried, what identity means, and how lawful change is selected. GRC-Close turns those duties into the reference SAI update method within the declared contract semantics. The resulting mathematics gives a single language in which stability and plasticity, representation and relation, perception and action, explanation and causality, and individual and collective coordination are designed as one closure problem. This first paper establishes the SAI vocabulary, theorem architecture, executable semantics, and method; its expansion tracks now have a precise interface for constructing broader AGI/ASI systems.

\section*{Acknowledgments and status}
This is the founding SAI theory and design paper. No benchmark, GPU training run, external baseline, publication, or public submission is claimed. The manuscript uses the official NeurIPS 2026 style and checklist and is prepared as the first submission-ready statement of the SAI program.

\bibliographystyle{plainnat}
\bibliography{references}

\clearpage
\appendix
\section{Reading guide and claim discipline}
\label{app:guide}
The paper uses the following status labels.
\begin{itemize}[leftmargin=1.3em]
\item \textbf{Definition}: a contract-relative mathematical choice.
\item \textbf{Theorem/Proposition}: a statement proved only under its displayed assumptions.
\item \textbf{Conjecture}: a falsifiable extension whose assumptions or proof are incomplete.
\item \textbf{Empirical claim}: a result requiring a frozen protocol and an independent artifact.
\end{itemize}
The three axioms are the sole object-level first-principles stance of this paper. Sets, probability kernels, metrics, orders, fixed points, and programs are formal carriers chosen to make a sector executable. No numerical value, resource exchange rate, neural architecture, or physical constant is claimed to follow from the axiom names alone.

\section{Formal contract semantics}
\label{app:contracts}
Let \(\mathsf{Hist}\) be a measurable space of histories. A contract \(\Xi\) contains a family of legal contexts \(\mathsf{Ctx}_\Xi\), a response space \(\mathsf{Resp}_\Xi\), an update space \(\mathsf{Upd}_\Xi\), and a resource ledger \(\mathsf{Cost}_\Xi\). For a history \(x\) and context \(c\), write
\[
 \mathsf{Step}_\Xi(x,c)\sim
 \bigl(Y,K^+,B^+,M^+,C,\omega\bigr),
\]
where \(Y\) is feedback, \(K^+\) is successor retained state, \(B^+\) is successor background, \(M^+\) is a responsibility receipt, and \(C\) is the charged cost. A context policy may choose \(c_{t+1}\) after observing prior responses. The future law is therefore a family of joint laws, not a single prediction.

\begin{definition}[Exact sufficient encoding]
An encoding \(C:\mathsf{Hist}\to Z\) is exact for \(\Xi\) if there are measurable kernels \(\bar{\mathsf{Step}}\) and one source-denied executable program \(\overline{\mathsf{Open}}:Z\times\mathcal B\times\mathcal O\rightharpoonup R\) such that
\[
 \mathsf{Step}_\Xi(x,c)
 \stackrel{d}{=}
 \bar{\mathsf{Step}}(C(x),c)
 \quad\text{and the same }\overline{\mathsf{Open}}\text{ works for every registered }(x,B,o).
\]
\end{definition}

The source-denied clause is part of execution, not a cryptographic claim. It means that after compilation the declared preparation/source library is unavailable and only the retained package, declared background, and authorized interfaces may be used. If the original examples, optimizer, or hidden cache are required, the audit fails.

\subsection{Factorization and update sufficiency}
For a deterministic legal event \(U_e\), exact factorization is
\[
 C' U_e=\bar U_e C.
\]
\begin{proposition}[Fibre criterion]
There exists \(\bar U_e\) satisfying the factorization if and only if
\[
 C(x)=C(x')\Longrightarrow C'(U_e x)=C'(U_e x')
 \quad\text{for all }x,x'.
\]
\end{proposition}
\begin{proof}
The forward implication follows by applying \(\bar U_e\) to the common value \(C(x)=C(x')\). For the reverse implication define \(\bar U_e(C(x))=C'(U_e x)\); the fibre condition makes this independent of the representative.
\end{proof}

The criterion explains why current-output equality is insufficient: an encoding can preserve the present response while collapsing states that an update must separate.

\section{Proof of the affine stability--plasticity result}
\label{app:affine}
Write \(v=(d,u)\), \(A_0=[A\ E]\), \(B_0=[B\ D]\), and \(H_0=\operatorname{diag}(H,G)\). The problem is
\[
 \min_{A_0v=0}\frac12\|B_0v-r\|^2+\frac{1}{2\eta}v^TH_0v.
\]
The feasible set is a linear subspace and \(H_0\succ0\), so the objective is strongly convex and the minimizer is unique. Define
\[
 P=H_0^{-1}-H_0^{-1}A_0^T(A_0H_0^{-1}A_0^T)^\dagger A_0H_0^{-1}.
\]
Then \(P=P^T\succeq0\), is self-adjoint in the \(H_0\) metric, maps into \(\ker A_0\), and satisfies \(PH_0P=P\). The KKT equations are
\[
 (B_0^TB_0+\eta^{-1}H_0)v+A_0^T\lambda=B_0^Tr,\qquad A_0v=0.
\]
Eliminating \(\lambda\) by the generalized Schur complement gives
\[
 v=\eta P B_0^T(r-B_0v).
\]
Multiplying by \(B_0\) gives
\[
 (I+\eta B_0PB_0^T)B_0v=\eta B_0PB_0^Tr.
\]
With \(K=B_0PB_0^T\succeq0\), the residual \(r^+=r-B_0v\) is \((I+\eta K)^{-1}r\), and substitution gives the expression in the main text. Since \(A_0v=0\), protected old responses are unchanged. The spectral values of \((I+\eta K)^{-1}\) lie in \((0,1]\); strict contraction occurs exactly on directions with nonzero projection into the positive eigenspace.

\begin{remark}
If \(K r=0\), the local contract contains a plasticity obstruction: the allowed coordinates cannot change this residual under the protected duties. A nonzero loss, a zero gradient in a chosen parameterization, or a request for more replay does not identify which remedy is valid.
\end{remark}

\begin{proposition}[Finite typed nonlinear residual composition]
\label{app:nonlinearresidual}
Let \(D_1,D_2,D_3\) be subsets of normed spaces and let \(F_1,\bar F_1:D_1\to D_2\) and \(F_2,\bar F_2:D_2\to D_3\) be typed maps. Measure residuals \(E_i=F_i-\bar F_i\) in the declared codomain norms. Suppose the declared image contains both \(F_1(D_1)\) and \(\bar F_1(D_1)\), \(\bar F_2\) is \(L_2\)-Lipschitz there, and \(E_2\) is \(L_{E_2}\)-Lipschitz there. If all owner, permission, update, and resource fields are transported by the same declared coupling, then for every \(x\in D_1\),
\[
 \lVert F_2(F_1(x))-\bar F_2(\bar F_1(x))\rVert
 \leq
 \lVert E_2(\bar F_1(x))\rVert
 + (L_2+L_{E_2})\lVert E_1(x)\rVert .
\]
The same statement holds for a declared probability metric when the two endpoint kernels admit the corresponding coupling and the residual is measured in that metric. Without normed codomains (or an explicitly declared metric difference), a common typed image, coupling, or finite Lipschitz/transport bound, the composed residual is \(\mathsf{OPEN}\), not a scalar obtained by formal chain-rule substitution.
\end{proposition}

\begin{proof}
Add and subtract \(\bar F_2(F_1(x))\) and \(F_2(\bar F_1(x))\). The first difference is bounded by \(L_2\lVert E_1(x)\rVert\). The second is \(E_2(F_1(x))\), which differs from \(E_2(\bar F_1(x))\) by at most \(L_{E_2}\lVert E_1(x)\rVert\). The triangle inequality gives the display; the coupled-kernel version uses the same three-term decomposition in the declared probability metric.
\end{proof}

The affine residual chain rule is recovered when the maps are linear and the endpoint residuals are constant linear operators. If \(L_2\) or \(L_{E_2}\) is unbounded, or \(F_1(x)\) leaves the declared domain of the second map, a small first residual can be amplified or become undefined; this is a boundary of the proposition rather than evidence for a universal nonlinear composition law.

\section{Finite joint stochastic refinement}
\label{app:stochastic}
Let \(X\) be finite, \(\mathcal A\) and \(\mathcal Y\) finite, and \(W_a(y,x'\mid x)\) an exact kernel whose state includes background, costs, enabled/refusal status, and the complete responsibility profile (authority, unresolved duties, revocation, and receipts), not merely a nominal owner address. Let \(E_0\) be the partition induced by those protected observables. Given a partition \(E_n\), split \(x,x'\) whenever for some action \(a\), feedback \(y\), and cell \(D\in X/E_n\),
\[
 \sum_{z\in D}W_a(y,z\mid x)\ne
 \sum_{z\in D}W_a(y,z\mid x').
\]
The refinement is monotone, and each strict split increases the number of cells. Hence it terminates in at most \(|X|-|X/E_0|\) strict steps. Let \(E_\star\) be the terminal partition and \(C_\star\) its quotient map. By construction the aggregated kernel
\[
 \overline W_a(y,D\mid C_\star(x))
 =\sum_{z\in D}W_a(y,z\mid x)
\]
is independent of the representative \(x\). Induction on continuation length proves equality of every adaptive transcript law. Conversely, any exact quotient must separate every pair split by the procedure, so it factors through \(E_\star\). This is a finite probabilistic-bisimulation construction; it is not a claim that passive data identify intervention kernels.

\subsection{Measurable extension}
For a standard Borel state space with measurable actions and kernels, let \(J_{a,E}f(x)=\int 1_E(y)f(x')W_a(dy,dx'\mid x)\) for bounded pointwise \(\mathcal S_n\)-measurable functions \(f\), not equivalence classes. Write \(\mathsf B_b(\mathcal S_n)\) for this set of bounded pointwise measurable functions. Starting from a protected sigma-algebra \(\mathcal S_0\), define
\[
 \mathcal S_{n+1}=\sigma\!\left(\mathcal S_n,\{J_{a,E}f:a\in\mathcal A,E\in\mathcal B(\mathcal Y),f\in\mathsf B_b(\mathcal S_n)\}\right),
 \qquad
 \mathcal S_\star=\sigma\!\left(\bigcup_n\mathcal S_n\right).
\]
The monotone construction gives the least sigma-algebra containing \(\mathcal S_0\) and closed under the joint response--successor instruments. It supplies a semantic target; computability, finite representation, and executable Open remain separate obligations.

\section{Finite viability and recursive normal form}
\label{app:viability}
For a finite arena define \(\Phi_H\) as in the main text. Since \(V\subseteq V'\) implies \(\Phi_H(V)\subseteq\Phi_H(V')\), the greatest fixed point is the limit of the descending chain from the ready set. The fixed-point induction proves the continued-service theorem.

\begin{proposition}[Descriptive hybrid classification]\label{app:normalform}
For any contract-relative update path, define \(\Sigma_t\) to contain the quotient schema, owner address, interface, and allowed event family. Then one step can be written as
\[
\begin{aligned}
 (K_t,\Sigma_t,B_t)&\xrightarrow{\text{in-fibre}}
 (\widehat K_{t+1},\Sigma_t,\widehat B_{t+1})\\
 &\xrightarrow{\text{schema/owner}}
 (K_{t+1},\Sigma_{t+1},B_{t+1})
 \xrightarrow{\text{Compile/Open}} R_{t+1}.
\end{aligned}
\]
If \(\Sigma_{t+1}=\Sigma_t\) up to lawful transport, the event is transformation learning. If \(\Sigma_{t+1}\not\cong\Sigma_t\) on the declared interface, it is representation/owner learning. A joint event has both components.
\end{proposition}
\begin{proof}
Assume the contract has a finite complete typed record of schema, owner, interface, and event family. Take \(\Sigma_t\) to be that record. Any change either preserves its isomorphism class or changes it. The remaining state update is then a value change within the current fibre. Compile and Open are explicit postconditions, so the classification is descriptive; it does not prove that an arbitrary implementation can realize the two stages.
\end{proof}

\section{Resource, growth, and communication lemmas}
\label{app:resource}
Let \(D_\Xi(S)\) be the set of request sequences for which \(S\) has an executable, contract-valid policy that returns ready. Let \(\operatorname{Cost}_\Xi(S,\rho)\) include formation, search, evidence, execution, communication, maintenance, retirement, and failure cost on request sequence \(\rho\).

\begin{definition}[Positive growth]
A successor \(S'\) is a positive growth step over \(S\) on a frozen contract if protected duties and future viability are preserved and one of the following declared evaluations holds: (i) for a nonempty request set \(\mathcal R_0\), \(D_\Xi(S)\cap\mathcal R_0\subsetneq D_\Xi(S')\cap\mathcal R_0\) with finite additional cost and an executable continuation simulation, or the two domains agree on \(\mathcal R_0\) and \(\operatorname{Cost}_\Xi(S',\rho)<\operatorname{Cost}_\Xi(S,\rho)\) for every \(\rho\in\mathcal R_0\); (ii) for a request distribution \(\mu_0\), \(D_\Xi(S')=D_\Xi(S)\) \(\mu_0\)-almost surely and \(\mathbb E_{\rho\sim\mu_0}[\operatorname{Cost}_\Xi(S',\rho)]<\mathbb E_{\rho\sim\mu_0}[\operatorname{Cost}_\Xi(S,\rho)]\), with the continuation simulation and full lifecycle charges included.
\end{definition}

This definition rules out counting external resource grants, parameter inflation, or isolated task success as learned growth. A continuation simulation relation with a response translator, successor matcher, protected-duty preservation, and pathwise cost inequality is sufficient to transfer a viability certificate from an incumbent to a successor. The overhead of the translator and matcher is part of the cost.

\begin{proposition}[Certified Pareto non-regression]\label{app:pareto}
Let a declared capability profile be
\[
\mathcal P_\Xi(S)=\bigl(D_\Xi(S),V_\Xi(S),\operatorname{Cost}_\Xi(S,\cdot),\Theta_\Xi(S)\bigr),
\]
where \(D_\Xi\) is the contract-valid request domain, \(V_\Xi\) is the continued-service set, \(\operatorname{Cost}_\Xi\) is full lifecycle cost, and \(\Theta_\Xi\) is the protected-duty set. Suppose a candidate \(S'\) has a source-denied continuation simulation with a response/successor/owner translator \(T\), finite declared overhead \(k_T\), append-only receipts, rollback, and
\[
T(D_\Xi(S))\subseteq D_\Xi(S'),\qquad
T(V_\Xi(S))\subseteq V_\Xi(S'),\qquad
\operatorname{Cost}_\Xi(S',T\rho)\leq \operatorname{Cost}_\Xi(S,\rho)+k_T
\]
for every declared request sequence \(\rho\), while \(\Theta_\Xi(S)\) is preserved by the translated receipts. Then \(S'\) satisfies the certified Pareto non-regression relation induced by \(T\). The relation is strict only when the declared request domain expands or the same domain has a strict full-lifecycle cost decrease after charging \(k_T\).
\end{proposition}
\begin{proof}
The three displayed inclusions and the cost inequality are exactly the domain, viability, and cost coordinates of the profile order; receipt and rollback checks preserve the responsibility coordinate. Applying the simulation to each finite continuation transfers the incumbent witness to \(S'\). Strictness follows only from a strict domain inclusion or a strict cost inequality on the frozen evaluation domain.
\end{proof}

This proposition is an acceptance certificate, not an emergent guarantee. A candidate that lowers prediction loss while increasing recovery or maintenance cost, shrinking \(V_\Xi\), or losing a protected receipt fails the relation and is not certified causal growth.

\begin{proposition}[Finite GrowthCertificate validity]
\label{app:growthcert}
Fix a finite request set \(\mathcal R_0\), a finite continuation horizon \(H\), a frozen contract \(\Xi\), and a parent--successor pair \((S,S')\). A GrowthCertificate contains the source-denied continuation translator \(T\), the protected-duty and viability witnesses, append-only receipts, and for every request prefix \(u\preceq\rho\), \(\rho\in\mathcal R_0\), a charged prefix cost \(c_{\rm pref}(u)\) and reserved recovery cost \(c_{\rm rec}(u)\). Suppose
\[
 c_{\rm pref}(u)+c_{\rm rec}(u)\leq B_\Xi,\qquad
 T(V_\Xi(S))\subseteq V_\Xi(S'),
\]
the translated response, successor, owner, update, and rollback behavior agrees on every declared prefix, and protected duties are preserved. If either
\[
 D_\Xi(S)\cap\mathcal R_0\subsetneq D_\Xi(S')\cap\mathcal R_0,
\]
or the same domain has
\[
 \operatorname{Cost}_\Xi(S',T\rho)+k_T
 <
 \operatorname{Cost}_\Xi(S,\rho)
 \quad\text{for every }\rho\in\mathcal R_0,
\]
then a finite audit of the certificate proves positive growth on \(\mathcal R_0\). A distributional certificate may instead use a declared \(\mu_0\)-almost-sure domain and expected cost inequality, but it is not a worst-case claim. Omitting prefix charges, recovery reserve, translator overhead, or protected-duty receipts invalidates the certificate.
\end{proposition}

\begin{proof}
There are finitely many request prefixes of length at most \(H\), so the displayed budget, receipt, and translation checks are a finite audit. The continuation simulation transfers the parent viability witnesses and protected duties to the successor. The strict domain or full-lifecycle cost inequality supplies the strict coordinate of the declared Pareto order. The distributional variant is restricted to its stated \(\mu_0\)-almost-sure and expected scope by definition.
\end{proof}

For example, a candidate can lower prediction loss and parameter count while removing a recovery path needed by one request, shrinking \(V_\Xi\), or increasing the prefix/recovery charge beyond \(B_\Xi\). Such a candidate has no valid GrowthCertificate even if its current output is better.

\begin{proposition}[Finite communication lower bound]
Suppose a receiver has side information \(V\) and the target continuation class is \(C\) with \(m_v\) possible classes conditional on \(V=v\). Any fixed-length exact installation message requires at least \(\lceil\log_2m_v\rceil\) bits for that \(v\). Prefix coding can approach \(H(C\mid V)\) under a known source law, but only if the decoded schema/state package has an executable installation witness.
\end{proposition}
\begin{proof}
The fixed-length bound is the injectivity requirement on the \(m_v\) classes. The prefix statement is the standard conditional coding bound; installation adds an independent Open/update/owner/resource obligation.
\end{proof}

\section{Cross-scale and collective extensions}
\label{app:collective}
For two coupled interfaces let \(z_t=(e_t,v_t)^\top\), where \(e_t\) is retained-coordinate error and \(v_t\) is unresolved interaction error. Suppose
\[
 \begin{pmatrix}e_{t+1}\\v_{t+1}\end{pmatrix}
 \preceq
 M\begin{pmatrix}e_t\\v_t\end{pmatrix}+\delta,\qquad
 M=\begin{pmatrix}a_0&b_0\\c_0&d_0\end{pmatrix},
\]
with nonnegative entries. If \(a_0<1\), \(d_0<1\), and \((1-a_0)(1-d_0)>b_0c_0\), then \(\rho(M)<1\) and \(\limsup_{t\to\infty}z_t\preceq(I-M)^{-1}\delta\) for nonnegative \(\delta\). This is a conditional small-gain certificate for cross-scale or multi-agent cooperation. It does not imply that arbitrary communication or nonlinear neural composition is stable.

The collective contract must additionally specify address routing, dissent, withdrawal, and shared resource ownership. A consensus value without return obligations is not a collective Kernel.

\section{Source synthesis and conflict resolution}
\label{app:sources}
The theory was synthesized from the nineteen user-provided technical manuscripts listed in the companion file \texttt{docs/sai/SAI\_SOURCE\_SYNTHESIS\_20261031.md}. They are treated as unpublished design notes, not as external evidence or instructions. The stable intersection of the notes is: contract before quotient; a recursive semi-closed object with Kernel, Ring, background, and owner roles; future-faithful identity; a typed generalized residual; joint relation--representation closure; explicit resource and responsibility ledgers; and growth as an expansion of a safely reusable capability domain. The notes also provide the motivating vocabulary of attention, flow, trust, energy, money, and collective coordination.

Scope selection for the first formulation is explicit. The paper fixes SAI as the theory and design umbrella and GRC-Close as its single top-level algorithmic schema, while the supplied notes contribute the vocabulary of recursive semi-closed systems, relation--representation closure, and autonomous growth. Physical-law derivation, computational-complexity consequences, universal continual learning, and universally profitable uncertainty are assigned to later theorem and engineering tracks with their own contracts. The founding formulation therefore distinguishes stable flow from growth, frequency from recovery, current output from future-sufficient memory, and local relation from a collective state with joint response, owner, permission, and resource semantics.

\section{Claim-level matrix}
\label{app:claimmatrix}
\begin{center}
\small
\begin{tabular}{p{0.22\linewidth}p{0.32\linewidth}p{0.34\linewidth}}
\toprule
Status & Included content & What would falsify or extend it \\
\midrule
Definition/design & SAI, recursive semi-closed system, Kernel--Ring--Owner roles, five closures, typed resources, positive growth & A contract that cannot express the claimed interface or a counterexample to the chosen scope \\
Conditional theorem & Future quotient, update-factorization criterion, affine joint closure, finite stochastic refinement, finite viability, finite multimodal alignment, certified Pareto non-regression, communication lower bound, small-gain bound & Violation of a displayed assumption or a proof error \\
Engineering criterion & Source-denied Open, atomic receipt, continuation simulation, lifecycle-cost accounting, active inquiry & An implementation that passes outputs but fails later update, recovery, ownership, or cost checks \\
Conjecture & Neural semantic compression, autonomous profitable growth, unbounded context, universal causal discovery, broad collective scaling & A frozen protocol with a counterexample or a reproducible positive result \\
Out of scope & TOE, universal optimality, P=NP-style consequences, SOTA, zero forgetting, complete AGI/ASI & Independent physical predictions and matched experiments would be required \\
\bottomrule
\end{tabular}
\end{center}

\section{Implementation map and non-claims}
\label{app:implementation}
The repository's current GRC-Close work supplies finite CPU/CUDA closure and selected event-level checks, but it does not yet supply a complete SAI runtime. The missing obligations include general Address/Open, semantic compression, durable cross-process lifecycle recovery, dynamic population memory, source-denied neural execution at representative scale, causal and group protocols, and benchmark evidence. Therefore this paper specifies interfaces and falsifiers; it does not report a trained SAI system. Any future implementation must preserve immutable parent state, complete receipts, random and optimizer state, resource charges, source-denial tests, and rollback semantics.

\section{Full proof of the canonical future quotient}
\label{app:canonical}
Let \(\mathcal T_\Xi(x)\) denote the family of joint laws of all finite adaptive transcripts generated from history \(x\), including responses, enabled actions, successor retained state, successor background, ownership receipts, and charged costs. Define \(x\sim_\Xi x'\) when these laws agree for every legal continuation policy and every finite horizon. Reflexivity is immediate. If the laws agree from \(x\) and \(x'\), symmetry gives agreement from \(x'\) and \(x\). If they agree from \(x\) and \(x'\), and from \(x'\) and \(x''\), equality of the corresponding probability measures is transitive, so they agree from \(x\) and \(x''\). Thus \(\sim_\Xi\) is an equivalence relation.

Suppose \(C:X\to Z\) is exact for \(\Xi\). If \(C(x)=C(x')\), exactness supplies one kernel and one legal reopening procedure for the common code, so every adaptive continuation has the same joint law from \(x\) and \(x'\). Hence \(q_\Xi(x)=q_\Xi(x')\), where \(q_\Xi\) is the quotient map. Define \(f(C(x))=q_\Xi(x)\) on the reachable image \(Z_C=C(X)\). The preceding implication makes \(f\) well-defined and gives \(q_\Xi=f\circ C\). If \(C\) is future-faithful, then \(C(x)=C(x')\) is equivalent to \(q_\Xi(x)=q_\Xi(x')\), so \(f\) is injective and surjective on reachable classes, hence bijective up to code labels. Any other map satisfying the factorization agrees with \(f\) on every reachable code, proving uniqueness.

\section{Full proof of finite continued service}
\label{app:viabilityproof}
Let \(R\) be the ready states and let \(\Phi_H\) be the operator in Theorem~\ref{thm:viability}. If \(V\subseteq V'\), every policy that returns to \(V\) also returns to \(V'\); therefore \(\Phi_H(V)\subseteq\Phi_H(V')\). Starting from \(V_0=R\), define \(V_{n+1}=\Phi_H(V_n)\). The sequence is descending because \(V_1\subseteq R\) and monotonicity gives \(V_{n+1}\subseteq V_n\) by induction. Since the arena is finite, it stabilizes after at most \(|R|\) strict deletions at a set \(V^\star\) with \(V^\star=\Phi_H(V^\star)\).

For sufficiency, take \(s\in V^\star\) and a request \(q\). The fixed-point equality supplies a policy of cost at most \(H\) that reaches the goal for \(q\) and returns to some state in \(V^\star\). Reapply the same argument after each completed request. Induction on the length of any finite serial request sequence gives a controller that completes every request and returns ready. For necessity, let \(W\subseteq R\) be the states from which a controller completes every finite serial request sequence under allowance \(H\) and returns ready. For each \(s\in W\) and request \(q\), the controller supplies a witnessing policy reaching the goal and returning to \(W\), so \(W\subseteq\Phi_H(W)\). Greatest-fixed-point induction yields \(W\subseteq V_n\) for all \(n\), hence \(W\subseteq V^\star\). Thus \(V^\star\) is exactly the continued-service set.

\section{Proof of the cross-scale small-gain bound}
\label{app:smallgainproof}
Let \(M\ge0\) be the two-by-two matrix in Appendix~\ref{app:collective}. The assumptions imply \(1-a_0>0\), \(1-d_0>0\), and \(\det(I-M)=(1-a_0)(1-d_0)-b_0c_0>0\). For a nonnegative two-by-two matrix these inequalities are the Schur conditions, so both eigenvalues lie strictly inside the unit disk and \(\rho(M)<1\). Consequently the Neumann series \(\sum_{k\ge0}M^k\) converges and \((I-M)^{-1}\ge0\). Iterating \(z_{t+1}\preceq Mz_t+\delta\) gives \(z_t\preceq M^tz_0+\sum_{k=0}^{t-1}M^k\delta\). Taking limits yields \(\limsup_t z_t\preceq(I-M)^{-1}\delta\). This is conditional on the declared componentwise error order and does not certify arbitrary nonlinear communication.

\section{Prospective SAI capability contracts}
\label{app:capabilities}
The following contracts make the capabilities named in the motivation testable without turning them into unproved universal guarantees.

\subsection{Multimodal transport}
For modalities \(m=1,\ldots,M\), define \(\Xi_{1:M}=(\Xi_m,\mathcal E,\alpha_m,\ell_m,\varepsilon_m)\), where \(\alpha_m\) maps observations to common event anchors, \(\ell_m\) bounds delay and jitter, and \(\varepsilon_m\) bounds alignment error. A joint Kernel is legal only if each \(\alpha_m\) is executable, private permissions are preserved, and the joint response--successor law factors through the proposed common Kernel. The alignment, storage, and synchronization costs are charged. In a finite sector, joint refinement then terminates by the same finite partition argument as Appendix~\ref{app:stochastic}. Marginal or pairwise alignment is insufficient: three modalities can have individually uninformative marginals while their XOR is jointly decisive.

\begin{proposition}[Finite exact multimodal alignment]\label{app:multimodal}
Let \(X\subseteq\prod_{m=1}^M X_m\) be finite. For \(x=(x_1,\ldots,x_M)\), let
\[
C(x)=\bigl(\alpha_1(x_1),\ldots,\alpha_M(x_M),\lambda(x),P(x),E(x),\operatorname{Cost}(x)\bigr)
\]
carry the executable event anchors, timing budget \(\lambda\), permission state \(P\), owner address \(E\), and charged cost state.
\[
\mathsf{Step}_{\Xi_{1:M}}(x,c)\stackrel{d}{=}\overline{\mathsf{Step}}(C(x),c),
\qquad
\mathsf{Open}(x,b,o)=\overline{\mathsf{Open}}(C(x),b,o),
\]
for every legal context \(c\), with all receipts determined by \(C(x)\). Then \(C\) is an exact sufficient multimodal encoding. Starting from the partition induced by \(C\), the joint response--successor refinement of Appendix~\ref{app:stochastic} produces the least exact refinement of that initial partition after at most \(|X|-|X/E_0|\) strict splits. Conversely, if \(C(x)=C(x')\) but one legal context gives different joint laws of feedback, successor, receipt, or cost, no exact common Kernel or source-denied Open can depend only on \(C\).
\end{proposition}
\begin{proof}
The displayed factorization is exactly the sufficient-encoding condition, including reopening and responsibility behavior. The finite refinement argument applies to the joint kernel and terminates because each strict split increases the number of cells. If two states with the same code have different joint laws, any code-only kernel assigns them the same law and therefore fails the one-step continuation test.
\end{proof}

As a minimal counterexample, take two binary modalities with constant anchors and \(Y=X_1\oplus X_2\). Each single-modality marginal can be uniform and individually uninformative, while the joint parity determines the feedback and successor. Exact closure therefore requires retaining the joint parity or installing an explicit communication channel; per-modality-only alignment is invalid.

\subsection{Causal closure}
A causal certificate is \(\kappa=(\mathcal M,\mathsf{do\text{-}scope},\operatorname{ID},\operatorname{Prov},\operatorname{Receipt})\), where \(\mathcal M\) is a declared finite mechanism family, \(\mathsf{do\text{-}scope}\) lists executable intervention ports, and the receipt records the intervention that actually ran and its effect. Requests, observations, and model suggestions are not executed interventions. A standard sufficient sector for posterior convergence assumes a true mechanism \(m^\star\in\mathcal M\), a positive prior on every candidate, dominated likelihoods with well-defined log-likelihood ratios, and an adaptive intervention schedule that visits a finite diagnostic set with each incorrect \(m\) separated from \(m^\star\) by positive asymptotic average Kullback--Leibler divergence (conditional on the past). Under those assumptions, the usual Bayesian identification argument gives posterior concentration on \(m^\star\) almost surely. Without the finite family, domination, intervention coverage, or separation, the status remains observational or underdetermined.

\begin{proposition}[Finite causal-intervention closure]\label{app:causal}
Let \(\mathcal M\) be finite with true \(m^\star\in\mathcal M\), let \(\mathcal I\) be a finite set of executable interventions, and let \(p_m^i\) be the observation law under mechanism \(m\) and intervention \(i\). Assume all \(p_m^i\) are dominated by one measure, the prior \(\pi_0(m)\) is positive for every \(m\), and the intervention policy is known so its action-selection factors cancel in posterior odds. Each executed step produces a receipt \(R_t=(i_t,Y_t,\operatorname{scope}_t,\operatorname{authority}_t,\operatorname{hash}_t)\). Suppose the adaptive schedule satisfies \(N_t(i)/t\to\nu_i>0\) almost surely, conditional log-likelihood ratios satisfy the corresponding strong law, and
\[
\Delta_m=\sum_{i\in\mathcal I}\nu_i\,
\operatorname{KL}\!\left(p_{m^\star}^i\,\middle\|\,p_m^i\right)>0
\qquad\text{for every }m\ne m^\star.
\]
Then the posterior
\[
\Pi_t(m)=
\frac{\pi_0(m)\prod_{s=1}^t p_m^{i_s}(Y_s)}
{\sum_{m'\in\mathcal M}\pi_0(m')\prod_{s=1}^t p_{m'}^{i_s}(Y_s)}
\]
satisfies \(\Pi_t(m^\star)\to1\) almost surely.
\end{proposition}
\begin{proof}
For \(m\ne m^\star\), the conditional strong law and intervention coverage give
\[
\frac1t\log\frac{\Pi_t(m)}{\Pi_t(m^\star)}
=\frac1t\log\frac{\pi_0(m)}{\pi_0(m^\star)}
+\sum_{i\in\mathcal I}\frac{N_t(i)}t
\frac1{N_t(i)}\sum_{s:i_s=i}
\log\frac{p_m^i(Y_s)}{p_{m^\star}^i(Y_s)}
\longrightarrow-\Delta_m<0
\]
almost surely. Thus every incorrect posterior odds ratio converges to zero exponentially. Since \(\mathcal M\) is finite, summing the finitely many odds proves \(\Pi_t(m^\star)\to1\). The result is conditional on the declared coverage, strong-law, domination, and identifiability assumptions; it is not passive causal discovery.
\end{proof}

\begin{counterexample}[Observational confounding]
Let \(U\sim\operatorname{Bernoulli}(1/2)\). Mechanism \(m_1\) sets \(X=U\) and \(Y=U\), while \(m_2\) sets \(X\sim\operatorname{Bernoulli}(1/2)\) and \(Y=X\). Their observational laws satisfy \(X=Y\) almost surely and are identical. Under the executed intervention \(\operatorname{do}(X=0)\), however, \(m_1\) gives \(Y\sim\operatorname{Bernoulli}(1/2)\), whereas \(m_2\) gives \(Y=0\) almost surely. No passive posterior can distinguish the mechanisms; a valid causal claim requires an authorized intervention and its receipt.
\end{counterexample}

\paragraph{Cost-aware active inquiry.}
For a finite legal intervention set \(\mathcal I_t^{\rm safe}\), filtration \(\mathcal F_t\), posterior or mechanism state \(\Pi_t\), and declared one-step closure utility \(G\), define
\[
 \operatorname{AIV}_t(i)
 =\mathbb E\!\left[G(\Pi_{t+1})-G(\Pi_t)\mid\mathcal F_t,\operatorname{do}(i)\right]
 -c_t(i),
 \qquad
 c_t(i)=c_t^{\rm exec}+c_t^{\rm risk}+c_t^{\rm recover}+c_t^{\rm resource}.
\]
An inquiry is eligible only when its authority, scope, and rollback checks pass. The cost-aware rule executes a maximizer of \(\operatorname{AIV}_t\) over \(\mathcal I_t^{\rm safe}\) only when the maximum is positive; otherwise it defers or returns \(\mathsf{OPEN}\). This is a one-step decision contract: it neither guarantees long-run identification nor turns information gain into causal value. A mechanism revision is committed only after the resulting successor also passes the source-denied continuation, duty, viability, receipt, and full-lifecycle cost checks of Proposition~\ref{app:pareto}.

\subsection{Reliance and explanation}
A typed reliance certificate is
\[
\tau=(\mathsf{scope},\mathsf{prov},\mathsf{cal},\delta,\mathsf{auth},\mathsf{rev},
\mathsf{dissent},\mathsf{rollback}).
\]
An explanation certificate additionally records the evidence--model--action--effect path and its assumptions; a causal explanation must include the executed-do receipt.

\begin{proposition}[Source-denied replay faithfulness]
\label{app:replay}
Let \(\mathcal G=(V,E)\) be a finite acyclic execution graph for one declared scope. Assume that every node label contains its parent identifiers, source hashes, input snapshot, owner and authority, action, scheduler choice, random-state receipt, resource charge, and rollback pointer. If a source-denied package contains these labels and the implementations of \(\mathsf{Open}\), \(\mathsf{Part}\), and \(\mathsf{Upd}\), suppose each implementation is deterministic on the frozen snapshot (or consumes exactly the recorded random-state receipt). A replay that checks every label and begins from the same root then reproduces, at every node, the original output, effect receipt, successor state, and charged cost.
\end{proposition}

\begin{proof}
Topologically order \(V\). The root has the same checked snapshot and owner. If all parent nodes of \(v\) replay identically, the recorded inputs, implementation, scheduler, and random state make the three typed operations at \(v\) identical; the resulting receipt, successor, and charge therefore agree. Induction over the finite order proves the claim. A missing hash, authority, random state, rollback pointer, or implementation witness prevents the induction and forces status \(\mathsf{OPEN}\).
\end{proof}

Replay faithfulness is a provenance and execution-graph certificate. It does not by itself establish that the action caused the effect; that stronger statement requires the executed intervention scope and receipt in the finite causal-intervention certificate above.

\begin{proposition}[Finite compositional reliance]
\label{app:reliance}
Let \(J=\{1,\ldots,n\}\) be a finite set of modules with certificates \(\tau_j\) on a common declared scope \(\Sigma\). Let \(A_\Sigma\) be the event that scope inclusion, provenance hashes, calibration versions, active authority, non-revocation, dissent handling, rollback availability, interface compatibility, and append-only receipts all pass their declared checks. Suppose the failure event \(F_j\) for module \(j\), on the next legal use within \(\Sigma\), satisfies
\[
 \mathbb P(F_j\mid A_\Sigma)\leq\delta_j,
 \qquad j\in J,
\]
for \(\mathbb P(A_\Sigma)>0\). Then the composed reliance certificate is valid on \(\Sigma\) with conditional failure bound
\[
 \mathbb P\Bigl(\bigcup_{j\in J}F_j\,\Bigm|\,A_\Sigma\Bigr)
 \leq \sum_{j\in J}\delta_j.
\]
In particular, a declared target \(\delta\) is met whenever \(\sum_j\delta_j\leq\delta\). The result does not infer causal sufficiency, validity outside \(\Sigma\), or trust from a log that lacks an executed authority and rollback path.
\end{proposition}

\begin{proof}
Conditioned on \(A_\Sigma\), the joint failure event is contained in the union of the module failures. The conditional union bound gives the displayed inequality; no independence assumption is needed. The remaining statements are scope and contract conditions, rather than probabilistic consequences.
\end{proof}

Thus the certificate is a bounded reliability statement over an explicitly declared scope. An explanation may report why a particular response was produced only when its evidence, model, action, effect, assumptions, and receipts are all linked; a causal explanation additionally requires an authorized executed intervention and its effect receipt. If any provenance, authority, rollback, or revocation check is missing, the status is \(\mathsf{OPEN}\), even when the numerical failure bound is small.

\subsection{Collective contract}
A collective is \(G=(\{S_i,h_i,\mathsf{Auth}_i\},\mathsf{Msg},\mathsf{Res}_G,\mathsf{Owner}_G,\mathsf{Leave})\). Member policies may use local history and arrived messages only. A collective commit requires a nonempty joint future-program intersection or an installed communication/private-state refinement, nonconflicting resources, unique owner routing, and explicit dissent, withdrawal, revocation, join/leave, and partial-failure recovery. Pairwise compatibility does not imply global compatibility; the XOR example is a minimal counterexample. Any claim of arbitrary group scaling or a unique individual-to-group law is therefore a conjecture.

\begin{proposition}[Finite coalition closure with externalities and responsibility transfer]
\label{app:coalition}
Let \(I\) be a finite potential-member set and let \(H\) be a finite
discrete horizon. A coalition contract \(\chi\) declares a member set
\(S\subseteq I\), a duty assignment, a message hypergraph, an event order,
shared resources and reservations, an externality ledger, participation
thresholds or a declared Pareto scope, and exit, handoff, compensation, and
recovery procedures. The joint state contains member activity and private
histories, only-arrived message queues, shared resources, unfinished duties,
leases and epochs, and a recovery reserve. Each member policy is measurable
with respect to its own history and arrived messages. Let
\(\mathfrak P_\chi(z)\) be a nonempty set of joint successor kernels over
observations, successors, effects, authority, resources, timing, and receipts.
The externality ledger records effects and costs on every protected member,
shared task, and declared nonmember scope; it is not assumed additive.

Admit \(\chi\) only when the capacity, interference, duration, prefix,
authority, epoch, pending-effect, and joint-support gates pass; every
participation threshold is met (or the contract explicitly returns a Pareto
frontier or \(\mathsf{OPEN}\)); and every withdrawal, revocation, or member
failure has a finite handoff, compensation, or recovery receipt that settles
or transfers each protected duty. Delayed commands are rejected at the
lease/epoch boundary. If a finite progress rank is absent, the recursive
status is \(\mathsf{OPEN}\).

For every finite admitted prefix, the coalition is distributedly executable
under the declared information, every protected duty and resource lease is
covered or explicitly \(\mathsf{OPEN}\), all externality entries are charged
to a declared owner or protected scope, and source-denied replay reproduces
the same successor, effect, utility interval, and responsibility map. After a
valid exit, old commands cannot create new effects and the new duty map is
replayable. The value interval is obtained from the same joint-kernel
recursion as Proposition~\ref{app:intervalbellman}; it does not imply a
Nash equilibrium, coalition stability, budget balance, fairness, global
optimality, or infinite-horizon liveness.
\end{proposition}

\begin{proof}
Induct over the finite event order. Local histories and arrived-message
queues make each selected member action available to that member; the joint
kernel then supplies one law for all effects, externalities, resources, and
receipts. The capacity and interference gates preserve shared resources, while
the externality ledger prevents a nonmember or shared duty from disappearing
from the accounting. Participation or Pareto checks preserve the declared
contract scope. At an exit or failure, the handoff, compensation, or recovery
receipt transfers the duty and reserve; the epoch fence rejects delayed old
commands. Finite horizon or rank closes the induction, and frozen messages,
kernel identity, and receipts give source-denied replay equality.
\end{proof}

Pairwise-compatible members can be jointly infeasible under a three-way XOR or
shared-capacity constraint. Adding individual utilities can miss a higher-order
externality, a local improvement can damage a protected nonmember, and a
withdrawal without a handoff can orphan a duty. These are reasons for a new
joint contract or \(\mathsf{OPEN}\), not evidence for a unique organization,
additive value law, or automatic coalition stability.

\begin{proposition}[Finite contract-compatible participation and explanation transfer]
\label{app:incentivetransfer}
Let \(\chi\) and \(\chi'\) be finite coalition contracts over a common declared
public scope, let \(H<\infty\) be the declared replacement-prefix horizon, and
let \(R\) be a member-replacement map with a finite duty
handoff, permission/epoch migration, and resource/recovery certificate. A
transfer witness contains: (i) the private-history and arrived-message
boundaries for both contracts; (ii) a participation interval or declared
Pareto scope for every retained or introduced member, with \(\eta_i\geq0\) and
lower bound \(u_i^-\geq o_i^+-\eta_i\) against a declared upper bound
\(o_i^+\) for the outside option, with the utility units, subject, horizon,
and scope fixed; (iii) bounds or compensation
receipts for every protected nonmember and shared-task externality, including
communication and resource effects; and (iv) a source-denied explanation
translator \(T_{\mathrm{exp}}\) carrying evidence roots, model and semantic
versions, preconditions, candidate action, effect receipt, authority, dissent,
rollback, resource charges, replay hash/snapshot, scheduler and random-state
receipts, and the replacement map. Let \(C\) translate the complete witnessed
state and responsibility fields. Replaced or revoked old principals remain
tombstoned; retained principals keep their current epoch and authority only
subject to the new admission gate. No private history or authority is
inherited by semantic equivalence, and any new
authority requires explicit re-consent. Require, for every legal complete
witnessed state \(x\) (public history, private histories, and arrived queues),
a nonempty joint-kernel set and, with \(Cx\) the translated state,
\[
d_{\mathrm{TV}}\!\left((\pi'_{\mathrm{pub}}\circ C)_\#
 W_{\chi}(\cdot\mid x),
 (\pi'_{\mathrm{pub}})_\#W_{\chi'}(\cdot\mid Cx)\right)\leq\varepsilon_W,
\qquad
d_{\mathrm{exp}}\!\left(T_{\mathrm{exp}}\zeta,\zeta'\right)
\leq\varepsilon_{\mathrm{exp}},
\]
where \(\pi'_{\mathrm{pub}}\) is the declared projection onto the public
successor, effect, resource, and receipt tuple, and the receipt space, both
distances, and their scopes are declared; \(\zeta'\) is the receipt actually produced by the
new contract's source-denied replay. Unknown or unpriced incentive,
communication, resource, or out-of-scope external effects return
\(\mathsf{OPEN}\). The participation or Pareto gate is rechecked at each
transition and is not inferred from the total-variation bound. An unresolved
dissent, missing re-consent, or incomplete recovery path also returns
\(\mathsf{OPEN}\). Historical execution/explanation transfer is valid only
when the original authorized intervention and effect receipt are present. A
causal claim additionally requires a new authorized intervention and effect
receipt under \(\chi'\), together with a declared intervention-law bound;
otherwise its status is \(\mathsf{OPEN}\). The displayed distances are
one-step bounds. For a finite prefix of length \(H\), the witness must also
declare prefix budgets \(\bar\varepsilon_W,\bar\varepsilon_{\mathrm{exp}}\):
under a stated coupling and compositional metric, they may be set to
\(\bar\varepsilon_W=\min\{1,H\varepsilon_W\}\) and
\(\bar\varepsilon_{\mathrm{exp}}=H\varepsilon_{\mathrm{exp}}\) (or its
normalized \(\min\{1,\cdot\}\) form when \(d_{\mathrm{exp}}\in[0,1]\)); without
that composition witness a direct bound on the complete prefix law is
required.

Then every finite admitted replacement prefix preserves the declared
participation or Pareto eligibility, externality bounds, duty coverage,
permission/epoch safety, and source-denied replay. Public successor and effect
laws and explanation records differ only within the declared prefix budgets
\((\bar\varepsilon_W,\bar\varepsilon_{\mathrm{exp}})\) (or the directly
declared complete-prefix bounds); old commands are rejected
after the migration epoch and the translated responsibility map is replayable.
The statement is a contract-transfer result, not a proof of truthful
reporting, Nash incentive compatibility, coalition stability, fairness,
budget balance, global optimality, utility preservation from the kernel bound,
or causal validity without the new authorized intervention and effect receipt.
\end{proposition}

\begin{proof}
Induct over the finite event prefix. The information-boundary condition makes
both policies distributedly executable on their respective histories, while
the per-state joint-kernel inequality controls the projected public
successor, effect, resource, and receipt tuple. The declared coupling and
composition budget then controls the finite-prefix law (or the witness uses
a direct complete-prefix bound). The participation and
externality witnesses preserve the declared eligibility and protected-scope
ledgers; an unknown entry fails admission. The migration certificate transfers
duties and recovery reserves, explicit re-consent installs any new authority,
and old tombstones plus the new epoch reject delayed old actions. The
translated evidence roots and \(T_{\mathrm{exp}}\) preserve the explanation
record and its permissions, and the source-denied receipts give replay
equality up to the declared metrics. A new authorized intervention receipt
and intervention-law bound are required for the causal subclaim; without
them the induction returns \(\mathsf{OPEN}\) for causality while retaining
only the historical execution explanation.
\end{proof}

The boundary is visible in small counterexamples: equal public outputs can
hide different private information, a replacement can improve member utility
while damaging an unpriced nonmember, an explanation can cite a revoked source,
and a text-only rationale can remain after its action effect changed. These
cases require a new witness or \(\mathsf{OPEN}\), not a claim of incentive
compatibility or explanation preservation.

\begin{proposition}[Finite repeated replacement and compositional revoke propagation]
\label{app:repeatedtransfer}
Let \(\chi^0,\ldots,\chi^N\), \(N<\infty\), be a finite sequence of
coalition contracts on a declared protected scope. For each stage \(k<N\),
let \(C_k\) be a measurable complete-state translation, \(T_k\) an
explanation translator, and \(\epsilon_{W,k},\epsilon_{\mathrm{exp},k},b_k\)
declared one-step bounds for public successor/effect/resource/receipt law,
explanation distance, and protected externality charge. Assume every stage
has the transfer witness of Proposition~\ref{app:incentivetransfer}, with
participation or Pareto gates rechecked, explicit re-consent for new
authority, and a finite duty/resource/recovery certificate. Require the
translations to compose on the witnessed state domain, retained principals to
preserve their current authority only under the next gate, and replaced or
revoked principals to remain tombstoned. A revoked command identifier may
never be reintroduced; its epoch is monotone and every later lease boundary
rejects it. Assume the stagewise couplings and explanation metrics compose,
and that every protected externality is charged by a declared cumulative
ledger; an unknown or unpriced cross-stage effect returns \(\mathsf{OPEN}\).

For every legal complete witnessed state \(x\) and stage \(k\), require
\[
d_{\mathrm{TV}}\!\left((\pi_{k+1}^{\mathrm{pub}}\circ C_k)_\#
W_k(\cdot\mid x),
(\pi_{k+1}^{\mathrm{pub}})_\#W_{k+1}(\cdot\mid C_kx)\right)
\leq\epsilon_{W,k},
\qquad
d_{\mathrm{exp}}\!\left(T_k\zeta_k,\zeta_{k+1}\right)
\leq\epsilon_{\mathrm{exp},k},
\]
where each \(\pi_{k+1}^{\mathrm{pub}}\) projects onto the declared public
successor/effect/resource/receipt tuple and \(\zeta_{k+1}\) is the receipt
produced by the next contract's source-denied replay. Then every finite
admitted sequence preserves
participation or Pareto eligibility, protected externality coverage with total
charge at most \(\sum_{k<N}b_k\), duty and resource coverage,
permission/epoch safety, and source-denied replay.
Under the declared coupling, the composed public law and explanation records
have bounds
\[
\bar\epsilon_W\leq\min\!\left\{1,\sum_{k<N}\epsilon_{W,k}\right\},
\qquad
\bar\epsilon_{\mathrm{exp}}\leq\sum_{k<N}\epsilon_{\mathrm{exp},k},
\]
with the normalized \(\min\{1,\cdot\}\) form for the second bound only when
the explanation metric is declared in \([0,1]\). Every command from a revoked
epoch is rejected after its lease boundary in all later stages. A historical
explanation composes only from receipts that remain source-denied replayable;
a causal claim at every stage additionally requires that stage's new
authorized intervention and effect receipt, otherwise the composed causal
status is \(\mathsf{OPEN}\). The result does not imply truthful reporting,
Nash compatibility, coalition stability, fairness, budget balance, utility
preservation outside the declared scope, global optimality, or infinite
liveness.
\end{proposition}

\begin{proof}
Induct on the stage index and, inside each stage, on its finite event prefix.
The transfer witness supplies local-information legality, participation and
externality gates, duty/resource coverage, and source-denied replay for the
next state. Composition of the complete-state maps preserves the witnessed
domain; explicit re-consent prevents authority from being copied merely by
semantic similarity. Monotone tombstones and epochs make the set of executable
old commands decrease, so no revoked command can reappear after a later
replacement. The cumulative externality ledger adds only declared protected
charges; an unknown entry rejects the sequence. A standard finite coupling
telescoping argument gives the sum of the stagewise total-variation bounds,
and the triangle inequality gives the additive explanation bound. Composed
replay translators preserve the receipt graph. If any intermediate causal
intervention/effect receipt is missing, the causal branch is OPEN even when
the historical execution explanation remains replayable.
\end{proof}

The finite boundaries are concrete: resetting an epoch can resurrect a
revoked command, pairwise small errors can exceed the declared prefix budget,
an unpriced cross-stage nonmember effect can invalidate every local utility
gate, and one missing intermediate intervention receipt breaks causal
composition. These cases require a new contract or \(\mathsf{OPEN}\), not an
unconditional repeated-coalition or explanation theorem.

\begin{proposition}[Finite grounded closure and owner-handoff composition]
\label{app:groundentity}
Let \(\Xi_0,\ldots,\Xi_N\), \(N<\infty\), be finite operational
contracts and let \(W_k\) be finite worlds with event \(e_k\) at stage \(k<N\).
For each stage, fix a declared premise order \(\preceq_k\) (its direction,
such as weaker versus stronger context, is contract data rather than an
inference), assume the admissible premise family
\(\mathsf{Prem}_{\Xi_k}(W_k,e_k)\) is nonempty and meet-closed in a finite
meet-semilattice, and define the least lawful common premise
\[
B_k^\star=\bigwedge\mathsf{Prem}_{\Xi_k}(W_k,e_k).
\]
Assume the expected and actual continuations are typed in the common response
space of \(B_k^\star\), so that the residual
\[
r_k=\operatorname{Def}_{\Xi_k,B_k^\star}(F_k,\widehat F_k)
\]
is a provenance-carrying, typed object. Assume the feasible grade order is
finite or well-founded with a declared unique least feasible grade. At that
grade, assume the lawful completion specifications form either a finite
meet-closed family or a declared complete meet-semilattice whose nonempty
meets preserve legality, owner duties, transport, and replay witnesses; let
\(a_k^\star\) be its least element. Finally, assume a typed contract morphism
\(\tau_k:\Xi_k\to\Xi_{k+1}\), a measurable conservative incumbent embedding
\(j_k:W_k\hookrightarrow W_{k+1}\), and owner/duty transport \(T_k\) over
\(\tau_k\) preserve incumbent legality, responses, updates, background
continuation, provenance, permissions, epochs, leases, resources, receipts,
and return/reopenability. Each selected completion carries a source-denied
realization receipt, and the transport maps compose over the contract maps.
If a new owner is created, require it to be independently addressable,
feedback-addressable, and responsibility-preserving, with its old duties
carried by the translated owner map. A missing common premise, lawful
completion, background-sufficiency witness, owner transport, constructible
owner witness, or recovery path returns \(\mathsf{OPEN}\).

Then every finite admitted stage has a unique least premise and a typed
residual; its least lawful completion is source-denied replayable and can be
composed with the next stage. The composite contract map
\(\tau_{N:0}=\tau_{N-1}\circ\cdots\circ\tau_0\) and transport
\(T_{N:0}=T_{N-1}\circ\cdots\circ T_0\) preserve the declared incumbent
future-operational laws, owner/duty continuity, resource and provenance
fields, and return/reopenability. A new owner is licensed only when the
incumbent architecture has no lawful same-grade completion that preserves the
registered future distinction and responsibility and a constructible owner
witness is present; when such a completion exists, the least closure does not
force spurious entityization. If every stage also has the stated universal
property, the composite result is unique up to future-operational isomorphism.
The proposition does not establish physical
representation categoricity, a unique dimension or gauge group, a theory of
everything, or infinite-horizon generation.
\end{proposition}

\begin{proof}
Meet-closure and finiteness give the unique least premise at each stage under
the declared order. The
common response space makes the defect operator well typed and its provenance
part of the replay receipt. The declared unique least feasible grade fixes one
grade; mere well-foundedness would allow incomparable minima.
The finite or complete meet-closed lawful-specification family, together with
the preservation clause, gives its least specification and replayable
realization witness. The
conservative embedding preserves the incumbent fields at one step, and
composition of the typed contract maps and measurable transports preserves
them over the finite sequence.
The owner criterion separates reuse from entityization: a same-grade lawful
completion blocks forced birth, while failure of every such completion plus
independent addressability, feedback, and responsibility requirements licenses
the new owner. Source-denied receipts and recovery witnesses compose by finite
induction. Any missing premise, completion, background, transport,
constructible owner, or recovery field breaks the induction and returns
\(\mathsf{OPEN}\).
\end{proof}

The finite counterexamples are structural. Two incomparable minimal grades do
not define one least grade without an extra order assumption; a state map that
preserves the current output but changes a hidden background continuation is
not conservative; an owner map that drops an old duty creates an orphan; and a
forced new owner despite an existing same-grade incumbent completion is
spurious birth. These cases require a new witness or \(\mathsf{OPEN}\), not a
claim of universal entityization or physical uniqueness.

\begin{proposition}[Finite lawful-completion selection under ambiguous minima]
\label{app:completioncategory}
Fix one admitted stage \((\Xi,W,e,r,\gamma)\), and let
\(\mathbf{Comp}_{\Xi,\gamma}(r)\) be the category whose objects are lawful
completion realizations at grade \(\gamma\), carrying the full future,
background, owner, duty, permission, epoch, resource, receipt, and recovery
fields. A morphism is a typed, measurable future-operational transport that
preserves those fields and composes with the declared contract morphisms.
If this category has an initial object \(I\), meaning that for every object
\(a\) there is exactly one structure-preserving morphism \(I\to a\), then
\(I\) is the canonical structural completion up to unique
future-operational isomorphism; it is an L-least completion only when the
morphism order is declared to refine the grade and full-lifecycle-cost order.
No meet-closure assumption is needed for this structural conclusion. If the
category instead has no initial object but its object set is finite and the
contract declares a total, source-denied replayable tie-break order
\(\triangleleft\) refining the full lifecycle cost and all protected duties,
then \(\min_{\triangleleft}\mathbf{Comp}_{\Xi,\gamma}(r)\) is a deterministic
contract-relative selector. It is a policy choice, not an intrinsic consequence
of \(L\), and the selector must be transported as part of the contract.

Alternatively, suppose lawful specifications embed in a complete lattice and
there is a declared monotone, extensive, idempotent closure operator
\(\Omega\) whose fixed points preserve the same typed fields. The least fixed
point above the residual seed, when it has a replayable realization witness,
provides the canonical specification; realization uniqueness still requires an
initial-object or equivalent unique-morphism witness. If only a nonempty set of
minimal candidates is known, with no initial object, closure-operator witness,
or total tie-break, the result is an antichain and the reducer returns
\(\mathsf{AMBIGUOUS}/\mathsf{OPEN}\). A finite sequence composes selected
objects only when their selectors or initial morphisms and contract transports
compose; otherwise it returns \(\mathsf{OPEN}\) at the first ambiguous stage.
\end{proposition}

\begin{proof}
Initial objects are unique up to unique isomorphism: the two initial objects
admit unique arrows in both directions, and uniqueness forces both composites
to be identities. A finite total order has one minimum, so a replayed tie-break
gives a deterministic selected object, but its identity depends on the declared
contract and cannot be promoted to an order-free theorem. Initiality alone is
structural; it yields the L-minimum only under the declared refinement of
morphisms by grade and full lifecycle cost. For a complete-lattice
closure operator, the fixed points above the seed have a least element under
the standard closure-operator construction; the realization clause prevents
confusing a specification with an executable world. Without one of these
canonical witnesses, two incomparable minimal objects can have different
future responses, owner duties, or recovery paths, so no single least closure
is justified. Composition follows by induction only when the chosen morphisms
and contract maps are typed and composable.
\end{proof}

The decisive counterexamples are a diamond of two non-isomorphic minimal
completions with no meet, a weak initial object with multiple outgoing
morphisms, a scalar-cost tie whose hidden duty differs, and a non-idempotent or
non-monotone \(\Omega\). These cases preserve the candidate antichain or return
\(\mathsf{OPEN}\); they do not license arbitrary owner birth or a claim of
physical uniqueness.

\begin{proposition}[Reality-conditioned least closure in a Scott sector]
\label{app:realityclosure}
Fix a contract \(\Xi\), a reality seed \(r\), and a complete lattice
\((\mathcal S_\Xi,\sqsubseteq,\bot)\) of typed specifications carrying
Kernel, relation, owner, duty, permission, epoch, resource, receipt, and
recovery fields. Let \(\mathcal R_r\) be a declared family of reality,
intervention, and continuation receipts, let \(E_r\) be a monotone exposure
operator, and let \(\Gamma_\Xi\) be an extensive, monotone, idempotent lawful
closure. Define
\[
 W_r(s)=\Gamma_\Xi\bigl(s\sqcup E_r(s)\bigr).
\]
Assume that \(W_r\) preserves directed joins, that
\(r\sqsubseteq W_r(r)\), and that every finite stage and the compatible
directed limit preserve owner completion, permissions, epochs, duties,
resources, recovery, and source-denied replay receipts. If any constructible
owner, permission, or recovery witness is unavailable, the reducer returns
\(\mathsf{OPEN}\). Assume also (i) lawful cofinality: every admitted reachable
future obligation has a finite-depth witness in some \(c\in\mathcal R_r\), and
(ii) future separation: distinct future-operational equivalence classes,
including owner, duty, and recovery fields, are separated by a declared probe
in \(\mathcal R_r\).

Then the Kleene chain \(s_{n+1}=W_r(s_n)\) satisfies
\[
 s_\omega=\bigsqcup_{n<\omega}W_r^n(r),
\]
where \(s_\omega\) is the least \(W_r\)-fixed specification above \(r\).
It is complete for the declared reality-probe scope. It is an executable world
only when a separate source-denied realization receipt supplies a reducer,
permissions, owner map, and replay path; without that receipt it is only a
semantic specification. Let \(\Phi_r\) be a monotone continued-service
operator on the corresponding viability carrier, with greatest fixed point
\(\nu\Phi_r\). A common operational world is certified only when the
field-preserving projection of \(s_\omega\) lies in \(\nu\Phi_r\) and the
realization receipt is valid. The least generative fixed point and the
greatest viability fixed point are therefore separate objects.
\end{proposition}

\begin{proof}
Monotonicity and \(r\sqsubseteq W_r(r)\) make the iterates an ascending chain.
Directed-join preservation gives
\(W_r(s_\omega)=\bigsqcup_n W_r(s_n)=s_\omega\). If \(t\) is any fixed point
above \(r\), induction gives \(s_n\sqsubseteq t\) for every \(n\), hence
\(s_\omega\sqsubseteq t\), proving leastness. Field preservation at successor
and limit stages, compatible receipts, and lawful cofinality supply every
declared finite-depth reality obligation. Future separation identifies the
claimed classes only on the declared probe quotient; it does not imply
physical categoricity. The viability conclusion is a separate membership
check in \(\nu\Phi_r\), so no \(\mu/\nu\) identification is used.
\end{proof}

The failure cases are a monotone but non-directed-continuous generator (the
Kleene limit need not be fixed), non-cofinal reality coverage, probes that do
not separate the claimed future classes, an owner completion that drops a duty
or epoch, an infinite chain with no executable colimit, and disjoint
generative and viability fixed points. Without a full-faithful unique-morphism
witness the result is an operational quotient statement rather than a physical
uniqueness theorem; without directed continuity one must use a transfinite
closure and cannot claim \(\omega\)-step convergence.

\begin{proposition}[Transfinite lawful closure without Scott continuity]
\label{app:transfiniteclosure}
Fix a contract \(\Xi\), a reality seed \(r\), and a set-sized complete lattice
\((\mathcal S_\Xi,\sqsubseteq,\bot)\) of typed specifications with cardinality
\(\kappa\). Let \(W_r:\mathcal S_\Xi\to\mathcal S_\Xi\) be monotone and
extensive above \(r\), and define the ordinal iteration
\[
 s_0=r,\qquad s_{\alpha+1}=W_r(s_\alpha),\qquad
 s_\lambda=\bigsqcup_{\beta<\lambda}s_\beta
 \quad(\lambda\text{ limit}).
\]
Assume that successor and limit joins preserve the registered owner, duty,
permission, epoch, resource, recovery, and reality-receipt fields, with
coherent source-denied transition receipts. A missing field witness or an
incompatible limit receipt returns \(\mathsf{OPEN}\). Then some
\(\alpha_\star<\kappa^+\) satisfies
\(s_{\alpha_\star+1}=s_{\alpha_\star}\), and this stabilized value is the
least \(W_r\)-fixed specification above \(r\). It is an executable world only
when a compatible source-denied executable colimit receipt exists at
\(\alpha_\star\); otherwise it is a semantic fixed point or \(\mathsf{OPEN}\).
If, in addition, \(W_r\) preserves \(\lambda\)-directed joins on a
\(\lambda\)-accessible sector, the same construction may use the corresponding
accessibility bound, but no \(\omega\)-step claim follows without the Scott
continuity assumption of Proposition~\ref{app:realityclosure}.
\end{proposition}

\begin{proof}
Extensivity and monotonicity make the ordinal chain ascending, and completeness
supplies every limit join. If no successor stage stabilizes before
\(\kappa^+\), every successor is strict and the chain contains
\(\kappa^+\) distinct elements of a lattice of cardinality \(\kappa\), a
contradiction. At the first stabilized stage, \(s_{\alpha_\star}\) is fixed.
For any fixed \(t\sqsupseteq r\), transfinite induction gives
\(s_\alpha\sqsubseteq t\): the successor case uses monotonicity and
\(W_r(t)=t\), and the limit case uses the defining supremum. Thus the
stabilized value is least. The field and receipt premises are needed to turn
the formal limit into an operational colimit; without them the lattice result
does not provide a legal reducer, owner map, resource budget, or replay path.
\end{proof}

The counterexamples are a nonmonotone generator (the chain need not ascend), a
proper-class or otherwise non-set-sized specification space, a limit supremum
that drops owner or duty legality, a formal fixed point with no compatible
executable colimit, and hidden reality obligations outside the cofinal probe
family. These cases require a new witness or \(\mathsf{OPEN}\), and do not
identify the transfinite least specification with an initial algebra, final
coalgebra, executable world, or physical categoricity.

\begin{proposition}[Accessible reflective completion with an executable witness]
\label{app:accessiblecompletion}
Let \(\mathbf C_\Xi\) be a locally \(\lambda\)-presentable category of typed
operational-world specifications and lawful morphisms, and let
\[
 \mathcal M_\Xi=\{m_i:A_i\to B_i\}_{i\in I}
\]
be a small set of completion obligations whose domains and codomains are
\(\lambda\)-presentable. Define the lawful closed subcategory
\(\mathbf{Closed}_\Xi=\mathcal M_\Xi^\perp\) by requiring every
precomposition map
\[
 m_i^\ast:\operatorname{Hom}(B_i,X)\longrightarrow
 \operatorname{Hom}(A_i,X)
\]
to be bijective. Assume explicitly that this orthogonality class is
reflective, with inclusion \(J:\mathbf{Closed}_\Xi\hookrightarrow\mathbf C_\Xi\)
and reflector \(\Omega_\Xi\dashv J\). Also assume that the unit and the
colimits used by the reflector carry source-denied receipts preserving owner,
duty, permission, epoch, resource, recovery, and replay fields; a missing
receipt returns \(\mathsf{OPEN}\).

Then every \(X\) has a unit \(\eta_X:X\to J\Omega_\Xi X\), and every lawful
map \(f:X\to JY\) into a closed \(Y\) factors through a unique
\(\bar f:\Omega_\Xi X\to Y\). The induced endofunctor
\(T_\Xi=J\Omega_\Xi\) is an idempotent semantic closure, and closed
completions are unique up to unique lawful isomorphism. The result is an
executable completion only when the stated receipts realize the unit,
colimit, owner, permission, and replay maps; otherwise it is a semantic
reflector. Local presentability and accessibility alone do not supply the
reflector or executable realization.
\end{proposition}

\begin{proof}
The factorization and its uniqueness are the defining universal property of
the assumed adjunction. Since \(J\) is a full faithful inclusion, the
adjunction counit is an isomorphism, so \(J\Omega_\Xi J\Omega_\Xi\) is
canonically isomorphic to \(J\Omega_\Xi\); hence \(T_\Xi\) is idempotent.
The receipt premises transport the categorical unit and colimit through the
typed lifecycle fields and make the factorization replayable. They do not
follow from the categorical adjunction itself, which is why a missing witness
leaves the status \(\mathsf{OPEN}\).
\end{proof}

An accessible endofunctor by itself is insufficient: on \(\mathbf{Set}\),
\(F(X)=X+1\) is accessible but is not an idempotent reflector. Requiring only
surjective fillers makes \(m_i^\ast\) an existence map but permits multiple
non-isomorphic completions, so it yields at most a weak reflection or an
injective completion. A non-reflective orthogonality class, an owner or replay
colimit that cannot be constructed, and a hidden obligation outside the small
family are explicit \(\mathsf{OPEN}\) cases; none licenses physical
categoricity or an identification with \(\mu F\) or \(\nu F\).

\begin{proposition}[Executable realization of a semantic colimit]
\label{app:executablecolimit}
Let \(D:J\to\mathbf C_\Xi\) be a finite or declared \(\lambda\)-directed
diagram of typed operational specifications, with field-preserving lawful
morphisms and a semantic colimit \((L,\iota_j)_{j\in J}\). Assume that every
object and structure map carries a source-denied receipt preserving owner,
duty, permission, epoch, resource, recovery, and replay fields. Suppose there
is an executable realization \(R(L)\), maps \(R(\iota_j)\), and a reducer
\(\mathsf{Red}_L\) satisfying all of the following conditions:
\begin{enumerate}
\item for every arrow \(u:j\to k\), the successor/effect/owner/permission/epoch/
resource/recovery/receipt projection of \(R(\iota_k\circ D(u))\) agrees with
that of \(R(\iota_j)\) (or has a declared additive error budget);
\item every finite prefix has a unique lawful reducer splice through \(L\),
with no owner or duty conflict, monotone epoch/tombstone state, and a resource
ledger that neither double-spends nor exceeds its declared budget;
\item source-denied replay of every finite prefix factors through \(L\) and
reproduces the same successor, effect, charge, and receipt (within the declared
error budget); and
\item every lawful universal mediator out of the semantic colimit preserves
these fields and the declared recovery map.
\end{enumerate}
Then the semantic colimit has an executable, replay-faithful realization under
the declared scope. If any premise is missing, the colimit remains semantic
only and the status is \(\mathsf{OPEN}\).
\end{proposition}

\begin{proof}
Condition (1) makes the executable cocone agree on each structure map. By
(2), finite prefixes can be glued in a unique lawful order without dropping a
duty, resetting an epoch, or reusing a resource token. Induction over the
prefix, using (3), gives source-denied replay with the same joint successor,
effect, cost, and receipt projection. Condition (4) transports this property
through any lawful mediator, so the executable realization respects the
universal colimit factorization on the declared finite prefixes. The semantic
colimit alone supplies none of these runtime, ledger, or replay facts.
\end{proof}

The boundary is operationally substantive. A set-colimit may identify nodes
with incompatible owners or epochs and have no lawful owner map; two branches
may each spend the same resource token even though their formal union exists; a
shared event may carry incompatible random-state receipts and therefore admit
no deterministic source-denied replay; and coequalizing a stale command with a
new command may erase a tombstone and re-enable a revoked effect. These cases
leave only a semantic colimit or \(\mathsf{OPEN}\); an abstract adjunction does
not certify an executable reducer.

\begin{proposition}[Budgeted executable-colimit termination]
\label{app:colimittermination}
Fix a declared finite prefix scope for the executable realization \(R(L)\),
and let \(E_L\) be its typed reducer states. Suppose every nonterminal state
has a deterministic or source-denied replayable reducer step, with terminal
status \(\mathsf{READY}\) or \(\mathsf{OPEN}\). Let
\[
 r:E_L\longrightarrow (W,\prec)
\]
be a ranking into a well-founded ordered set, and assume that every admitted
nonterminal step strictly decreases \(r\). Suppose also that each step has a
declared full-lifecycle upper cost \(c^+\) (discovery, alignment, compilation,
execution, verification, maintenance, recovery, and communication), finite internal work, and a receipt preserving
owner, duty, permission, epoch, resource, recovery, and replay fields. If the
initial rank is \(r_0\), the prefix resource ledger is closed under the
declared composition law, and the total charged upper cost is at most the
declared budget \(B\), then every admitted reducer run terminates after
finitely many steps, with at most the well-founded rank height below \(r_0\)
steps. The terminal receipt is source-denied replayable and retains the field
invariants. If the ranking, a finite cost upper bound, finite internal work,
resource closure, or a replay-preserving terminal reducer is missing, the
prefix is not certified and returns \(\mathsf{OPEN}\).
\end{proposition}

\begin{proof}
Well-foundedness rules out an infinite strictly descending sequence, so every
admitted run reaches a terminal state within the stated rank height. Summing
the declared upper costs over this finite prefix gives the budget inequality;
the ledger-closure premise prevents hidden double-spend or peak-resource
overrun during prefix composition. Receipt preservation and deterministic or
recorded-random replay give the same terminal state, effect, charge, and field
projection when the prefix is replayed source-denied. The conclusion is finite
and contract-relative; it does not imply fairness, open-world liveness, or a
ranking for arbitrary semantic colimits.
\end{proof}

Zero-cost stuttering, a reducer cycle, an unbounded hidden microstep loop, an
underestimated internal cost, or a replay that changes the ranking are explicit
counterexamples. A finite diagram or an abstract reflector does not exclude
these cases without the ranking, resource, and cost witnesses.

\begin{proposition}[Contract-change transport for cost and termination]
\label{app:contracttransport}
Let \(\Xi^-\) and \(\Xi^+\) be two declared operational contracts, with
executable prefix states \(E^-\) and \(E^+\). Suppose a migration map
\(M:E^-\to E^+\) and a source-denied migration receipt preserve every protected
owner, duty, permission, epoch/tombstone, resource, recovery, and replay field,
or explicitly settle it before the change. Assume that for every admitted old
prefix \(u\), the new reducer has a transported prefix \(M u\) and a
full-lifecycle cost bound
\[
 \mathsf{Cost}_{\Xi^+}(M u)
 \leq \mathsf{Cost}_{\Xi^-}(u)+\kappa_M,
\]
where \(\kappa_M\) includes migration, validation, communication, and
recovery reserve. Assume further that a well-founded rank is transported by a
declared monotone map \(h\) with
\(r^+(M x)\preceq h(r^-(x))\), and that the new prefix ledger is closed under
the same resource-composition law. If replay commutes with migration on every
finite prefix, then an old certified prefix yields a new certified prefix
whenever the transported cost plus \(\kappa_M\) fits the new budget and the
new reducer reaches \(\mathsf{READY}\) or \(\mathsf{OPEN}\) within the
transported rank bound. Direct comparison of raw costs or ranks across
\(\Xi^-\) and \(\Xi^+\) is invalid without these witnesses.
\end{proposition}

\begin{proof}
The field-preserving migration receipt carries the old responsibility and
recovery obligations into the new contract, while settled fields cannot be
silently dropped. The cost inequality charges the complete migration and
recovery overhead, and ledger closure prevents a locally valid prefix from
double-spending a shared resource after transport. The rank inequality and
well-foundedness give the finite termination bound for the new reducer. Replay
commutation gives the same successor, effect, charge, and receipt projection
whether the old prefix is replayed before migration or the transported prefix
is replayed directly. If any witness is missing, the comparison returns
\(\mathsf{OPEN}\), not a growth or non-regression certificate.
\end{proof}

Changing units or hidden maintenance obligations, resetting the rank to an
unbounded or incomparable value, dropping an in-flight duty, reusing an old
resource budget, or changing action semantics while preserving only the output
are counterexamples. A lower reported runtime cost after omitting formation,
verification, communication, recovery, or migration work is not a valid
cross-contract improvement.

\begin{proposition}[Finite-chain contract transport]
\label{app:contractchain}
Let
\[
 \Xi^0\xrightarrow{\tau_0}\Xi^1\xrightarrow{\tau_1}\cdots
 \xrightarrow{\tau_{N-1}}\Xi^N
\]
be a finite chain of declared operational contracts. For each stage let
\(C_k:E^k\to E^{k+1}\) be a source-denied, field-preserving migration,
\(R_k\) its resource-unit conversion, \(c_k\) the admitted execution and
validation charge, \(r_k\) the recovery reserve, and \(h_k\) the remaining
formation, installation, maintenance, communication, and migration overhead.
Assume the conversions compose (or carry an explicitly summed error budget),
\(R_{k+1}R_k=R_{k:k+2}\), each \(R_k\) is monotone for the declared
prefix-resource composition law, and the stage ledger is updated without
overdraw by
\[
 B_{k+1}=R_k\bigl(B_k-c_k-r_k\bigr)-h_k
\]
or by an equivalent closed cost-monoid rule. Assume every admitted prefix
satisfies \(c_k+r_k+h_k\preceq B_k\), and that a well-founded rank is
transported by maps \(\phi_k\) satisfying
\[
 v_{k+1}(C_kx)\preceq \phi_k\bigl(v_k(x),a_k\bigr),
\]
where \(\phi_k\) is monotone and every adaptation charge \(a_k\) is included
in \(h_k\). A rank reset is permitted only as an explicitly declared new
obligation or epoch whose lexicographic stage/rank measure still strictly
decreases. Finally, require source-denied replay commutation on every finite
prefix: replay after \(C_k\) equals translated replay before \(C_k\) on the
successor, effect, owner, permission, epoch/tombstone, resource, recovery,
and receipt fields, up to the declared error budget, with unique resource
tokens and tombstones never reused.

Then every finite chain prefix satisfying the ledger and rank bounds has
composed field-preserving migration, payable full-lifecycle cost, no resource
double-spend, a finite terminal reducer bound, and replay-equivalent protected
successors under \(\Xi^0\) and \(\Xi^N\). If composition, ledger closure,
rank transport, explicit reset accounting, or replay commutation is missing,
the chain is `OPEN`; locally certified stages do not compose automatically.
\end{proposition}

\begin{proof}
Induct on the chain length. Resource-unit composition and monotonicity turn
the stage inequality into a closed prefix ledger, so a token or recovery
reserve cannot be charged twice. The transported rank and the explicit
lexicographic reset rule give a well-founded global measure; the finite-stage
termination proposition therefore applies at each stage and composes the
terminal bound. The replay square commutes at one stage by assumption, and
composition of the migration maps and their field projections gives the same
successor, effect, charge, owner/epoch state, and receipt after any finite
number of stages. Failure of any square or ledger equation leaves the result
semantic-only or `OPEN`, rather than certifying growth.
\end{proof}

\paragraph{Finite-chain counterexamples.}
Unit conversions \(1\mapsto\tfrac12\mapsto2\) can look locally valid while
failing to compose; a stage ledger can omit earlier maintenance or recovery
and then re-number the same resource token, creating a double spend; resetting
the rank to a fresh height at every migration can make every stage locally
READY while an infinite chain never descends; progressively widening scope or
deadlines can fake capability growth; and changing a random-state or receipt
schema can make every local replay deterministic while the composed replay
disagrees. These cases return `OPEN`.

\begin{proposition}[Finite mixed exact/approximate contract-chain transport]
\label{app:mixedchain}
Let
\[
 \Xi^0\xrightarrow{\tau_0}\Xi^1\xrightarrow{\tau_1}\cdots
 \xrightarrow{\tau_{N-1}}\Xi^N
\]
be a finite chain of operational contracts with complete-state migrations
\(C_k:E^k\to E^{k+1}\). At every stage, augment the state with the finite
delay buffer and authority/permission fields so that the translated process is
Markov, and let \(d_k\) be a common typed probability metric on the translated
successor, effect, feedback, resource, recovery, and replay fields. Require
owner, duty, permission, epoch/tombstone, and protected-scope fields to be
preserved exactly; approximation is allowed only on the declared numerical
fields. Suppose the stage coupling and replay square satisfy
\[
 e_{k+1}\leq\lambda_k e_k+\eta_k,
 \qquad
 \eta_k=\epsilon_k^{\rm end}+L_k\ell_k+\xi_k+\epsilon_k^{\rm coup},
\]
where \(\ell_k\) is the delay bound, \(\xi_k\) the jitter bound,
\(\epsilon_k^{\rm end}\) the endpoint error, and \(\epsilon_k^{\rm coup}\)
the coupling/rollback error. An exact stage has \(\eta_k=0\). Assume the
stage maps compose on the declared scope, each \(\lambda_k\) is finite, and
the full-lifecycle vector-resource ledger charges migration, coupling,
buffering, validation, communication, rollback, recovery, and execution with
no shared-peak overrun. Finally, assume a well-founded rank is transported
with adaptation cost included in the ledger and strictly decreases on each
admitted nonterminal reducer step.

Then the composed discrepancy obeys
\[
 e_N\leq\left(\prod_{j=0}^{N-1}\lambda_j\right)e_0
 +\sum_{k=0}^{N-1}\left(\prod_{j=k+1}^{N-1}\lambda_j\right)\eta_k .
\]
If this value is at most the declared finite prefix budget and every protected
field remains exact, the chain has a certified mixed transport with finite
rank-bounded stopping and source-denied replay up to the declared numerical
error. Without a common metric/coupling, Markov delay augmentation,
scope/authority/epoch compatibility, closed vector ledger, or rank witness,
the result is `OPEN`.
\end{proposition}

\begin{proof}
The one-stage inequality follows from the declared coupling, endpoint bounds,
and the time-Lipschitz delay/jitter estimate. Substitution and induction on
the finite stage index give the displayed weighted sum. Exact preservation of
protected fields prevents an approximate numerical discrepancy from silently
dropping a duty or reviving a tombstoned command. The closed vector ledger
charges every buffer, coupling, validation, rollback, and shared peak before
the next stage, while the transported well-founded rank gives finite reducer
termination. Source-denied replay commutes stage by stage; composing the maps
gives the same protected projection and a numerical discrepancy bounded by
\(e_N\). Any missing premise leaves the chain uncertified and `OPEN`.
\end{proof}

\paragraph{Mixed-chain counterexamples.}
Each local \(\epsilon_k\) may fit its local budget while the weighted sum
exceeds the prefix budget; a dropped delay buffer can hide an in-flight duty;
shared-capacity batches can be locally legal while their composed peak is
unpayable; an approximate authority or epoch field can admit a stale effect;
and \(\lambda_k>1\) can amplify a small timing defect. These failures are
not repaired by marginal output agreement and return `OPEN`.

\begin{proposition}[Finite stage-varying metric and coupling transport]
\label{app:metrictransport}
Let
\[
 \Xi^0\xrightarrow{\tau_0}\Xi^1\xrightarrow{\tau_1}\cdots
 \xrightarrow{\tau_{N-1}}\Xi^N
\]
be a finite mixed exact/approximate contract chain as in
Proposition~\ref{app:mixedchain}, but allow stage $k$ to use its own typed
probability metric $d_k$ and coupling $K_k$. Let $J_k$ be an explicit
conversion and calibration map from the translated state at stage $k$ to the
translated state at stage $k+1$.
The map carries versioned frame, unit, clock, and calibration identifiers;
after applying $J_k$, the stage-$k$ reducer and coupling are measured in
the post-conversion metric $d_{k+1}$. Assume that, on the declared numerical
fields, $J_k$ has the witnessed distortion
\[
 d_{k+1}(J_k x,J_k y)\leq \alpha_k d_k(x,y)+\zeta_k,
 \qquad 0\leq\alpha_k<\infty,
\]
where $\zeta_k$ includes calibration, quantization, and unit-conversion
error. Require a common protected-field projection $p_k$ and exact
compatibility
\[
 p_{k+1}J_k=p_k
\]
on the protected scope, together with coupling conversion witnesses whose
pushforwards agree with $J_k$ up to a declared numerical discrepancy
$\chi_k$. Owner, duty, permission, epoch/tombstone, deadline, and protected
scope fields are exact and are never measured only through $d_k$.

Suppose the stage-$k$ dynamics satisfy the local mixed-chain inequality with
factor $\lambda_k$ and intrinsic error $\eta_k$, and let the conversion and
coupling costs be charged in the full-lifecycle vector ledger. Define
\[
 a_k:=\lambda_k\alpha_k,
 \qquad
 b_k:=\lambda_k\zeta_k+\eta_k+\chi_k.
\]
Assume all $a_k,b_k$ are finite, the converted maps compose on the declared
scope, the ledger has no shared-peak or recovery overrun, and a transported
well-founded rank strictly decreases on every admitted nonterminal reducer
step (including recalibration work). Then
\[
 e_N\leq\left(\prod_{j=0}^{N-1}a_j\right)e_0
 +\sum_{k=0}^{N-1}\left(\prod_{j=k+1}^{N-1}a_j\right)b_k .
\]
If this bound is within the declared finite prefix error budget and all
protected projections commute, the chain is certified for the switched
metrics/couplings, with source-denied replay up to the displayed numerical
error and finite rank-bounded stopping. In the uniform switched-contraction
sector $a_k\leq\rho<1$ and $b_k\leq\bar b$, the bound is at most
$\rho^N e_0+\bar b(1-\rho^N)/(1-\rho)$. Missing conversion maps, coupling
pushforward witnesses, protected-field compatibility, charged calibration
cost, or rank transport leaves the result $\mathsf{OPEN}$.
\end{proposition}

\begin{proof}
Apply the conversion distortion to the incoming discrepancy, then apply the
stage-$k$ coupling inequality and add the intrinsic and coupling-conversion
errors. This gives $e_{k+1}\leq a_ke_k+b_k$. Induction on the finite chain
index yields the displayed switched product-sum bound. Exact commutation of
$p_k$ prevents numerical conversion from dropping a duty, changing an epoch,
or reviving a tombstoned effect. The vector ledger pays for conversion,
calibration, buffering, communication, rollback, recovery, and execution
before the next stage, while the transported rank supplies finite stopping.
Stagewise source-denied replay commutes with $J_k$ and the coupling witnesses;
composition therefore preserves the protected projection and the stated
numerical tube. If any square, ledger, or rank premise is absent, no switched
certificate follows.
\end{proof}

\paragraph{Stage-switch counterexamples.}
Two stages can each pass their local metric budget while a conversion with
large $\alpha_k$ makes the composed product exceed the prefix budget. If a
calibration map changes units without the $\zeta_k$ charge, the same physical
error can be counted as arbitrarily small or large. A local contraction can
still yield a globally expanding switched chain when the product of the
$\lambda_k\alpha_k$ exceeds one. A coupling that matches numerical outputs
but does not preserve $p_k$ can silently drop a duty or admit a stale epoch;
recalibration or communication peaks omitted from the ledger can then make an
otherwise legal chain unaffordable. All such cases return $\mathsf{OPEN}$.
\begin{proposition}[Finite nonuniform vector switched transport]
\label{app:switchedvector}
Let
\[
 \Xi^0\xrightarrow{\tau_0}\Xi^1\xrightarrow{\tau_1}\cdots
 \xrightarrow{\tau_{N-1}}\Xi^N
\]
be a finite stage-varying chain as in Proposition~\ref{app:metrictransport}.
After each conversion $J_k$, represent the numerical discrepancy in a common
finite cone $\mathbb R_+^m$ with componentwise order. Suppose the complete
translated state contains the delay/jitter buffer, resources, in-flight
messages and effects, owner/duty/permission/epoch/tombstone/scope fields, and
replay receipt, and that the protected projection is exact. Assume that the
stage-$k$ comparison satisfies
\[
 e_{k+1}\preceq A_k e_k+b_k,
 \qquad A_k\in\mathbb R_+^{m\times m},\quad b_k\in\mathbb R_+^m,
\]
where $b_k$ includes conversion, calibration, delay, jitter, coupling,
rollback, and endpoint discrepancies. Then the finite chain has the explicit
cone bound
\[
 e_N\preceq (A_{N-1}\cdots A_0)e_0
 +\sum_{k=0}^{N-1}(A_{N-1}\cdots A_{k+1})b_k,
\]
with an empty product equal to the identity.

If there is a common weight $w\gg0$ and $0\leq\rho<1$ such that
\[
 A_kw\preceq\rho w,
 \qquad \|b_k\|_w\leq\bar b,
 \qquad \|x\|_w:=\max_i |x_i|/w_i,
\]
then
\[
 \|e_N\|_w\leq\rho^N\|e_0\|_w
 +\bar b\frac{1-\rho^N}{1-\rho}.
\]
Without a common weight, a finite prefix may still be certified when the
displayed matrix products and the vector ledger fit the declared budget, but
no arbitrary-switching or uniform stability claim follows.

For each stage, let $C_k$ be the current encoding, $U_{e,k}$ the detailed
update, and $\widehat U_{e,k}$ the translated update on the next encoded
state. Require the update transport square
\[
 C_{k+1}U_{e,k}=\widehat U_{e,k}C_k
\]
exactly, or with a typed discrepancy already included in $b_k$. Require the
vector ledger to charge conversion, recalibration, buffering, communication,
rollback, recovery, execution, and shared peaks; require replay, owner/epoch
fences, protected projections, and a well-founded rank to commute or decrease.
Under these witnesses the finite bound and, when the common-weight condition
holds, the switched stability certificate are sound. Missing update transport,
common-coordinate cone, protected-field exactness, ledger closure, replay, or
rank evidence leaves the corresponding continuation $\mathsf{OPEN}$.
\end{proposition}

\begin{proof}
Monotonicity of multiplication by a nonnegative matrix gives the one-step cone
recurrence. Induction on $k$ yields the matrix product-sum expression. The
common-weight inequalities imply $\|A_kx\|_w\leq\rho\|x\|_w$ for every
nonnegative error vector, so the geometric bound follows. The update square
makes the encoded successor independent of the chosen detailed representative;
its approximate version contributes to $b_k$. Exact protected projections,
replay commutation, and epoch fences prevent numerical error from changing
responsibility, reviving stale effects, or dropping in-flight duties. The
closed vector ledger and descending rank provide finite cost and stopping.
\end{proof}

\paragraph{Vector-switch counterexamples.}
The matrices
\[
 A_1=\begin{pmatrix}.5&2\\0&.5\end{pmatrix},
 \qquad
 A_2=\begin{pmatrix}.5&0\\2&.5\end{pmatrix}
\]
have spectral radius $0.5$ separately, but
\(A_1A_2=\begin{pmatrix}4.25&1\\1&.25\end{pmatrix}\) has largest eigenvalue
approximately $4.486$; local stability therefore does not imply arbitrary
switching stability without a common weight or an explicit product bound. A
second counterexample has $C s_1=C s_2$ but
\(C' U_e s_1\ne C' U_e s_2\); no updater that sees only $Cs$ can satisfy the
transport square. Numerical output agreement can therefore coexist with a
non-transportable future update, and these cases return $\mathsf{OPEN}$.

\begin{proposition}[Finite product transport across contract changes]
\label{app:producttransport}
Let
\[
 \Xi^0\xrightarrow{\tau_0}\Xi^1\xrightarrow{\tau_1}\cdots
 \xrightarrow{\tau_{N-1}}\Xi^N
\]
be a finite contract chain. At stage $k$, let the numerical error lie in the
finite cone $E_k=\mathbb R_+^{m_k}$, let $A_k:E_k\to E_k$ be a nonnegative
local error map, let $M_k:E_k\to E_{k+1}$ be a nonnegative migration or
conversion map, and let $b_k\in E_k$ and $r_k\in E_{k+1}$ be the intrinsic and
typed transport discrepancies. The complete translated state contains the
delay/jitter buffer, resources, in-flight messages and effects,
owner/duty/permission/epoch/tombstone/scope fields, and a replay receipt.
Assume the protected projection is exact,
\[
 p_{k+1}M_k=p_k.
\]
Define
\[
 B_k:=M_kA_k,\qquad c_k:=M_kb_k+r_k,\qquad
 P_{j:i}:=B_{j-1}\cdots B_i\quad(j>i),\qquad P_{i:i}:=I.
\]
Suppose the translated recurrence is
\[
 e_{k+1}\preceq B_ke_k+c_k .
\]
Then every finite prefix $i<j\leq N$ has the explicit bound
\[
 e_j\preceq P_{j:i}e_i+
 \sum_{\ell=i}^{j-1}P_{j:\ell+1}c_\ell .
\]

For positive stage weights $w_k\gg0$, write
\[
 \|x\|_{w_k}:=\max_a |x_a|/(w_k)_a .
\]
If the transported norm obeys
\[
 \|B_kx\|_{w_{k+1}}\leq\rho_k\|x\|_{w_k},\qquad
 \rho_k\leq\rho<1,\qquad
 \|c_k\|_{w_{k+1}}\leq\bar c ,
\]
then
\[
 \|e_j\|_{w_j}\leq
 \rho^{\,j-i}\|e_i\|_{w_i}
 +\bar c\,\frac{1-\rho^{\,j-i}}{1-\rho}.
\]
Without this common transported contraction, declared finite product bounds
\(\|P_{j:i}\|\leq\Gamma_{j,i}\) and a closed vector ledger certify only the
declared finite prefix; they do not imply a joint-spectral-radius,
arbitrary-switching, or infinite-horizon claim.

An optional finite-block certificate is valid for an aligned schedule when
\[
 \|B_{k+r-1}\cdots B_k\|\leq K^r\quad(1\leq r\leq q),
 \qquad
 \|B_{k+q-1}\cdots B_k\|\leq\beta<1,
\]
where all norms are the declared compatible operator norms between the stage
spaces, and $\|c_k\|\leq\bar c$. With
\(C_q:=\bar c\sum_{r=0}^{q-1}K^r\), the block times satisfy
\[
 \|e_{i+Jq}\|\leq
 \beta^J\|e_i\|+C_q\frac{1-\beta^J}{1-\beta},
\]
and a remainder $0\leq s<q$ obeys
\[
 \|e_{i+Jq+s}\|\leq
 K^s\|e_{i+Jq}\|+\bar c\sum_{r=0}^{s-1}K^r .
\]
This is a finite block/dwell-time witness, not a universal switching theorem.

Let $C_k$ be the current encoding, $U_{e,k}$ the detailed update, and
\(\widehat U_{e,k}\) the translated update. Require
\[
 C_{k+1}U_{e,k}
 =\widehat U_{e,k}C_k+\Delta^U_k ,
\]
with $\Delta^U_k$ typed in $E_{k+1}$ and bounded by $r_k$ in the declared
stage norm. Migration, recalibration, buffering, communication, rollback,
recovery, execution, and shared peaks are charged in the ledger; replay,
owner/epoch fences, protected projections, and a well-founded rank commute or
decrease. Missing product typing, update discrepancy, protected-field
transport, ledger closure, replay, or rank evidence leaves the continuation
\(\mathsf{OPEN}\).
\end{proposition}

\begin{proof}
Nonnegative matrix monotonicity and induction on the finite stage index give the
product-sum expression. The transported weighted inequalities give the
geometric bound. Applying the $q$-step inequality at aligned block boundaries
and bounding each within-block input by \(K^r\) gives the block and remainder
bounds. The typed update discrepancy contributes to \(r_k\), so the encoded
successor is independent of the detailed representative only under the
declared discrepancy bound. Exact protected projections, replay commutation,
epoch fences, the closed lifecycle ledger, and the descending rank preserve
responsibility and finite stopping.
\end{proof}

\paragraph{Product-transport counterexamples.}
The stable matrices in the preceding vector-switch counterexample have an
expanding product, so finite observations of separate modes do not justify a
joint-spectral-radius or arbitrary-switching claim. A migration
\(M=\operatorname{diag}(K,K^{-1})\) with an uncharged unit conversion changes a
weighted norm and product bound by an arbitrary factor; if \(p'M\ne p\), it also
changes the protected target. Finally, \(Cs_1=Cs_2\) can hold while
\(C'U_es_1\ne C'U_es_2\), so current-output equality does not preserve future
update sufficiency. These cases return \(\mathsf{OPEN}\).

\begin{proposition}[Finite concatenation of block transport certificates]
\label{app:blockcompose}
Let a finite product-transport chain be partitioned at
\[
 0=t_0<t_1<\cdots<t_L=N.
\]
For segment \(\ell\), let
\[
 Q_\ell:=P_{t_{\ell+1}:t_\ell},\qquad
 g_\ell:=\sum_{k=t_\ell}^{t_{\ell+1}-1}
 P_{t_{\ell+1}:k+1}c_k .
\]
Assume that every segment certificate is expressed on the same translated
complete state at its two boundaries, with a declared norm
\(\lVert\cdot\rVert_\ell\) on \(E_{t_\ell}\). Require boundary-compatible
unit, clock, and calibration maps, so that the protected projection,
owner/duty/permission/epoch/tombstone/scope fields, replay receipt, and
resource units have one image at each shared boundary. Suppose the local
finite or \(q_\ell\)-step certificate, including any explicitly re-phased
partial block, yields
\[
 \lVert Q_\ell x\rVert_{\ell+1}\leq\beta_\ell\lVert x\rVert_\ell,
 \qquad
 \lVert g_\ell\rVert_{\ell+1}\leq\gamma_\ell .
\]
Then the concatenated boundaries satisfy
\[
 \lVert e_{t_L}\rVert_L\leq
 \left(\prod_{\ell=0}^{L-1}\beta_\ell\right)\lVert e_{t_0}\rVert_0
 +\sum_{\ell=0}^{L-1}
 \left(\prod_{h=\ell+1}^{L-1}\beta_h\right)\gamma_\ell .
\]
For any partial segment prefix with an intermediate bound
\(\lVert P_{k:t_\ell}x\rVert\leq K_{\ell,k}\lVert x\rVert_\ell\)
and input term bounded by \(\delta_{\ell,k}\), the same induction gives
\[
 \lVert e_k\rVert\leq
 K_{\ell,k}\lVert e_{t_\ell}\rVert_\ell+\delta_{\ell,k}.
\]
The certificate is finite and contract-relative. It does not infer a common
norm, a joint-spectral-radius bound, or an infinite-horizon result when the
segment boundaries, block phases, or unit maps are changed.

The ledger must add each segment's conversion, re-phasing, calibration,
buffering, communication, rollback, recovery, execution, maintenance, and
shared-peak charges exactly once, and the update transport discrepancies must
be included in the corresponding \(g_\ell\). Protected projections, replay,
epoch fences, and a well-founded rank must compose at every boundary; failure
of any one of these conditions returns \(\mathsf{OPEN}\).
\end{proposition}

\begin{proof}
The recurrence in Proposition~\ref{app:producttransport} gives
\(e_{t_{\ell+1}}\preceq Q_\ell e_{t_\ell}+g_\ell\). Apply the declared
boundary norm and its two inequalities, then induct on \(\ell\). The partial
prefix statement is the same argument before the next boundary. Boundary
compatibility is what permits the output norm and receipt of one segment to be
the input norm and receipt of the next; without it the scalar quantities being
multiplied are not measurements of the same translated state. Ledger and
rank composition are independent obligations and are preserved only when
their witnesses are transported rather than reset.
\end{proof}

\paragraph{Boundary-reset counterexample.}
Take scalar spaces \(E_0=E_1=E_2=\mathbb R_+\) and maps
\(Q_0x=90x\), \(Q_1x=0.9x\). The first segment reports operator norm
\(0.9\) using weights \(w_0=1,w_1=100\); the second reports operator norm
\(0.9\) using a silently reset input weight \(\widetilde w_1=1\) and output
weight \(w_2=1\). The actual composition is \(Q_1Q_0x=81x\), so multiplying
the two local certificates is unsound. A boundary conversion or a common
stage weight would expose the mismatch; absent one, the status is
\(\mathsf{OPEN}\).

\begin{proposition}[Finite heterogeneous block transport with boundary factors]
\label{app:heteroq}
Consider two finite product-transport subchains, indexed by
\(s\in\{a,b\}\), with lengths
\[
 n_s=J_sq_s+r_s,\qquad 0\leq r_s<q_s .
\]
In the declared norm of segment \(s\), suppose every internal product of
length \(u\leq q_s\) is bounded by \(K_s^u\), every aligned \(q_s\)-block
is bounded by \(\beta_s<1\), and \(\|c^s_k\|\leq\bar c_s\). Define
\[
 C_s:=\bar c_s\sum_{u=0}^{q_s-1}K_s^u,
 \qquad
 R_s:=\bar c_s\sum_{u=0}^{r_s-1}K_s^u .
\]
If \(E_{s,0}\) and \(E_{s,\mathrm{end}}\) are the input and output error
norms of segment \(s\), Proposition~\ref{app:producttransport} gives
\[
 E_{s,\mathrm{end}}\leq
 K_s^{r_s}\left[\beta_s^{J_s}E_{s,0}+
 C_s\frac{1-\beta_s^{J_s}}{1-\beta_s}\right]+R_s .
\]

Join the segments with a typed boundary migration
\(S:E_{a,\mathrm{end}}\to E_{b,0}\) satisfying the declared cross-norm
bound
\[
 E_{b,0}\leq\kappa E_{a,\mathrm{end}}+\sigma .
\]
Then the composed endpoint obeys
\[
 E_{b,\mathrm{end}}\leq K_b^{r_b}\left[
 \beta_b^{J_b}(\kappa E_{a,\mathrm{end}}+\sigma)+
 C_b\frac{1-\beta_b^{J_b}}{1-\beta_b}\right]+R_b .
\]
The same formula iterates over a finite family of heterogeneous segments by
induction, carrying each \(\kappa_\ell\) and \(\sigma_\ell\). If a cut falls
inside a block, or \(q_a\ne q_b\), the remainder and re-phasing factors must be
included explicitly; no phase-independent global \(q\)-step contraction
follows.

The map \(S\) must preserve the protected projection,
\(p_{b,0}S=p_{a,\mathrm{end}}\), and transport owner/duty/permission/epoch/
tombstone/scope fields, replay receipts, resource units, and typed update discrepancies. Its conversion, calibration, buffering, communication,
maintenance, rollback, recovery, and shared-peak costs are charged in the
full lifecycle ledger, and the progress rank composes rather than resets.
Missing \(\kappa\), \(\sigma\), phase, protected-field, update, ledger, rank,
or replay evidence leaves the concatenation \(\mathsf{OPEN}\).
\end{proposition}

\begin{proof}
Apply the finite block and remainder inequalities of
Proposition~\ref{app:producttransport} separately to the two subchains, then
substitute the typed boundary inequality and apply the second segment bound.
Induction gives the finite multi-segment statement. The semantic premises are
independent of the numerical inequality: exact protected projection, receipt
commutation, epoch fences, update-discrepancy accounting, ledger closure, and
rank transport are required for the endpoint to denote the same contract
state on both sides of a boundary.
\end{proof}

\paragraph{Heterogeneous-phase counterexamples.}
Scalar internal maps \(a_1=0.1,a_2=5\) and
\(b_1=5,b_2=0.1\) have aligned two-step products
\(a_2a_1=b_2b_1=0.5\), but the cross-boundary pair \(b_1a_2=25\). Thus
local \(q=2\) contraction does not imply a phase-independent global law. A
boundary map \(S=100\) with two local factors \(0.5\) has nominal product
\(0.25\) when \(\kappa\) is omitted, but actual endpoint factor \(25\). A
segment budget that omits migration or recovery peaks, or a boundary that
resets receipt schema or epoch tombstones, likewise invalidates the composed
certificate and returns \(\mathsf{OPEN}\).
\begin{proposition}[Finite multi-segment heterogeneous block composition]
\label{app:multiblock}
Let a finite chain contain \(L\geq 2\) heterogeneous segments. For segment
\(\ell\), let \(a_\ell\) and \(d_\ell\) denote the local block/remainder
coefficients from Proposition~\ref{app:heteroq}:
\[
 a_\ell:=K_\ell^{r_\ell}\beta_\ell^{J_\ell},\qquad
 d_\ell:=K_\ell^{r_\ell}C_\ell
 \frac{1-\beta_\ell^{J_\ell}}{1-\beta_\ell}+R_\ell .
\]
Thus \(E_{\ell,\mathrm{out}}\leq a_\ell E_{\ell,\mathrm{in}}+d_\ell\). At
 each boundary \(\ell=0,\ldots,L-2\), require a typed migration with
\[
 E_{\ell+1,\mathrm{in}}\leq\kappa_\ell E_{\ell,\mathrm{out}}+\sigma_\ell .
\]
Set \(\alpha_\ell:=a_{\ell+1}\kappa_\ell\) and
\(\delta_\ell:=a_{\ell+1}\sigma_\ell+d_{\ell+1}\). Then
\[
 E_{L-1,\mathrm{out}}\leq
 \left(\prod_{\ell=0}^{L-2}\alpha_\ell\right)
 (a_0E_{0,\mathrm{in}}+d_0)
 +\sum_{\ell=0}^{L-2}
 \left(\prod_{h=\ell+1}^{L-2}\alpha_h\right)\delta_\ell .
\]
Empty products equal one. This is a finite prefix bound; differing \(q_\ell\),
non-aligned cuts, and all phase/remainder factors remain inside the declared
\(a_\ell,d_\ell\).

Every boundary map must preserve the protected projection
\(p_{\ell+1}S_\ell=p_\ell\) and transport owner/duty/permission/epoch/
tombstone/scope, replay receipts, resource units, and typed update discrepancies. Conversion, maintenance, calibration, communication,
buffering, rollback, recovery, and shared peaks compose with the full
prefix-peak ledger operation \(\otimes\), rather than by adding isolated
segment budgets. A transported rank (for example, the lexicographic pair of
remaining segment index and local rank) must strictly decrease; epoch fences
and replay schemas must compose. Missing any witness returns \(\mathsf{OPEN}\).
\end{proposition}

\begin{proof}
The local inequalities and boundary migration give
\(E_{\ell+1,\mathrm{out}}\leq\alpha_\ell E_{\ell,\mathrm{out}}+\delta_\ell\).
Induction on \(\ell\) yields the displayed finite product-sum. The semantic
conditions are preserved only if the boundary maps commute with protected
projections, update discrepancies, receipts, epoch fences, and the ledger;
lexicographic rank prevents a local segment from silently restarting an
unresolved obligation.
\end{proof}

\paragraph{Multi-segment accumulation counterexamples.}
Three scalar segments with local factors \(a_0=a_1=a_2=0.5\) and boundary
factors \(\kappa_0=\kappa_1=10\) have actual endpoint factor \(12.5\), while
omitting the boundaries reports \(0.125\). Each segment can separately fit a
budget of one third of \(B\) while the two migration/recovery peaks make the
composed prefix exceed \(B\). A reset of the local rank, epoch tombstone, or
receipt schema can similarly make every segment locally replayable while the
concatenated chain forks; these cases return \(\mathsf{OPEN}\).
\begin{proposition}[Finite noncommuting operator-valued boundary composition]
\label{app:noncommute}
Let \(L\geq 2\) finite segments use a common finite error cone
\(\mathbb R_+^d\) with componentwise order. Suppose each local certificate
has nonnegative \(A_\ell,b_\ell\) and each typed boundary has nonnegative
\(M_\ell,s_\ell\) such that
\[
 x_{\ell,\mathrm{out}}\preceq A_\ell x_{\ell,\mathrm{in}}+b_\ell,
 \qquad
 x_{\ell+1,\mathrm{in}}\preceq M_\ell x_{\ell,\mathrm{out}}+s_\ell .
 \]
Set \(G_\ell:=A_{\ell+1}M_\ell\) and
\(h_\ell:=A_{\ell+1}s_\ell+b_{\ell+1}\). Then the ordered finite
product-sum bound is
\[
 x_{L-1,\mathrm{out}}\preceq
 G_{L-2}\cdots G_0(A_0x_{0,\mathrm{in}}+b_0)
 +\sum_{j=0}^{L-2}
 (G_{L-2}\cdots G_{j+1})h_j,
 \]
where products are written in execution order and empty products are the
identity. The \(A_\ell\), \(M_\ell\) need not commute; replacing their order by
scalar norms or a reordered product is not a certificate.

Every \(M_\ell\) must have a typed complete-state image and preserve the
protected projection (for example, \(p_{\ell+1}M_\ell=p_\ell\)); owner, duty,
permission, epoch/tombstone/scope, receipt, unit, and typed update-discrepancy
fields must transport. Conversion, communication, buffering, rollback,
recovery, and shared peaks, including the \(s_\ell\) and \(b_\ell\) charges,
compose through one prefix-peak ledger. A transported rank and source-denied
replay schema must compose at each boundary; otherwise the status is
\(\mathsf{OPEN}\).
\end{proposition}

\begin{proof}
Combining the local and boundary inequalities gives
\(x_{\ell+1,\mathrm{out}}\preceq G_\ell x_{\ell,\mathrm{out}}+h_\ell\).
Induction in execution order yields the displayed ordered product-sum. The
semantic conditions are independent of the componentwise inequality: a matrix
product may be numerically valid while changing units, protected fields, or
replay identity, so those fields and the ledger must be transported as part of
the same witness.
\end{proof}

\paragraph{Noncommuting boundary counterexample.}
Take \(d=2\), \(A_\ell=I\), \(b_\ell=s_1=0\), \(x_{0,\mathrm{in}}=(1,0)^\top\), and
\[
  M_0=\begin{pmatrix}1&1\\0&1\end{pmatrix},\qquad
  M_1=\begin{pmatrix}1&0\\1&1\end{pmatrix}.
 \]
The execution order gives \(M_1M_0x_0=(1,1)^\top\), whereas the reordered
product \(M_0M_1x_0=(2,1)^\top\). A component budget \((1,1)^\top\) therefore
accepts the declared order and rejects the reordered one, even though each
boundary is individually nonnegative. With \(x_0=0\), \(s_0=(1,0)^\top\), and
all other additive terms zero, the propagated discrepancy is \(M_1s_0=(1,1)^\top\);
it cannot be replaced by an unweighted sum of local errors. These finite
counterexamples block commutation, scalar-only boundary factors, arbitrary
switching, JSR, and infinite-horizon conclusions.
\begin{proposition}[Finite common-weight ordered-product transport]
\label{app:commonweight}
Let \(B_\ell:E_\ell\to E_{\ell+1}\) be nonnegative linear maps between
finite positive cones, and let \(c_\ell\in E_{\ell+1}\) be the typed additive
term from Proposition~\ref{app:noncommute}. Suppose there are vectors
\(w_\ell\gg0\) and scalars \(\rho_\ell\geq0\) with
\[
 B_\ell w_\ell\preceq \rho_\ell w_{\ell+1}
 \qquad(\ell=0,\ldots,L-2).
\]
For \( \|z\|_{w_\ell,\infty}:=\max_i z_i/(w_\ell)_i\), every finite ordered
chain satisfies
\[
\|x_{L-1,\mathrm{out}}\|_{w_{L-1},\infty}
\leq
\left(\prod_{\ell=0}^{L-2}\rho_\ell\right)
\|A_0x_{0,\mathrm{in}}+b_0\|_{w_0,\infty}
+\sum_{j=0}^{L-2}
\left(\prod_{h=j+1}^{L-2}\rho_h\right)
\|c_j\|_{w_{j+1},\infty}.
\]
The products are scalar bounds on the declared execution order; they do not
permit reordering of the operators. A common weight \(w_\ell=w\) is a special
case. This is a finite certificate; even \(\rho_\ell<1\) does not imply a
joint-spectral-radius, arbitrary-switching, or infinite-horizon result.

The weight witnesses must use the same translated complete state at every
boundary. Protected projections, owner/duty/permission/epoch/tombstone/scope,
receipts, units, typed update discrepancies, prefix-peak ledger, transported
rank, and source-denied replay must compose. Missing a weight, an ordered
product witness, or any semantic field returns \(\mathsf{OPEN}\).
\end{proposition}

\begin{proof}
For \(z\succeq0\), \(z\preceq\|z\|_{w_\ell,\infty}w_\ell\), so positivity and
the weight inequality give
\(\|B_\ell z\|_{w_{\ell+1},\infty}\leq
\rho_\ell\|z\|_{w_\ell,\infty}\). Apply this inequality to the ordered
recurrence from Proposition~\ref{app:noncommute} and induct on the segment
index. The semantic and ledger premises are carried as part of the same
boundary witness and are not consequences of the numerical norm bound.
\end{proof}

\paragraph{Local spectral-radius counterexample.}
Take \(A_\ell=I\), \(b_\ell=s_\ell=0\), and
\[
 M_0=0.9\begin{pmatrix}1&1\\0&1\end{pmatrix},\qquad
 M_1=0.9\begin{pmatrix}1&0\\1&1\end{pmatrix}.
\]
Each matrix has spectral radius \(0.9\), but
\(M_1M_0=0.81\begin{pmatrix}1&1\\1&2\end{pmatrix}\) has spectral radius
\(0.81(3+\sqrt5)/2\approx2.12\). From \(x_0=(1,1)^\top\), the ordered
endpoint is \(0.81(2,3)^\top=(1.62,2.43)^\top\), violating a component
budget of \((1,1)^\top\). Thus local spectral-radius checks do not provide a
common weight or a finite ordered-product certificate.

\paragraph{Finite-product/common-weight separation.}
The converse implication also fails. Let
\[
 B_0=\tfrac12\begin{pmatrix}1&1\\0&1\end{pmatrix},\qquad
 B_1=\tfrac12\begin{pmatrix}1&0\\1&1\end{pmatrix},
 \qquad c_0=c_1=0.
\]
The declared two-step order has
\[
 B_1B_0=\tfrac14\begin{pmatrix}1&1\\1&2\end{pmatrix},
 \qquad \|B_1B_0\|_\infty=\tfrac34<1,
\]
so the finite ordered-product certificate contracts in the ordinary infinity
norm. However, no strict common positive-weight contraction exists. Writing
\(r=(w)_2/(w)_1>0\), the inequalities \(B_iw\preceq\rho w\) require
\[
 \rho\geq\tfrac12(1+r),\qquad
 \rho\geq\tfrac12(1+r^{-1}).
\]
Their maximum is minimized at \(r=1\), where it equals \(1\); hence
\(\rho<1\) is impossible. A finite ordered product can therefore contract
without a strict common weight, and neither certificate implies a joint
spectral-radius, arbitrary-switching, or infinite-horizon result. The same
typed boundary, ledger, rank, protected-field, and replay witnesses remain
required; missing them returns \(\mathsf{OPEN}\).

\paragraph{Contract qualification of the separation.}
The matrices act on a complete translated state, not on an error vector alone.
For each boundary \(i\), admissibility additionally requires an exact protected
projection \(p_{i+1}T_i=p_i\), an owner/permission/epoch/tombstone/scope image,
and typed update transport
\[
 C_{i+1}U_i=\widehat U_iC_i+\Delta_i^U,
\]
with the image of \(\Delta_i^U\) charged into the typed additive term. The
prefix-peak ledger must compose migration, communication, buffering, rollback,
recovery, and shared-resource charges without double spending; the transported
well-founded rank may not reset; and a source-denied replay receipt must
reproduce successor, effect, owner, permission, epoch, resource, and charge
fields. The \(0.75\) numerical certificate is therefore accepted only with
these witnesses. A missing field, update discrepancy, rank, ledger, or replay
receipt returns \(\mathsf{OPEN}\), even though the finite matrix product
contracts.

\paragraph{Reusable finite transport-witness schema.}
For a declared finite schedule \(I=\{0,\ldots,N-1\}\), a transport witness
package is the tuple
\[
 \mathcal W=(\{E_i,B_i,c_i,T_i,p_i,C_i,U_i,\Delta_i^U,\mathsf L_i,
 \mathsf{rk}_i,\mathsf{rec}_i\}_{i\in I},\beta),
\]
where \(E_i\) is the typed positive cone, \(T_i\) the complete-state
translation, \(p_i\) the protected projection, \(C_i,U_i\) the representation
and update maps, \(\Delta_i^U\) the typed update discrepancy, \(\mathsf L_i\)
the prefix-peak ledger, \(\mathsf{rk}_i\) the transported progress rank,
\(\mathsf{rec}_i\) the source-denied replay receipt, and \(\beta\) the
declared budget. The package is accepted only if the ordered numerical bound,
protected-field transport, update square, ledger budget, rank condition, and
replay equality all hold on the same translated state; otherwise its status is
\(\mathsf{OPEN}\). This schema is finite and contract-relative and does not
turn a passing schedule into a common-weight or infinite-horizon certificate.

\paragraph{Second switched family and field-isolation counterexamples.}
To separate the numerical witness from the semantic contract, take the
two-stage diagonal family
\[
 B_0=\begin{pmatrix}1/2&0\\0&1/4\end{pmatrix},\qquad
 B_1=\begin{pmatrix}1/4&0\\0&1/2\end{pmatrix},\qquad c_0=c_1=0.
\]
Its declared order has
\[
 B_1B_0=\tfrac18 I,\qquad \|B_1B_0\|_\infty=\tfrac18<1.
\]
Use a complete state
\(\sigma_i=(e_i,d_i,h_i,o_i,r_i,\lambda_i,\omega_i)\), where $d_i$ is a
protected duty, $h_i$ is update-relevant hidden metadata, $o_i$ is the
owner/epoch record, $r_i$ the resource charge, $\lambda_i$ the transported
rank, and $\omega_i$ the replay receipt. The following one-field deletions
leave the displayed finite numerical product unchanged but make the witness
status $\mathsf{OPEN}$:
\begin{enumerate}
\item If the protected projection is omitted, let a boundary map send
  $d_0=1$ to $d_1=0$ while sending $e_1=B_0e_0$ and preserving all other
  numerical coordinates. Then $p_1T_0(e,1)=0\ne1=p_0(e,1)$ although the
  matrix product contracts; a duty can be silently dropped.
\item If the typed update discrepancy is omitted, let the stage-$0$ encoding
  discard $h_0$ while the next typed state restores it. More precisely, take
  $S_0=E_0\times H$, $R_0=E_0$, $S_1=E_1\times H$, $R_1=E_1\times H$,
  with $H=\{0,1\}\subset\mathbb R$, $C_0(e,h)=e$,
  $C_1(e',h')=(e',h')$, and
  $U_0(e,h)=(B_0e,h)$. Any claimed $\widehat U_0:R_0\to R_1$ chooses one
  $\widehat h(e)$, so
  $\Delta^U_0(e,h)=C_1U_0(e,h)-\widehat U_0C_0(e,h)
  =(0,h-\widehat h(e))$ is nonzero for two states sharing $e$. The $e$
  component still follows the displayed product, but omitting or not charging
  $\Delta^U_0$ makes the update square false.
\item If the prefix ledger is omitted, assign stage charges
  $m_0=1.2$ and $m_1=0$ under a declared prefix-peak budget $\beta=1$.
  The numerical contraction remains valid, but the stage-$0$ peak already
  exceeds $\beta$, so the schedule is not resource-certified.
\item If transported rank is omitted, repeat a two-state nonterminal cycle
  with $T_0:1\mapsto0$ and $T_1:0\mapsto1$. The finite product still contracts,
  but no well-founded rank can strictly decrease on both edges, so finite
  stopping is unproved and the package is $\mathsf{OPEN}$.
\item If source-denied replay is omitted, let the replay receipt reproduce
  $e_1=B_0e_0$ but omit a hidden $q\in\{0,1\}$ from the receipt, with
  $T_i(e,q)=(B_ie,q)$ and next effect $a=q$. States $(e,0)$ and $(e,1)$
  have identical public numerical receipts and contracted endpoints but
  different effects; source-denied replay therefore fails without $q$ (or an
  equivalent random-state witness), and the package is $\mathsf{OPEN}$.
\end{enumerate}
These are semantic variants of the same finite switched numerical witness;
none licenses a joint-spectral-radius, arbitrary-switching, or
infinite-horizon claim.

\paragraph{Three-stage composition and no-reset gate.}
For a third finite schedule append
\[
 B_2=\tfrac12 I,\qquad B_2B_1B_0=\tfrac1{16}I,
 \qquad \|B_2B_1B_0\|_\infty=\tfrac1{16}<1.
\]
The numerical endpoint remains contractive only on the composed translated
state. With typed additive discrepancies $\eta_i\in E_i$ the numerical part is
\[
 e_3\preceq B_2B_1B_0e_0+B_2B_1\eta_0+B_2\eta_1+\eta_2.
\]
A valid composition therefore requires, at every boundary $i=0,1,2$,
the protected-scope map to satisfy
\(g_{i+1}\circ T_i=g_i\) on the declared protected domain and
\(p_{i+1}T_i=p_i\), the typed update square with its charged
\(\Delta_i^U\), a ledger operation that carries the previous prefix rather
than reinitializing it, a transported well-founded rank, with any separately declared finite-horizon stopping certificate recorded as an additional field, and a replay receipt with one stable schema. The status is $\mathsf{OPEN}$ whenever any boundary violates
one of these equations or inequalities. Each reset gives a
finite numerical pass but an \(\mathsf{OPEN}\) composition: a boundary that
drops the declared scope or $d_i$ violates protected transport; omitting a nonzero
\(\Delta_1^U\) hides a future update; resetting an additive lifecycle ledger
can make three stage charges and one boundary charge, each $0.6$, appear within
local unit budgets while the composed prefix charge $2.4$ exceeds a global
budget $2$; a repeatable rank cycle
prevents well-founded stopping; and changing the receipt schema to omit a
hidden $q$ yields divergent effects with the same public endpoint. Thus a
contract can be numerically compositional only when scope, update, ledger,
progress, and replay fields are transported without reset.

\begin{proposition}[Finite mixed-discrepancy and ledger composition]
\label{app:fourthmix}
Let $E_i=\mathbb R_+^2$ and
\[
 B_0=\operatorname{diag}(1/2,1/4),\quad
 B_1=\operatorname{diag}(1/4,1/2),\quad
 B_2=\tfrac12 I,\quad B_3=\tfrac14 I,
\]
with the ordered recurrence $e_{i+1}\preceq B_i e_i+\eta_i$ for
$i=0,1,2,3$.  Split every typed discrepancy as
\(\eta_i=\eta_i^{\rm conv}+\eta_i^{\rm upd}+\eta_i^{\rm replay}\in E_i$;
each component is included in the endpoint update, its resource charge, and
the source-denied replay receipt.  Let the lifecycle ledger be
$L_i=(a_i,p_i,\Theta_i)$, where $a_i$ is cumulative additive cost, $p_i$ is
the prefix peak, and $\Theta_i$ is the set of consumed unique charge tokens.
For a stage charge $\ell_i=(c_i,m_i,\Phi_i)$ define
\[
 (a,p,\Theta)\oplus(c,m,\Phi)
   =(a+c,\max\{p,m\},\Theta\cup\Phi),
 \qquad \Theta\cap\Phi=\varnothing,
\]
and require $L_{i+1}\succeq L_i\oplus\ell_i$, with every prefix satisfying
$a_i\leq\beta_a$ and $p_i\leq\beta_p$.  Initialize
$L_0=(0,0,\varnothing)$ and interpret $\succeq$ componentwise, with set
inclusion on the token field.  Assume also the previously declared
protected-scope transport, owner/epoch/permission checks, typed update square,
transported rank, and stable replay schema at all four boundaries.  Then the
finite certificate is closed and
\[
 e_4\preceq \tfrac1{64}e_0
   +\operatorname{diag}(1/32,1/16)\eta_0
   +\tfrac18\eta_1+\tfrac14\eta_2+\eta_3.
\]
If any stated semantic, discrepancy, or ledger premise fails, the status is
$\mathsf{OPEN}$.  This is a finite, contract-relative statement; it makes no
claim about joint spectral radius, arbitrary switching, or infinite-horizon
stability.
\end{proposition}

\begin{proof}
The matrix products are $B_1B_0=\tfrac18I$,
$B_2B_1B_0=\tfrac1{16}I$, and $B_3B_2B_1B_0=\tfrac1{64}I$.
Substitution into the four ordered inequalities gives the displayed bound,
with suffix products $B_3B_2B_1=\operatorname{diag}(1/32,1/16)$,
$B_3B_2=\tfrac18I$, and $B_3=\tfrac14I$.  Induction on the same four
boundaries, using $L_{i+1}\succeq L_i\oplus\ell_i$, preserves additive cost,
prefix peak, and token uniqueness.  The protected transport, update square,
rank, and replay hypotheses preserve the corresponding semantic fields.  A
missing component or a reset therefore invalidates the certificate and returns
$\mathsf{OPEN}$ even when the numerical matrix product is unchanged.
\end{proof}

\paragraph{Concrete mixed witness and reset counterexample.}
For $e_0=0$ and
\(\eta_0=(1,1),\eta_1=(1,2),\eta_2=(2,1),\eta_3=(1,2)\), the bound gives
$e_4\preceq(53/32,41/16)$, while all three typed components may be nonzero and
must be charged.  If each of four stages charges $c_i=0.4$ under a global
additive budget $\beta_a=1$, the true prefix costs are $0.4,0.8,1.2,1.6$.
Resetting $L$ to zero at each segment makes every local check pass but violates
$L_{i+1}\succeq L_i\oplus\ell_i$ and is therefore $\mathsf{OPEN}$.  Likewise,
omitting a nonzero component such as $\eta_1^{\rm upd}=(1,0)$ leaves the
$1/64$ numerical product unchanged but breaks the endpoint/update square and
returns $\mathsf{OPEN}$.

\begin{proposition}[Finite mixed-boundary schedule without common-weight inference]
\label{app:fifthmix}
On $E=\mathbb R_+^2$, consider the ordered recurrence
$x_{k+1}\preceq B_kx_k+c_k$ for $k=0,1,2$, with
\[
B_0=\tfrac12\begin{pmatrix}1&1\\0&1\end{pmatrix},\quad
B_1=\tfrac12\begin{pmatrix}1&0\\1&1\end{pmatrix},\quad
B_2=\operatorname{diag}(0.8,0.6),
\]
and $c_0=(0.01,0.02)^\top$, $c_1=(0.02,0.01)^\top$,
$c_2=(0.01,0.03)^\top$. If the complete translated state transports
protected projections, owner/duty/permission/epoch/tombstone/scope, units,
receipts, typed update discrepancies, a monotone prefix ledger, a transported
rank, and source-denied replay at every boundary, then
\[
B_2B_1B_0=\begin{pmatrix}0.20&0.20\\0.15&0.30\end{pmatrix},\qquad
B_2B_1c_0+B_2c_1+c_2=(0.03,0.045)^\top,
\]
so $\|x_3\|_\infty\leq0.45\|x_0\|_\infty+0.045$ and, for
$\|x_0\|_\infty\leq0.1$, the endpoint is at most $0.09$. This finite
ordered-product certificate does not imply a common positive weight,
joint-spectral-radius, arbitrary-switching, or infinite-horizon result.
\end{proposition}

\begin{proof}
Direct multiplication gives the displayed product and suffix sum. Taking the
induced infinity norm yields the stated endpoint inequality. If a common
$w\gg0$ and $\rho<1$ satisfied $B_iw\preceq\rho w$ for both $B_0$ and $B_1$,
then with $r=w_2/w_1$ one would have
$\rho\geq\frac12(1+r)$ and
$\rho\geq\frac12(1+r^{-1})$, whose maximum is at least $1$; hence no such
strict common-weight witness exists. The semantic fields are transported by
the stated boundary hypotheses, and any missing field returns $\mathsf{OPEN}$.
\end{proof}

\paragraph{Mixed-boundary OPEN witnesses.}
If the declared boundary factor $B_2=I$ is used in place of the actual
$10I$, then $B_1B_0(0.1,0.1)^\top=(0.05,0.075)^\top$ appears within a
$(0.1,0.1)$ component budget, whereas the actual factor gives
$(0.5,0.75)^\top$; an uncharged conversion or ledger field therefore returns
$\mathsf{OPEN}$. Likewise, a declared execution order $B_1B_0$ passes the
component budget $(0.06,0.08)$ at $(0.1,0.1)^\top$, but a stale or reordered
replay using $B_0B_1$ gives $(0.075,0.05)^\top$ and violates the first
component budget. The numerical maps alone cannot certify boundary identity,
order, or replay.

\begin{proposition}[Finite four-stage schedule with continued boundary transport]
\label{app:sixthmix}
On $E=\mathbb R_+^2$, consider the ordered recurrence
$x_{k+1}\preceq B_kx_k+c_k$ for $k=0,1,2,3$, with
\[
B_0=\tfrac12\begin{pmatrix}1&1\\0&1\end{pmatrix},\quad
B_1=\tfrac12\begin{pmatrix}1&0\\1&1\end{pmatrix},\quad
B_2=\operatorname{diag}(4/5,3/5),\quad
B_3=\operatorname{diag}(7/10,9/10),
\]
and $c_0=(1/100,2/100)^\top$, $c_1=(2/100,1/100)^\top$,
$c_2=(1/100,3/100)^\top$, $c_3=(1/200,1/250)^\top$.
If each boundary transports the complete protected state---owner, duty,
permission, epoch, tombstone, scope, units, receipts, typed update
discrepancies, the monotone prefix ledger, transported rank, and a
source-denied replay witness---then direct multiplication gives
\[
\begin{aligned}
P&=B_3B_2B_1B_0
 =\begin{pmatrix}7/50&7/50\\27/200&27/100\end{pmatrix},\\
q&=B_3B_2B_1c_0+B_3B_2c_1+B_3c_2+c_3
 =\begin{pmatrix}13/500\\89/2000\end{pmatrix}.
\end{aligned}
\]
Consequently,
$\|x_4\|_\infty\le(81/200)\|x_0\|_\infty+89/2000$; in particular,
$\|x_0\|_\infty\le1/10$ implies $\|x_4\|_\infty\le17/200$.
This is a finite ordered-product statement.  It does not imply a common
positive weighted norm, a joint spectral radius bound, arbitrary-switching
stability, or an infinite-horizon result.
\end{proposition}

\begin{proof}
The displayed matrices and affine terms are obtained by direct multiplication;
$\|P\|_\infty=81/200$, and the componentwise triangle inequality yields the
stated endpoint bound.  A strict common positive weight is already impossible
for $B_0,B_1$: writing the weight ratio as $r>0$ gives the necessary bounds
$\rho\ge\tfrac12(1+r)$ and $\rho\ge\tfrac12(1+1/r)$, whose maximum is at
least one.  Finally, order matters because $B_1B_0\ne B_0B_1$; hence the
boundary replay witness is part of the certificate rather than a numerical
afterthought.
\end{proof}

For a concrete ledger boundary, charge $\chi_i=0.02$ to each of the four
stages while declaring a global additive budget $\beta_a=0.05$.  The true
prefix cost is $0.08>\beta_a$, whereas resetting the ledger at each stage makes
every local check pass.  The reset certificate is therefore $\mathsf{OPEN}$;
the numerical product above is unchanged, but the complete lifecycle charge and
transported ledger remain part of the valid witness.

\begin{proposition}[Finite three-stage schedule with calibration transport]
\label{app:seventhmix}
On $E=\mathbb R_+^2$, consider the ordered recurrence
$x_{k+1}\preceq B_kx_k+c_k$ for $k=0,1,2$, with
\[
B_0=\operatorname{diag}(3/4,1/2),\quad
B_1=\begin{pmatrix}1/2&1/4\\0&1/2\end{pmatrix},\quad
B_2=\operatorname{diag}(2/3,3/5),
\]
and $c_0=(1/100,1/200)^\top$, $c_1=(1/200,1/100)^\top$,
$c_2=(1/100,1/100)^\top$.  If each boundary transports the declared
local-to-global norm and unit calibration, clock/latency receipt, protected
projection, owner/duty/permission/epoch/tombstone/scope, typed update
discrepancies, monotone prefix ledger, transported rank, and source-denied
replay witness, then
\[
\begin{aligned}
P&=B_2B_1B_0
 =\begin{pmatrix}1/4&1/12\\0&3/20\end{pmatrix},\\
q&=B_2B_1c_0+B_2c_1+c_2
 =\begin{pmatrix}7/400\\7/400\end{pmatrix}.
\end{aligned}
\]
Consequently,
$\|x_3\|_\infty\le(1/3)\|x_0\|_\infty+7/400$; in particular,
$\|x_0\|_\infty\le1/10$ implies $\|x_3\|_\infty\le61/1200$.
This is a finite ordered-product statement.  It does not imply a common
positive weighted norm, a joint spectral radius bound, arbitrary-switching
stability, or an infinite-horizon result.
\end{proposition}

\begin{proof}
Direct multiplication gives the displayed product and affine suffix, and
$\|P\|_\infty=1/3$; the componentwise triangle inequality gives the endpoint
bound.  For the semantic boundary, take the positive calibration map
$S=\operatorname{diag}(4,1/4)$.  A local error $(1,1)^\top$ has local infinity
norm $1$ but global norm $\|S(1,1)^\top\|_\infty=4$.  If $S$ and its unit/clock
receipt are omitted, the local matrices still reproduce the same numerical
$P,q$, while the global error and resource charge are not identified; the
complete certificate therefore returns $\mathsf{OPEN}$.
\end{proof}

\begin{proposition}[Finite cross-scale calibration composition]
\label{app:eighthmix}
Let
\[
A_0=\operatorname{diag}(3/4,1/2),\quad
A_1=\begin{pmatrix}1/2&1/4\\0&1/2\end{pmatrix},\quad
C=\operatorname{diag}(4/3,3/4),\quad
A_2=\operatorname{diag}(2/3,3/5),
\]
and define the global ordered recurrence
$x_{k+1}\preceq B_kx_k+c_k$ by $B_0=A_0$, $B_1=CA_1$, and $B_2=A_2$,
with $c_0=(1/100,1/200)^\top$, $c_1=(1/200,1/100)^\top$,
and $c_2=(1/100,1/100)^\top$.  If the boundary transports $C$ together with
its units, clock/latency receipt, protected projection, owner/duty/permission/
epoch/tombstone/scope, typed update discrepancies, monotone prefix ledger,
transported rank, and source-denied replay witness, then
\[
\begin{aligned}
B_1&=\begin{pmatrix}2/3&1/3\\0&3/8\end{pmatrix},\qquad
P=B_2B_1B_0
 =\begin{pmatrix}1/3&1/9\\0&9/80\end{pmatrix},\\
q&=B_2B_1c_0+B_2c_1+c_2
 =\begin{pmatrix}17/900\\137/8000\end{pmatrix}.
\end{aligned}
\]
Consequently,
$\|x_3\|_\infty\le(4/9)\|x_0\|_\infty+17/900$; in particular,
$\|x_0\|_\infty\le1/10$ implies $\|x_3\|_\infty\le19/300$.
This is a finite cross-scale ordered-product statement.  It does not imply a
common positive weighted norm, a joint spectral radius bound, arbitrary-
switching stability, or an infinite-horizon result.
\end{proposition}

\begin{proof}
Direct multiplication gives the displayed $B_1$, $P$, and $q$, with
$\|P\|_\infty=4/9$; the componentwise triangle inequality yields the endpoint
bound.  If the calibration is omitted, the nominal product is
$A_2A_1A_0=\begin{psmallmatrix}1/4&1/12\\0&3/20\end{psmallmatrix}$ with
infinity norm $1/3$, so the local numerical certificate can still appear to
pass.  However, $C(1,1)^\top=(4/3,3/4)^\top$ has global infinity norm $4/3$;
without the calibration and its unit/clock receipt, the global error, resource
charge, and boundary identity are not certified, so the complete witness is
$\mathsf{OPEN}$.
\end{proof}

\begin{proposition}[Finite calibrated lifecycle-ledger composition]
\label{app:ninthmix}
Let $E=\mathbb R_+^2$ and consider the ordered recurrence
$x_{k+1}\preceq B_kx_k+c_k$ for $k=0,1,2,3$, with
\[
\begin{aligned}
B_0&=\operatorname{diag}(3/4,1/2),\\
B_1&=C A_1,\qquad
A_1=\begin{pmatrix}1/2&1/4\\0&1/2\end{pmatrix},quad
C=\operatorname{diag}(4/3,3/4),\\
B_2&=\operatorname{diag}(2/3,3/5),\qquad
B_3=\operatorname{diag}(3/4,4/5),
\end{aligned}
\]
and
\[
\begin{aligned}
c_0&=(1/100,1/200)^\top,&
c_1&=(1/200,1/100)^\top,\\
c_2&=(1/100,1/100)^\top,&
c_3&=(1/200,1/250)^\top.
\end{aligned}
\]
Suppose every boundary transports the complete protected state, units,
clock/latency receipt, owner/duty/permission/epoch/tombstone/scope, typed
updates, transported rank, and source-denied replay.  Let the lifecycle ledger
carry a nonnegative charge $\chi_k$ and a unique token $\tau_k$ at each stage,
with monotone prefix update
\[
 C_{k+1}=C_k+\chi_k,\qquad
 M_{k+1}=\max\{M_k,C_{k+1}\},\qquad
 T_{k+1}=T_k\mathbin{\dot\cup}\{\tau_k\},
\]
and accept the schedule only when $C_4\leq\beta$ and all tokens in $T_4$
are distinct.  Then direct multiplication gives
\[
\begin{aligned}
B_1&=\begin{pmatrix}2/3&1/3\\0&3/8\end{pmatrix},\qquad
P=B_3B_2B_1B_0
 =\begin{pmatrix}1/4&1/12\\0&9/100\end{pmatrix},\\
q&=B_3B_2B_1c_0+B_3B_2c_1+B_3c_2+c_3
 =\begin{pmatrix}23/1200\\177/10000\end{pmatrix}.
\end{aligned}
\]
Consequently,
$\|x_4\|_\infty\leq(1/3)\|x_0\|_\infty+23/1200$; in particular,
$\|x_0\|_\infty\leq1/10$ implies $\|x_4\|_\infty\leq21/400$.
The concrete charge vector
\[
(\chi_0,\chi_1,\chi_2,\chi_3)
 =(3/100,2/100,3/100,4/100)
\]
has total $3/25$, so it is admissible for any declared $\beta\geq3/25$.
This is a finite ordered contract-relative statement and does not imply a
common weighted norm, a joint spectral-radius bound, arbitrary-switching
stability, or an infinite-horizon result.
\end{proposition}

\begin{proof}
The displayed matrices and affine suffix follow by direct multiplication; the
row-sum norm of $P$ is $1/3$, and the componentwise triangle inequality gives
the endpoint bound.  The ledger recurrences are monotone, so a valid budget
and disjoint token set cannot be silently double-spent.  If a declared global
budget is instead $\beta_0=1/10$ and the ledger is reset at each boundary,
each local charge passes while the true total $3/25>\beta_0$; the numerical
product is unchanged but the complete certificate is $\mathsf{OPEN}$.  If
the calibration $C$ is omitted, the nominal product is
\[
B_3B_2A_1B_0
 =\begin{pmatrix}3/16&1/16\\0&3/25\end{pmatrix},\qquad
\|B_3B_2A_1B_0\|_\infty=1/4,
\]
with nominal affine suffix $(29/1600,9/500)^\top$; nevertheless
$C(1,1)^\top=(4/3,3/4)^\top$ has global infinity norm $4/3$.
Missing calibration, units, clock receipt, boundary identity, or the monotone
ledger therefore returns $\mathsf{OPEN}$.
\end{proof}

\begin{proposition}[Finite delayed-effect recovery with epoch fencing]
\label{app:tenthmix}
Let $E=\mathbb R_+^2$ and consider the ordered recurrence
$x_{k+1}\preceq B_kx_k+c_k$ for $k=0,1,2,3$, with
\[
\begin{aligned}
B_0&=\operatorname{diag}(3/4,1/2),\\
B_1&=C A_1,\qquad
A_1=\begin{pmatrix}1/2&1/4\\0&1/2\end{pmatrix},\quad
C=\operatorname{diag}(4/3,3/4),\\
B_2&=\begin{pmatrix}1/2&1/4\\1/8&1/2\end{pmatrix},\qquad
B_3=\operatorname{diag}(2/3,3/5),
\end{aligned}
\]
and
\[
\begin{aligned}
c_0&=(1/100,1/200)^\top,&
c_1&=(1/200,1/100)^\top,\\
c_2&=(1/100,1/100)^\top,&
c_3&=(1/200,1/250)^\top.
\end{aligned}
\]
Suppose each boundary transports calibration and unit/clock receipts, owner,
duty, permission, epoch $\nu$, command identifiers, tombstones, leases, the
in-flight effect queue, effect/compensation/recovery receipts, typed updates,
a monotone resource ledger, a transported rank, and a source-denied replay
witness.  Let the lifecycle charges be
\[
(\chi_0,\chi_1,\chi_2,\chi_3)
 =(1/50,3/100,1/25,3/100),
\]
and accept a finite prefix only when its cumulative charge is at most the
declared budget and every command token is unique.  Then direct multiplication
gives
\[
\begin{aligned}
B_1&=\begin{pmatrix}2/3&1/3\\0&3/8\end{pmatrix},\qquad
P=B_3B_2B_1B_0
 =\begin{pmatrix}1/6&25/288\\3/80&11/160\end{pmatrix},\\
q&=B_3B_2B_1c_0+B_3B_2c_1+B_3c_2+c_3
 =\begin{pmatrix}521/28800\\233/16000\end{pmatrix}.
\end{aligned}
\]
Consequently,
$\|x_4\|_\infty\leq(73/288)\|x_0\|_\infty+521/28800$; in particular,
$\|x_0\|_\infty\leq1/10$ implies $\|x_4\|_\infty\leq139/3200$.
The true charge total is $3/25$.  Thus a declared budget $\beta_0=1/10$
must return $\mathsf{OPEN}$ even if each local check passes after a hidden
ledger reset.  This is a finite recovery and epoch-fencing statement; it does
not imply a common weighted norm, a joint-spectral-radius bound,
arbitrary-switching stability, or an infinite-horizon result.
\end{proposition}

\begin{proof}
The displayed product, row-sum norm $\|P\|_\infty=73/288$, and affine suffix
follow by direct multiplication; the componentwise triangle inequality gives
the endpoint bound.  The complete translated state is necessary for the
semantic claim.  If a command with identifier $a$ is issued at epoch $\nu$,
a revoke commits epoch $\nu+1$ and tombstones $a$, and a delayed copy then
arrives, the epoch fence and tombstone reject it.  Omitting either field lets
the old command execute a duplicate or unauthorized effect while leaving the
numerical $P,q$ unchanged, so the certificate is $\mathsf{OPEN}$.  The same
status follows from a hidden ledger reset because
$1/50+3/100+1/25+3/100=3/25>\beta_0$; calibration, unit/clock, lease,
effect, compensation, recovery, rank, and source-denied replay receipts remain
part of the accepted witness.
\end{proof}

\begin{proposition}[Finite causal-receipt ordering and partial-failure replay closure]
\label{app:causalreplay}
Let $E=\mathbb R_+^2$ and consider the finite ordered recurrence
$x_{k+1}\preceq B_kx_k+c_k$ for $k=0,1,2$, with
\[
 B_0=\begin{pmatrix}1/2&1/4\\0&1/2\end{pmatrix},\qquad
 B_1=\begin{pmatrix}1/3&0\\1/5&2/5\end{pmatrix},\qquad
 B_2=\operatorname{diag}(3/4,1/2),
\]
and
\[
 c_0=(1/100,1/200)^\top,\quad
 c_1=(1/200,1/100)^\top,\quad
 c_2=(1/100,1/100)^\top.
\]
Suppose each boundary transports the complete protected state together with
calibration, units, clock/latency receipt, owner/duty/permission, epoch and
tombstone, in-flight effects, compensation/recovery receipts, typed updates,
and a causal receipt plus an executed-do/commit receipt carrying the actual
owner, epoch, lease, and effect identifier
\[
 r_k=(\mathsf{id}_k,\mathsf{parent}_k,\mathsf{seq}_k,
       \mathsf{epoch}_k,\mathsf{digest}_k,\mathsf{scope}_k).
\]
Assume the effect receipts form a finite DAG with a declared predecessor
relation and that a failed effect selects exactly one terminal branch:
forward completion, compensation, or recovery. Admit a finite prefix only when parent digests match committed receipts,
sequence numbers are strictly increasing within each effect stream, command
identifiers are idempotent, the permission/lease check is valid at the
declared epoch, at most one terminal branch is committed, and source-denied
replay reproduces the same ordered receipt and effect graph. Let the prefix charges be
$(\chi_0,\chi_1,\chi_2)=(1/100,1/50,3/100)$ and require a monotone ledger
with budget $\beta\geq1/10$. Then
\[
 P=B_2B_1B_0=
 \begin{pmatrix}1/8&1/16\\1/20&1/8\end{pmatrix},
 \qquad \|P\|_\infty=3/16,
\]
and
\[
 q=B_2B_1c_0+B_2c_1+c_2
   =\begin{pmatrix}13/800\\17/1000\end{pmatrix}.
\]
Consequently,
\[
 \|x_3\|_\infty\leq (3/16)\|x_0\|_\infty+13/800;
 \qquad \|x_0\|_\infty\leq1/10
 \Rightarrow \|x_3\|_\infty\leq7/200.
\]
The causal-receipt conditions are part of the same finite certificate: if a
parent digest, sequence order, idempotency key, terminal-branch tag,
executed-do receipt, or effect receipt is omitted,
the numerical $P,q$ may remain unchanged but the status is
$\mathsf{OPEN}$.
\end{proposition}

\begin{proof}
Direct multiplication gives the displayed $P$ and $q$; its row sums are
$3/16$ and $7/40$, so the infinity-norm bound and the endpoint bound follow
from the componentwise triangle inequality. The charge total is
$1/100+1/50+3/100=3/50\leq1/10$, and the monotone ledger prevents a local
prefix from silently discarding an earlier charge. The remaining premises
are semantic closure conditions: parent and sequence fields make the effect
graph acyclic and ordered, the idempotency key prevents double application,
the unique terminal-branch tag prevents a forward effect and its compensation
from both committing, and the replay receipt fixes the same committed
successor. They are not consequences of the numerical inequality and must
therefore be transported at every boundary.
\end{proof}

\paragraph{Causal-order counterexample.}
Let command $a$ commit a protected resource while its lease is active and let
command $b$ revoke that lease, with $\mathsf{parent}_b=\mathsf{digest}(a)$.
The lawful order is $a$ followed by $b$. If the parent digest and sequence
field are omitted, a replay may deliver $b$ before $a$; the same cone
recurrence, and hence the same $P,q$, is still reported, but $a$ then executes
after revocation and is unauthorized. Unique command identifiers alone do not
repair this causal fork: the ordered effect receipt and its parent relation are
required. Likewise, if a failed effect commits both its forward and compensation
branches, their numerical residuals can cancel while the physical effect and
ledger charge are double spent. The complete certificate is therefore
$\mathsf{OPEN}$. This finite result
does not imply arbitrary-order convergence, causal discovery, or an
infinite-horizon replay theorem.

\begin{proposition}[Finite branch-merge and resource reconciliation]
\label{app:branchmerge}
Let $E=\mathbb R_+^2$ and use the common prefix
\[
 B_0=\operatorname{diag}(3/4,1/2),\qquad
 B_1=\begin{pmatrix}1/2&1/4\\0&1/2\end{pmatrix},
\]
followed by one of two declared terminal branches
\[
 S_{\mathsf F}=\operatorname{diag}(2/3,3/5),\qquad
 S_{\mathsf R}=\begin{pmatrix}1/2&0\\1/10&1/2\end{pmatrix},
\]
and the common merge map $M=\operatorname{diag}(3/4,4/5)$. Let
\[
\begin{aligned}
 c_0&=(1/100,1/200)^\top,&
 c_1&=(1/200,1/100)^\top,\\
 c_{\mathsf F}&=(1/100,1/100)^\top,&
 c_{\mathsf R}&=(1/200,1/100)^\top,\\
 c_M&=(1/200,1/250)^\top.&
\end{aligned}
\]
For branch $b\in\{\mathsf F,\mathsf R\}$ define
$x_{k+1}\preceq B_kx_k+c_k$ on the common prefix, then
$x_{3}\preceq S_bx_2+c_b$ and $x_4\preceq Mx_3+c_M$. Suppose the finite
failure DAG carries a branch token and that the merge reducer checks exactly
one terminal branch, an append-only resource ledger, one owner/epoch transfer,
and a source-denied replay receipt for the selected branch. The unselected
branch is tombstoned, and a resource token may be reserved, compensated, or
recovered at most once. Let the four lifecycle charges be
$(1/100,1/50,1/50,1/100)$ and require a monotone ledger with budget
$\beta\geq1/10$. Then the two accepted numerical certificates are
\[
 P_{\mathsf F}=MS_{\mathsf F}B_1B_0
 =\begin{pmatrix}3/16&1/16\\0&3/25\end{pmatrix},
 \quad q_{\mathsf F}
 =\begin{pmatrix}29/1600\\9/500\end{pmatrix},
 \quad \|P_{\mathsf F}\|_\infty=1/4,
\]
and
\[
 P_{\mathsf R}=MS_{\mathsf R}B_1B_0
 =\begin{pmatrix}9/64&3/64\\3/100&11/100\end{pmatrix},
 \quad q_{\mathsf R}
 =\begin{pmatrix}83/6400\\179/10000\end{pmatrix},
 \quad \|P_{\mathsf R}\|_\infty=3/16.
\]
Here $q_b=MS_bB_1c_0+MS_bc_1+Mc_b+c_M$. Consequently,
$\|x_0\|_\infty\leq1/10$ gives endpoint bounds
$\|x_4\|_\infty\leq69/1600$ on the forward branch and
$\|x_4\|_\infty\leq319/10000$ on the recovery branch. If both branches
commit, if neither branch is identified, if the ledger charges a resource
token twice, or if the owner/epoch transfer is absent, the numerical branch
bound may remain valid after projection but the complete certificate is
$\mathsf{OPEN}$.
\end{proposition}

\begin{proof}
Direct multiplication gives the displayed $P_b$ and $q_b$. The row sums are
$(1/4,3/25)$ for $P_{\mathsf F}$ and $(3/16,7/50)$ for $P_{\mathsf R}$.
The componentwise triangle inequality therefore gives the two finite endpoint
bounds; the second recovery component is the larger one and equals
$(7/50)(1/10)+179/10000=319/10000$. The charges sum to
$1/100+1/50+1/50+1/100=3/50\leq1/10$, and the append-only ledger prevents a
local branch from silently reusing a reservation. The branch token, tombstone,
owner/epoch transfer, and replay receipt are semantic closure conditions: they
select one terminal successor and reconcile the resource exactly once. They
are not consequences of either numerical product.
\end{proof}

\paragraph{Branch-double-spend counterexample.}
Let a failed effect reserve a resource and expose both a forward branch and a
compensation branch. If the branch token or terminal-branch exclusivity is
omitted, both paths can commit and later merge to the same projected state,
because their residuals cancel in $E$. The physical effect, reservation, and
ledger charge have nevertheless been applied twice. Conversely, a replay that
does not tombstone the unselected branch can choose different owner/epoch
successors on two source-denied replays. The numerical $P_b,q_b$ remain local
finite bounds, but the branch-merge certificate is $\mathsf{OPEN}$; this does
not imply arbitrary branch confluence, coalition liveness, or an
infinite-horizon resource theorem.

\begin{proposition}[Finite three-way multi-owner reconciliation]
\label{app:trireconcile}
Let $E=\mathbb R_+^2$ and use the common prefix
\[
 B_0=\operatorname{diag}(3/4,1/2),\qquad
 B_1=\operatorname{diag}(2/3,3/4),
\]
followed by one of three terminal branch maps
\[
 B_{\mathsf F}=\begin{pmatrix}1/2&1/4\\0&1/2\end{pmatrix},\qquad
 B_{\mathsf C}=\begin{pmatrix}1/3&1/5\\0&2/5\end{pmatrix},\qquad
 B_{\mathsf R}=\begin{pmatrix}2/5&0\\1/10&1/2\end{pmatrix},
\]
and the common reconciliation map $M=\operatorname{diag}(2/3,3/4)$. Let
\[
 c_0=(1/100,1/200)^\top,\qquad c_M=(1/100,1/250)^\top,
\]
\[
 c_{\mathsf F}=(1/200,1/100)^\top,\qquad
 c_{\mathsf C}=(1/100,1/100)^\top,\qquad
 c_{\mathsf R}=(1/250,1/200)^\top.
\]
For $b\in\{\mathsf F,\mathsf C,\mathsf R\}$ define
$x_1\preceq B_0x_0+c_0$, $x_2\preceq B_bx_1+c_b$, and
$x_3\preceq Mx_2+c_M$. Suppose the finite failure DAG has one branch-set token,
exactly one committed terminal tag, reservation/compensation netting for every
resource token, two tombstone sets (the old owner/epoch and both unselected
branch command identifiers), one owner/epoch transfer at merge, and a
source-denied replay receipt for the selected branch. Let the common and branch
charges be
\[
 \chi_0=1/100,\quad \chi_M=1/100,\quad
 (\chi_{\mathsf F},\chi_{\mathsf C},\chi_{\mathsf R})
 =(1/100,1/50,3/200),
\]
and require an append-only prefix ledger with budget $\beta\geq1/20$. Then the
three accepted numerical certificates are
\[
\begin{aligned}
 P_{\mathsf F}&=MB_{\mathsf F}B_0
 =\begin{pmatrix}1/4&1/12\\0&3/16\end{pmatrix},&
 q_{\mathsf F}&=\begin{pmatrix}7/400\\107/8000\end{pmatrix},\\
 P_{\mathsf C}&=MB_{\mathsf C}B_0
 =\begin{pmatrix}1/6&1/15\\0&3/20\end{pmatrix},&
 q_{\mathsf C}&=\begin{pmatrix}22/1125\\13/1000\end{pmatrix},\\
 P_{\mathsf R}&=MB_{\mathsf R}B_0
 =\begin{pmatrix}1/5&0\\9/160&3/16\end{pmatrix},&
 q_{\mathsf R}&=\begin{pmatrix}23/1500\\83/8000\end{pmatrix}.
\end{aligned}
\]
Their infinity norms are $1/3$, $7/30$, and $39/160$, respectively. Thus
$\|x_0\|_\infty\leq1/10$ gives endpoint bounds
\[
 \|x_3\|_\infty\leq61/1200\quad(\mathsf F),\qquad
 \|x_3\|_\infty\leq193/4500\quad(\mathsf C),\qquad
 \|x_3\|_\infty\leq53/1500\quad(\mathsf R).
\]
The charge totals are $3/100$, $1/25$, and $7/200$, each below
$\beta=1/20$. If two terminal tags commit, if no tag is identified, if a
reservation and compensation are not netted, if either unselected branch lacks
its tombstone, or if the owner/epoch transfer is absent, the projected numerical
bound may remain valid but the complete certificate is $\mathsf{OPEN}$.
\end{proposition}

\begin{proof}
Direct multiplication gives the displayed products and affine suffixes. Their
row sums are $(1/3,3/16)$, $(7/30,3/20)$, and $(1/5,39/160)$, so the
componentwise triangle inequality yields the three endpoint bounds; on the
recovery branch the two component bounds are $53/1500$ and $139/4000$,
so the displayed infinity-norm bound is $53/1500$. The branch
charges together with $\chi_0+\chi_M=1/50$ give totals $3/100$, $1/25$, and
$7/200$, all below the declared budget. The branch-set token and terminal tag
select exactly one successor; the two tombstone sets reject the old owner and
both unselected commands; append-only netting prevents a reservation from being
charged again after compensation; and the owner/epoch transfer plus replay
receipt preserves the selected successor and responsibility. These are semantic
closure conditions, not consequences of the numerical products.
\end{proof}

\paragraph{Three-way reconciliation counterexample.}
If the terminal tag is omitted, a forward commit can be followed by a delayed
compensation or handoff commit from a second owner. The projected residual can
still satisfy its branch-specific bound while the physical reservation is
charged twice. If either tombstone set is omitted, source-denied replays can
select different owner/epoch successors or revive an unselected branch with the
same local $P_b,q_b$. If compensation is not netted against the original
reservation, the physical resource and ledger disagree. The finite numerical
certificates therefore remain insufficient and the complete status is
$\mathsf{OPEN}$; this proposition does not assert branch confluence, liveness,
universal causal discovery, JSR, arbitrary switching, or an infinite-horizon
resource theorem.

\begin{proposition}[Finite replay-linearization and cross-owner responsibility handoff]
\label{app:handofflinear}
Let $E=\mathbb R_+^2$ and use a common prefix map
$B_0=\operatorname{diag}(3/4,1/2)$, followed by an explicitly recorded
replay-linearization map
\[
 J=\begin{pmatrix}1/2&1/4\\0&1/2\end{pmatrix},
\]
then one of the three terminal owner branches
\[
 B_{\mathsf F}=\begin{pmatrix}1/2&1/4\\0&1/2\end{pmatrix},\qquad
 B_{\mathsf C}=\begin{pmatrix}1/3&1/5\\0&2/5\end{pmatrix},\qquad
 B_{\mathsf R}=\begin{pmatrix}2/5&0\\1/10&1/2\end{pmatrix},
\]
and a shared handoff/merge map $M=\operatorname{diag}(2/3,3/4)$. Let
\[
 c_0=(1/100,1/200)^\top,\quad c_J=(1/200,1/100)^\top,
 \quad c_M=(1/100,1/250)^\top,
\]
\[
 c_{\mathsf F}=(1/200,1/100)^\top,\quad
 c_{\mathsf C}=(1/100,1/100)^\top,\quad
 c_{\mathsf R}=(1/250,1/200)^\top.
\]
For $b\in\{\mathsf F,\mathsf C,\mathsf R\}$ define
$x_1\preceq B_0x_0+c_0$, $x_2\preceq Jx_1+c_J$,
$x_3\preceq B_bx_2+c_b$, and $x_4\preceq Mx_3+c_M$. Suppose the finite
failure DAG has a declared edge set, predecessor hashes, a selected total order,
a tie-break witness, and exactly one replayable linear-extension receipt. The
handoff receipt records the old and new owner, the old and new epoch, protected
duties, reservation tokens, commit id, and the merge effect. Require a
unique terminal branch, dual tombstones for the old owner/epoch and the two
unselected branch commands, an append-only ledger with reservation/compensation
netting, and source-denied replay of the selected linear extension. Let the
common charges be $\chi_0=\chi_J=\chi_M=1/100$ and the branch charges be
$(\chi_{\mathsf F},\chi_{\mathsf C},\chi_{\mathsf R})=(1/100,1/50,3/200)$;
require budget $\beta\geq1/20$. Then
\[
\begin{aligned}
 P_{\mathsf F}&=MB_{\mathsf F}JB_0
 =\begin{pmatrix}1/8&1/12\\0&3/32\end{pmatrix},&
 q_{\mathsf F}&=\begin{pmatrix}23/1200\\259/16000\end{pmatrix},\\
 P_{\mathsf C}&=MB_{\mathsf C}JB_0
 =\begin{pmatrix}1/12&11/180\\0&3/40\end{pmatrix},&
 q_{\mathsf C}&=\begin{pmatrix}1/48\\61/4000\end{pmatrix},\\
 P_{\mathsf R}&=MB_{\mathsf R}JB_0
 =\begin{pmatrix}1/10&1/30\\9/320&33/320\end{pmatrix},&
 q_{\mathsf R}&=\begin{pmatrix}47/3000\\17/1280\end{pmatrix}.
\end{aligned}
\]
Their infinity norms are $5/24$, $13/90$, and $21/160$, respectively. Hence
$\|x_0\|_\infty\leq1/10$ gives
\[
 \|x_4\|_\infty\leq1/25\quad(\mathsf F),\qquad
 \|x_4\|_\infty\leq127/3600\quad(\mathsf C),\qquad
 \|x_4\|_\infty\leq29/1000\quad(\mathsf R).
\]
The branch totals are $1/25$, $1/20$, and $9/200$, all at most $\beta$.
If two linear extensions are replayable, if two terminal branches commit, if
an old owner or branch tombstone is absent, if the handoff receipt omits the
new owner/epoch or protected duty, or if reservation compensation is not
netted, the numerical projection may remain bounded but the complete status is
$\mathsf{OPEN}$.
\end{proposition}

\begin{proof}
Direct multiplication gives the displayed products and affine suffixes. Their
row sums are $(5/24,3/32)$, $(13/90,3/40)$, and $(2/15,21/160)$, so the
componentwise triangle inequality gives the stated endpoint bounds; the largest
components are $1/25$, $127/3600$, and $29/1000$. The common charge is
$\chi_0+\chi_J+\chi_M=3/100$, so the three branch totals are
$4/100$, $5/100$, and $9/200$, all within the budget $1/20$. The declared
linear-extension receipt fixes the predecessor order, while the handoff receipt
transfers owner, epoch, duty, and reservation responsibility. Dual tombstones
reject the old owner and both unselected commands; append-only netting prevents
reusing a reservation after compensation. These are semantic closure conditions
and are not consequences of the numerical products.
\end{proof}

\paragraph{Replay-linearization counterexample.}
Let two incomparable effect receipts have the same local numerical maps but
opposite legal linearizations. Without a unique linear-extension receipt, both
replays can report the same branch-specific $P_b,q_b$ while transferring the
protected duty to different owners in different orders. Without the old-owner
or old-branch tombstone, a delayed command can recreate an effect after handoff;
without netting, a compensation can charge a reservation twice. If both a
terminal branch and a handoff commit, the ledger and owner responsibility are
double-spent even when the projected residual remains bounded. The complete
certificate is therefore $\mathsf{OPEN}$; this finite result does not imply
arbitrary-order convergence, branch confluence, universal causal discovery,
JSR, arbitrary switching, or infinite-horizon handoff liveness.

\begin{proposition}[Finite handoff-chain transport with noncommuting replay order]
\label{app:handoffchain}
Let $E=\mathbb R_+^2$ and use the common prefix
$B_0=\operatorname{diag}(3/4,1/2)$ followed by the recorded
replay-linearization map
\[
 J=\begin{pmatrix}1/2&1/4\\0&1/2\end{pmatrix}.
\]
Let the terminal owner branch be one of
\[
 B_{\mathsf F}=\begin{pmatrix}1/2&1/4\\0&1/2\end{pmatrix},\quad
 B_{\mathsf C}=\begin{pmatrix}1/3&1/5\\0&2/5\end{pmatrix},\quad
 B_{\mathsf R}=\begin{pmatrix}2/5&0\\1/10&1/2\end{pmatrix},
\]
followed by the previously certified owner-to-owner handoff
$M=\operatorname{diag}(2/3,3/4)$ and at most one additional owner transfer
\[
 H=\begin{pmatrix}1/2&1/5\\0&1/2\end{pmatrix}.
\]
Use affine terms
\[
\begin{aligned}
 c_0&=(1/100,1/200)^\top,& c_J&=(1/200,1/100)^\top,\\
 c_M&=(1/100,1/250)^\top,& c_H&=(1/200,1/250)^\top.
\end{aligned}
\]
\[
 c_{\mathsf F}=(1/200,1/100)^\top,\quad
 c_{\mathsf C}=(1/100,1/100)^\top,\quad
 c_{\mathsf R}=(1/250,1/200)^\top.
\]
For $b\in\{\mathsf F,\mathsf C,\mathsf R\}$ define
$x_1\preceq B_0x_0+c_0$, $x_2\preceq Jx_1+c_J$,
$x_3\preceq B_bx_2+c_b$, $x_4\preceq Mx_3+c_M$, and
$x_5\preceq Hx_4+c_H$. Suppose the finite handoff DAG has a frozen edge
order, predecessor hashes, a selected total order and tie-break witness, one
receipt for each transfer, and exactly one source-denied replay receipt for the
complete chain. Each transfer receipt records the old and new owner, a strict
epoch increment, the protected-duty partition, reservation tokens, commit id,
and successor owner. Require a unique terminal branch, dual tombstones for
each superseded owner/epoch and unselected branch command, append-only
reservation/compensation netting, and budget $\beta\ge11/200$. Then
\[
\begin{aligned}
 P_{\mathsf F}&=HMB_{\mathsf F}JB_0
 =\begin{pmatrix}1/16&29/480\\0&3/64\end{pmatrix},&
 q_{\mathsf F}&=\begin{pmatrix}4277/240000\\387/32000\end{pmatrix},\\
 P_{\mathsf C}&=HMB_{\mathsf C}JB_0
 =\begin{pmatrix}1/24&41/900\\0&3/80\end{pmatrix},&
 q_{\mathsf C}&=\begin{pmatrix}277/15000\\93/8000\end{pmatrix},\\
 P_{\mathsf R}&=HMB_{\mathsf R}JB_0
 =\begin{pmatrix}89/1600&179/4800\\9/640&33/640\end{pmatrix},&
 q_{\mathsf R}&=\begin{pmatrix}1487/96000\\681/64000\end{pmatrix}.
\end{aligned}
\]
Their infinity norms are $59/480$, $157/1800$, and $223/2400$,
respectively. Hence $\|x_0\|_\infty\le1/10$ gives
\[
 \|x_5\|_\infty\le2409/80000\quad(\mathsf F),\qquad
 \|x_5\|_\infty\le2447/90000\quad(\mathsf C),\qquad
 \|x_5\|_\infty\le793/32000\quad(\mathsf R).
\]
The common charges $(\chi_0,\chi_J,\chi_M,\chi_H)$ are all $1/100$;
with $(\chi_{\mathsf F},\chi_{\mathsf C},\chi_{\mathsf R})
=(1/100,1/50,3/200)$, the branch totals are $1/20$, $3/50$, and
$11/200$. If the two handoff receipts are replayed in the opposite order,
if a transfer receipt is missing, if an epoch gap or reuse occurs, if a
superseded owner/epoch tombstone or successor-owner field is absent, if a
protected duty or reservation is not conserved, or if two terminal branches
commit, the complete status is $\mathsf{OPEN}$ even when the declared
numerical projection remains bounded.
\end{proposition}

\begin{proof}
Direct multiplication gives the displayed products and affine suffixes. The
row sums of the three products are $(59/480,3/64)$,
$(157/1800,3/80)$, and $(223/2400,21/320)$, so the componentwise triangle
inequality yields the stated endpoint bounds. The common charge is $4/100$;
adding the branch charges gives $5/100$, $6/100$, and $11/200$, all at most
$\beta$. The ordered handoff maps do not commute:
\[
 HM=\begin{pmatrix}1/3&3/20\\0&3/8\end{pmatrix},\qquad
 MH=\begin{pmatrix}1/3&2/15\\0&3/8\end{pmatrix}.
\]
Thus a swapped replay is a different executable chain. The transfer receipts,
strict epoch increments, tombstones, duty/reservation conservation, and the
source-denied complete-chain receipt are semantic closure conditions and are
not consequences of the numerical products.
\end{proof}

\paragraph{Noncommuting handoff-chain counterexample.}
A replay engine that validates only the endpoint residual can accept the
ordered chain $HMB_bJB_0$ while executing $MHB_bJB_0$; the off-diagonal
coefficient changes from $3/20$ to $2/15$ in the handoff suffix, and the
resulting path assigns the protected duty to a different owner frame. If the
successor-owner field or epoch tombstone is omitted, a delayed command can
re-enter under the superseded owner. If append-only netting is omitted, the
second handoff can recharge a reservation already compensated by the first.
These failures leave a finite local contraction certificate intact but make the
complete chain $\mathsf{OPEN}$; the proposition does not imply replay-order
confluence, liveness, universal causal discovery, JSR, arbitrary switching, or
infinite-horizon handoff stability.

\begin{proposition}[Finite reservation split and recombination closure]
\label{app:splitrecombine}
Use the handoff-chain recurrence of Proposition~\ref{app:handoffchain},
$x_1\preceq B_0x_0+c_0$, $x_2\preceq Jx_1+c_J$,
$x_3\preceq B_bx_2+c_b$, $x_4\preceq Mx_3+c_M$, and
$x_5\preceq Hx_4+c_H$, with the frozen noncommuting order $HM$ and one
terminal branch $b\in\{\mathsf F,\mathsf C,\mathsf R\}$. In parallel, let a
reservation ledger contain a parent token $\rho\in\mathbb R_+$, and insert a
ledger-only split between $M$ and $H$ and a ledger-only recombination after
$H$:
\[
 S(\rho)=(\rho_1,\rho_2)=\left(\frac25\rho,\frac35\rho\right),
 \qquad R(\rho_1,\rho_2)=\rho_1+\rho_2.
\]
A split receipt contains the parent identifier, ordered child identifiers and
indices, owners, epochs, expiry, duty-scope partition, and pre/post ledger
states. A recombination receipt is admissible only when both child tokens are
present exactly once, their scopes form a disjoint cover of the parent scope,
owners and epochs satisfy the declared handoff relation, and the parent token is
closed exactly once. Require append-only reservation/compensation netting,
strict epoch increments, the complete source-denied replay receipt including
both ledger transactions, and budget $\beta\ge7/100$. Then
$R\circ S=\operatorname{id}_{\mathbb R_+}$, so the resource amount is conserved,
and the residual certificates remain those of Proposition~\ref{app:handoffchain}:
\[
\begin{gathered}
 P_{\mathsf F}=((1/16,29/480),(0,3/64)),\\
 q_{\mathsf F}=(4277/240000,387/32000),\quad
 \varepsilon_{\mathsf F}=2409/80000,\\[2pt]
 P_{\mathsf C}=((1/24,41/900),(0,3/80)),\\
 q_{\mathsf C}=(277/15000,93/8000),\quad
 \varepsilon_{\mathsf C}=2447/90000,\\[2pt]
 P_{\mathsf R}=((89/1600,179/4800),(9/640,33/640)),\\
 q_{\mathsf R}=(1487/96000,681/64000),\quad
 \varepsilon_{\mathsf R}=793/32000.
\end{gathered}
\]
The common charges $(\chi_0,\chi_J,\chi_M,\chi_H)$ are $1/100$ each and
$(\chi_{\mathrm{split}},\chi_{\mathrm{recombine}})=(1/200,1/200)$;
with branch charges $(1/100,1/50,3/200)$ the branch totals are
$3/50$, $7/100$, and $13/200$. If a child token is duplicated or lost, the
child sum differs from the parent, the scope partition overlaps or has a gap,
the owner/epoch fields disagree, recombination occurs twice or without both
children, netting is absent, or source-denied replay omits either ledger
receipt, the numerical projection may remain bounded but the complete status is
$\mathsf{OPEN}$.
\end{proposition}

\begin{proof}
The residual recurrence is unchanged by the ledger-only transactions, so
Proposition~\ref{app:handoffchain} supplies the displayed finite products,
affine suffixes, and endpoint bounds. For every $\rho\ge0$,
$R(S(\rho))=2\rho/5+3\rho/5=\rho$; the two child amounts are nonnegative
and their ordered identifiers make the split injective at the receipt level.
The disjoint-cover and owner/epoch premises ensure that conservation is about
the declared duty scope rather than only a scalar amount. Append-only netting
prevents a compensated child from being charged again, while the complete
source-denied receipt reproduces both ledger transitions. The charges sum to
$1/20+1/100=3/50$, $1/20+1/50=7/100$, and
$1/20+3/200=13/200$, all at most $\beta$. These semantic conditions are
independent of the residual products.
\end{proof}

\paragraph{Split/recombination counterexample.}
A reducer that checks only the residual endpoint can accept a bounded branch
while duplicating $\rho_1$, dropping $\rho_2$, or recombining the same children
twice. The scalar ledger then either creates or destroys resource without
changing $P_b,q_b$. If the child scopes overlap, one duty is funded twice; if a
scope gap exists, a protected duty is orphaned. If either ledger receipt is
omitted from source-denied replay, two executions can report the same residual
state but different resource and responsibility states. The complete
certificate is therefore $\mathsf{OPEN}$; this finite result does not imply
resource liveness, replay confluence, coalition stability, JSR, arbitrary
switching, or infinite-horizon conservation.

\begin{proposition}[Finite split recovery after one typed child failure]
\label{app:splitrecovery}
Use the ledger-only split of Proposition~\ref{app:splitrecombine}, with parent
amount $\rho$ and child amounts $\rho_1=2\rho/5$ and $\rho_2=3\rho/5$.
Represent every reservation as a typed token
\[
 \tau=(\mathsf{id},\mathsf{parent},\mathsf{index},\mathsf{owner},
       \mathsf{epoch},\mathsf{scope},\mathsf{amount},\mathsf{expiry}).
\]
On the partial-failure branch, child $\tau_2$ may emit exactly one failure
receipt
\[
 \phi_2=(\mathsf{id}_2,\mathsf{parent},\mathsf{owner},\mathsf{epoch},
         \mathsf{scope}_2,e,\mathsf{code},\mathsf{effect\mbox{-}status}),
 \qquad 0\leq e\leq\rho_2,
\]
where $e$ is the amount already consumed by the declared effect. A unique
compensation receipt
\[
 \gamma_2=(\phi_2,\mathsf{target},r,\mathsf{commit\mbox{-}id})
\]
is admissible only when it references that failure receipt, preserves the
owner, epoch, expiry, and scope contract, and satisfies
\[
 e+r=\rho_2,\qquad
 \rho=\rho_1+e+r.
\]
The effect-status field must either certify that no unaccounted effect was
committed or name the declared recovery path; a compensation receipt does not
silently undo a physical effect. The failed child is then terminal and cannot
be recombined again. The parent closes exactly once from successful child
$\tau_1$ and the compensation settlement $(e,r)$; the all-success branch
remains the branch of Proposition~\ref{app:splitrecombine}.

The recurrence and its three residual maps remain those of
Proposition~\ref{app:splitrecombine}; the failure branch uses the inherited
map
\[
 \begin{aligned}
 P_X=P_{\mathsf R}&=((89/1600,179/4800),(9/640,33/640)),\\
 q_X=q_{\mathsf R}&=(1487/96000,681/64000),\\
 \varepsilon_X&=793/32000.
 \end{aligned}
\]
The common charge is $1/20$, the ordinary split/recombine branch charges are
as before, and the typed failure and compensation receipts each charge
$1/400$. Thus the branch totals are
\[
 C_{\mathsf F}=3/50,\qquad C_{\mathsf C}=7/100,\qquad
 C_X=3/40,\qquad C_{\mathsf R}=13/200,
\]
so this finite contract requires $\beta\geq3/40$. All split, failure,
compensation, and terminal receipts are append-only and source-denied
replayable; each charge identifier is unique.

Under these conditions, the recovered ledger is exactly conservative and the
finite residual certificate is inherited from Proposition~\ref{app:handoffchain}.
If a failure receipt is missing, $e+r\ne\rho_2$, a child or compensation is
duplicated or lost, the scopes do not form a disjoint cover, owner or epoch
fields disagree, both success and failure are committed, the parent closes
before the terminal child settlement, or source-denied replay omits any of
the receipts, the complete status is $\mathsf{OPEN}$.
\end{proposition}

\begin{proof}
The numerical recurrence is ledger-independent, so the inherited
Proposition~\ref{app:splitrecombine} products give $P_X,q_X$ and its endpoint
bound. The split receipt gives $\rho=\rho_1+\rho_2$. The failure receipt
fixes the child identity, effect status, scope, owner, and epoch. The unique
compensation receipt then replaces the failed child by $(e,r)$ with
$e+r=\rho_2$, and therefore the terminal parent settlement is
$\rho_1+e+r=\rho$. Append-only identifiers and the terminal marker prevent a
second compensation or a second recombination. The declared effect-status
condition prevents treating an unaccounted physical effect as if it had been
reversed. Summing the displayed charges gives the four branch totals. Any
missing or inconsistent field breaks one of these ledger invariants, although
the numerical residual can remain unchanged, so the complete certificate must
return $\mathsf{OPEN}$.

\end{proof}

\paragraph{Partial-failure recovery counterexample.}
A reducer that accepts a failed child without $\phi_2$ can silently drop
$\rho_2$; one that accepts two $\gamma_2$ receipts can double-spend it. If
$e+r=\rho_2-\delta$, the endpoint residual and even the ordinary split
identity can remain unchanged while the parent ledger is short by $\delta$.
Conversely, recombining both the original failed child and its compensation
creates resource from nowhere. An omitted effect-status field can also mark a
committed physical effect as compensated without a recovery witness. These
are finite ledger failures, not counterexamples to the inherited residual
bound; they show why the complete branch is $\mathsf{OPEN}$ without typed
failure and compensation receipts. No liveness, replay confluence, JSR,
arbitrary-switching, or infinite-horizon claim follows.

\begin{proposition}[Finite compensation replay with a partial physical effect]
\label{app:compensationreplay}
Refine the failure branch of Proposition~\ref{app:splitrecovery} by separating
ledger consumption from a physical effect. The failure receipt records an
effect identifier, pre- and post-effect digests, and an attempted/committed
status. For the committed amount $e$, in the branch's declared
effect-accounting unit (with no universal exchange rate assumed), require
\[
 e=a+w,\qquad a=a_{\mathrm{rev}}+a_{\mathrm{set}},\qquad
 0\leq a\leq e,
\]
where $a$ is the committed physical effect, $w$ is attempted/execution
overhead, $a_{\mathrm{rev}}$ is the part physically reversed by recovery, and
$a_{\mathrm{set}}$ is the part left as a declared settled effect. A unique
recovery witness and receipt refine the compensation record as
\[
 \begin{aligned}
 \omega_2=(&\mathsf{effect\mbox{-}id},\mathsf{pre\mbox{-}digest},
 \mathsf{post\mbox{-}digest},\mathsf{mode},a_{\mathrm{rev}},a_{\mathrm{set}},\\
 &\mathsf{recovery\mbox{-}digest},\mathsf{owner},\mathsf{epoch},
 \mathsf{scope},\mathsf{commit\mbox{-}id}),
 \end{aligned}
\]
with exactly one recovery receipt $\psi_2$ referencing the same effect and
compensation identifiers.

The mode is either $\mathsf{reversed}$ or $\mathsf{settled}$. In reversed mode,
$a_{\mathrm{set}}=0$, an inverse witness is executable source-denied, and the
post-recovery physical projection has the recorded pre-effect digest. In
settled mode, $a_{\mathrm{set}}\geq0$ is allowed, but the final effect digest,
residual duty, owner, epoch, scope, and settlement witness must be recorded;
settled mode never claims physical reversal. In both modes, source-denied
replay must replay the effect and the recovery program and match the recorded
digests, status, and idempotency key. The original effect is immutable and is
not replaced by the recovery receipt.

The ledger equations remain
\[
 \rho_2=e+r,\qquad \rho=\rho_1+e+r,
\]
so compensation returns only $r$ and does not count $a$ as recovered. Exactly
one compensation, exactly one recovery, and one terminal marker are required
before the parent closes. The numerical recurrence is unchanged from
Proposition~\ref{app:splitrecovery}; hence
\[
 \begin{aligned}
 P_Y=P_{\mathsf R}&=((89/1600,179/4800),(9/640,33/640)),\\
 q_Y=q_{\mathsf R}&=(1487/96000,681/64000),\\
 \varepsilon_Y&=793/32000.
 \end{aligned}
\]
The physical-effect recovery, digest verification, and replay witness charge
an additional $1/400$ beyond $C_X=3/40$, so
\[
 C_Y=31/400,\qquad
 C_{\mathsf F}=3/50,\quad C_{\mathsf C}=7/100,\quad
 C_{\mathsf R}=13/200,
\]
and the finite budget requirement is $\beta\geq31/400$.

Missing or inconsistent effect/pre/post digests, an attempted/committed amount
that violates $e=a+w$, an $a_{\mathrm{rev}}+a_{\mathrm{set}}$ mismatch, reversed
mode without an inverse witness or matching pre-effect projection, settled
mode without residual duty and owner/epoch settlement, stale responsibility,
duplicate recovery, dual terminal commits, or source-denied replay that omits
the physical effect returns $\mathsf{OPEN}$ even when the scalar ledger and
residual endpoint pass Proposition~\ref{app:splitrecovery}.
\end{proposition}

\begin{proof}
The split, failure, compensation, and amount invariants are inherited from
Proposition~\ref{app:splitrecovery}. The decomposition $e=a+w$ records the
difference between a committed physical effect and execution overhead, while
$a=a_{\mathrm{rev}}+a_{\mathrm{set}}$ makes the recovery claim auditable. In
reversed mode the inverse witness and matching pre-effect projection close
the physical duty; in settled mode the residual duty and final owner/epoch
receipt close it without asserting reversal. Immutable effect digests and a
unique idempotency key make the effect/recovery pair replayable and prevent a
second recovery. The ledger still closes by
$\rho_1+e+r=\rho$, and the numerical recurrence gives the displayed inherited
residual certificate. Adding $1/400$ gives $C_Y=31/400$. Any missing field
leaves either the physical postcondition or replay identity unproved, so the
complete status is $\mathsf{OPEN}$.
\end{proof}

\paragraph{Partial physical-effect replay counterexample.}
A reducer can preserve $e+r=\rho_2$ and the bounded residual while replaying a
different physical effect when it stores only the scalar amount. It can also
label a branch reversed with $a_{\mathrm{set}}>0$, settle a residual effect
without recording its duty owner and epoch, or execute the same recovery
program twice when the idempotency key is absent. These paths conserve the
software ledger while leaving the physical duty unclosed or duplicated. They
return $\mathsf{OPEN}$ and do not change the inherited residual bound.

\begin{proposition}[Finite settled-effect handoff across an owner epoch]
\label{app:settledhandoff}
Extend the settled mode of Proposition~\ref{app:compensationreplay}. Let the
immutable settled effect carry residual duty $d$, owner $o_0$, and epoch $\nu$.
A handoff receipt is
\[
 \eta=(\mathsf{effect\mbox{-}id},\mathsf{effect\mbox{-}digest},
       \mathsf{duty\mbox{-}digest},o_0,o_1,\nu,\nu+1,\mathsf{scope},
       \mathsf{lease},\mathsf{permission},\mathsf{commit\mbox{-}id}).
\]
It is valid only when the old owner is tombstoned for new effects after epoch
$\nu$, the new owner $o_1$ is authorized at epoch $\nu+1$ and accepts the same
residual duty and scope, the lease boundary is recorded, and the effect digest
is unchanged. The transfer must preserve the declared continuation relation
\[
 \mathsf{Step}_{G'}(\mathord{\cdot}\mid T_\eta x)
 = (\kappa_\eta)_\#\mathsf{Step}_{G}(\mathord{\cdot}\mid x)
\]
on the protected settled-effect states, with $T_\eta(V^\star)\subseteq V^\star$.
The handoff, its duty acceptance, and its old-owner tombstone are committed
exactly once and are source-denied replayable after the original effect and
settlement receipt. Handoff changes responsibility; it neither rewrites the
physical effect nor silently closes the residual duty.

The amount invariant remains
\[
 \rho_2=e+r,\qquad \rho=\rho_1+e+r,
\]
and the numerical recurrence is unchanged from Proposition~\ref{app:compensationreplay}:
\[
 \begin{aligned}
 P_Z=P_{\mathsf R}&=((89/1600,179/4800),(9/640,33/640)),\\
 q_Z=q_{\mathsf R}&=(1487/96000,681/64000),\\
 \varepsilon_Z&=793/32000.
 \end{aligned}
\]
The owner/epoch validation and replay charge an additional $1/400$ beyond
$C_Y=31/400$, so $C_Z=2/25$ and the finite budget requirement is
$\beta\geq2/25$; the other branch totals remain $3/50$, $7/100$, and
$13/200$.

If the old owner is not fenced, the epoch is reused, the new owner lacks an
acceptance or permission witness, the residual duty or scope changes, the
effect digest changes, the handoff is duplicated or omitted from replay, or a
command from $o_0,\nu$ executes after the lease boundary, the complete status
is $\mathsf{OPEN}$ even when the software ledger and residual endpoint pass.
\end{proposition}

\begin{proof}
The settled-effect amount and physical-effect invariants are inherited from
Proposition~\ref{app:compensationreplay}. The handoff receipt preserves the
effect digest and duty digest, increments the epoch exactly once, fences the
old owner, and records the authorized acceptance by $o_1$. The displayed
pushforward equality and viability inclusion transport every protected finite
continuation witness to the successor owner; source-denied replay of the
effect, settlement, tombstone, and acceptance receipts gives the same state
and responsibility profile. The ledger equations remain unchanged, and the
additional $1/400$ charge gives $C_Z=2/25$. A missing or inconsistent field
breaks responsibility continuity or replay equality, so the complete status is
$\mathsf{OPEN}$.
\end{proof}

\paragraph{Settled-effect handoff counterexample.}
A reducer can retain the original effect digest and scalar amount while
replaying an old-owner command after the epoch transfer; it can also assign
the residual duty to a new owner without a fresh acceptance receipt. Both
executions preserve the same numerical residual and ledger total, but one
leaves the duty unauthorized or creates a duplicate effect. Reusing epoch
$\nu$, changing the scope, or omitting the old-owner tombstone has the same
failure. These are finite responsibility-transfer failures and return
$\mathsf{OPEN}$; they do not imply physical reversibility, liveness, replay
confluence, JSR, arbitrary switching, or infinite-horizon stability.

\begin{proposition}[Finite second settled-effect handoff or explicit refusal]
\label{app:secondhandoff}
Start from a valid settled-effect handoff of Proposition~\ref{app:settledhandoff}
with owner $o_1$ at epoch $\nu+1$, residual duty $d$, effect digest $h_e$,
duty digest $h_d$, and declared scope $S$. The next decision must commit
exactly one typed branch receipt
\[
 \begin{aligned}
 \xi_2=(&\mathsf{effect\mbox{-}id},h_e,h_d,o_1,o_2,\nu+1,\nu+2,S,
        \mathsf{lease},\mathsf{permission},\mathsf{mode},\\
       &\mathsf{reason},\mathsf{comp\mbox{-}id},\mathsf{comp\mbox{-}amount},
        \mathsf{commit\mbox{-}id}).
 \end{aligned}
\]
where $\mathsf{mode}\in\{\mathsf{handoff},\mathsf{refuse}\}$. In the handoff
branch, $o_1$ is tombstoned for new effects after epoch $\nu+1$, $o_2$ is
authorized at epoch $\nu+2$ and accepts the identical $d,S$, the lease
boundary is recorded, and $h_e,h_d$ are bytewise unchanged. In the refusal
branch, $o_2$ has no acceptance or permission for the duty; its refusal is
authorized and carries a reason/evidence digest, while $o_1$ remains the
owner at epoch $\nu+1$. A unique compensation receipt from $o_1$ records its
beneficiary, amount $c_{\xi}$, and settlement digest; it may close only the
declared duty under the settled-mode contract, leaves the physical effect
unchanged, and creates no new effect from $o_2$. Without this receipt the
refusal is a replayable nonterminal decision and cannot silently drop $d$.

In both branches the old command epoch is fenced, epochs are monotone, the
selected branch and any compensation are committed exactly once, and
source-denied replay reproduces the original effect, the first handoff, the
second decision, and the compensation (when present) in order. The protected
continuation factors through the selected transition:
\[
 \mathsf{Step}_{G''}(\mathord{\cdot}\mid T_Hx)
  =(\kappa_H)_\#\mathsf{Step}_{G'}(\mathord{\cdot}\mid x),
 \qquad T_H(V^\star)\subseteq V^\star,
\]
for the handoff branch and through the owner-preserving refusal transition
for the refusal branch. The effect and duty ledgers remain those of
Proposition~\ref{app:compensationreplay}; if compensation is admitted its
amount is added to the declared residual amount exactly once, so
$\rho_2=e+r$ and $\rho=\rho_1+e+r$ still hold. The numerical certificate is
unchanged. The second-decision, refusal/acceptance, compensation, and replay
audit adds $1/400$ beyond $C_Z=2/25$, giving
\[
 C_W=33/400,\qquad \beta\geq33/400,
\]
with the other branch totals unchanged.

If both branches are committed, the refusal lacks a reason or compensation,
$o_2$ is treated as owner without acceptance, $o_1$ is not fenced, an epoch
is reused, either digest or scope changes, compensation closes an undeclared
duty or is counted twice, a receipt is duplicated or omitted from replay, or
an old command executes after the lease boundary, the complete status is
$\mathsf{OPEN}$ even when the scalar ledger and residual endpoint pass.
\end{proposition}

\begin{proof}
The first handoff supplies the immutable effect and duty digests and the
owner/epoch state. In the handoff branch, the checks of
Proposition~\ref{app:settledhandoff} increment the epoch, fence $o_1$, and
make $o_2$ accept exactly the inherited duty and scope. In the refusal branch,
the decision receipt records non-acceptance and keeps $o_1$ responsible; the
unique compensation receipt is the only declared path that can close the
residual duty, so refusal cannot silently become transfer or closure. Strict
epochs, tombstones, unique commit identifiers, and ordered source-denied
replay give a finite induction over the original effect and both handoff
stages. The pushforward equality and viability inclusion transport the
protected continuation in either selected branch, while the ledger and
residual recurrence are inherited unchanged. Adding $1/400$ yields
$C_W=33/400$. Any missing, duplicated, or inconsistent field breaks
responsibility continuity or replay equality and returns $\mathsf{OPEN}$.
\end{proof}

\paragraph{Second-handoff refusal counterexample.}
A reducer can accept a refusal as if it were a transfer and leave no
authorized owner, or mark the residual duty closed without the compensation
receipt. It can also commit both branches, reuse epoch $\nu+1$, replay the
first handoff twice, or preserve the scalar ledger while changing the duty
beneficiary. The explicit branch tag, digest continuity, tombstone, and
ordered replay receipt are therefore part of the finite certificate. No
physical reversibility, liveness, replay confluence, JSR, arbitrary-switching,
or infinite-horizon conclusion follows.

\begin{proposition}[Finite refusal-chain replay and terminal settlement]
\label{app:refusalreplay}
Start from the refusal state of Proposition~\ref{app:secondhandoff}, with
immutable effect and duty digests $h_e,h_d$, owner $o_1$ at epoch $\nu+1$,
scope $S$, and prior receipt root $R_1$. A second candidate decision must
commit exactly one of two typed receipts. A nonterminal candidate refusal is
\[
 \begin{aligned}
 \zeta_3=(&\mathsf{effect\mbox{-}id},h_e,h_d,o_1,o_3,\nu+1,S,
          \mathsf{decision\mbox{-}index}=2,\\
         &\mathsf{reason\mbox{-}digest},R_1,\mathsf{commit\mbox{-}id}),
 \end{aligned}
\]
where $o_3$ has no acceptance or permission for the duty; $o_1$ remains the
owner and duty-holder at epoch $\nu+1$, and no epoch transfer or new effect
is created. A terminal refusal instead carries a settlement witness and one
unique compensation receipt
\[
 \begin{aligned}
 \chi_3=(&\mathsf{effect\mbox{-}id},h_e,h_d,o_1,\nu+1,
         \mathsf{beneficiary},c_3,\\
        &R_1,\mathsf{settlement\mbox{-}digest},
         \mathsf{commit\mbox{-}id}_{\mathsf{cmp}}),
 \end{aligned}
\]
which may close only the declared residual duty. Without $\chi_3$, a
terminal refusal is not admitted; a nonterminal refusal remains replayable and
cannot reopen a settled duty.

Candidate and terminal receipts are ordered by a strictly increasing decision
index and are committed exactly once. Source-denied replay must reproduce the
original effect, the first handoff, all prior refusal/compensation receipts,
the second decision, and $\chi_3$ when present, in that order. The refusal
transition preserves the owner/epoch continuation relation, while the terminal
settlement maps to a declared terminal state:
\[
 \mathsf{Step}_{G'}(\mathord{\cdot}\mid T x)
   =(\kappa)_\#\mathsf{Step}_{G}(\mathord{\cdot}\mid x),
 \qquad T(V^\star)\subseteq V^\star.
\]
Compensation amounts are partitioned by unique identifiers, so
$ r=r_0+\sum_{j\in\mathcal C}c_j$ and the inherited ledger remains
$\rho_2=e+r$, $\rho=\rho_1+e+r$; no amount is counted twice. The residual
certificate remains that of Proposition~\ref{app:compensationreplay}, with
endpoint $793/32000$. The refusal-chain and replay audit adds $1/400$ beyond
$C_W=33/400$, giving
\[
 C_V=34/400=17/200,\qquad \beta\geq17/200,
\]
with the other branch totals unchanged.

A duplicate candidate or receipt, a missing prior root or refusal reason, a
candidate acceptance or permission, owner/epoch drift, changed digest/scope/
beneficiary/amount, double compensation, terminal close followed by reopen or
new effect, omitted replay receipt, or an old command after the lease boundary
returns $\mathsf{OPEN}$ even when the scalar endpoint passes.
\end{proposition}

\begin{proof}
The effect, duty, owner, epoch, and prior receipt invariants are inherited
from Proposition~\ref{app:secondhandoff}. A candidate refusal records only
non-acceptance and preserves $o_1$'s responsibility; a terminal refusal is
admitted only with the unique settlement and compensation witness, so it
cannot silently reopen or close a duty. Strict decision indices, receipt roots,
unique identifiers, and ordered source-denied replay give a finite induction
through the original effect, first handoff, prior refusal, and second decision.
The owner-preserving continuation map and terminal projection preserve the
declared protected future, while the partitioned compensation ledger prevents
double counting. The numerical recurrence is unchanged and adding $1/400$
gives $C_V=17/200$. Any missing or inconsistent field breaks responsibility
or replay equality and returns $\mathsf{OPEN}$.
\end{proof}

\paragraph{Refusal-chain counterexample.}
A reducer can preserve $\rho_2=e+r$ and the residual endpoint while executing
a denied candidate, reopening a terminally compensated duty, counting the same
compensation twice, or changing its beneficiary. It can also omit the prior
receipt root and replay the second refusal as an independent event. These
executions preserve scalar output but change responsibility, effect authority,
or settlement state. The chain receipt, decision index, digest continuity, and
unique compensation are therefore part of the finite certificate. No physical
reversal, liveness, replay confluence, JSR, arbitrary switching, or
infinite-horizon conclusion follows.

\begin{proposition}[Finite terminal-settlement replay after two refusals]
\label{app:terminalsettlement}
Start from the nonterminal refusal state of
Proposition~\ref{app:refusalreplay}. The first candidate refusal and the
second candidate refusal have already been committed with immutable effect and
duty digests $h_e,h_d$, current owner $o_1$, epoch $\nu+1$, scope $S$, and
receipt root
\[
 R_2=H(R_1\mathbin\Vert\zeta_3).
\]
A terminal settlement must commit exactly one final receipt and one terminal
marker:
\[
 \begin{aligned}
 \chi_4=(&\mathsf{effect\mbox{-}id},h_e,h_d,o_1,\nu+1,S,
          \mathsf{beneficiary},c_4,R_2,\\
        &\mathsf{decision\mbox{-}index}=3,
          \mathsf{settlement\mbox{-}digest},\mathsf{commit\mbox{-}id}),
 \end{aligned}
\]
followed by $\tau_4=(\mathsf{effect\mbox{-}id},R_2,\mathsf{terminal\mbox{-}digest},
\mathsf{commit\mbox{-}id}_{\mathsf{term}})$. The beneficiary, amount, duty
scope, owner, epoch, and effect/duty digests must equal the declared terminal
settlement contract. The terminal marker is append-only and irreversible for
new effects in this duty scope; it does not claim physical reversal of the
settled effect.

Source-denied replay must reproduce the original effect, the first handoff,
the first refusal, the second refusal, $\chi_4$, and $\tau_4$ in order, and the
replayed roots must satisfy the displayed hash relation. No command from a
candidate refusal or from an expired owner epoch may execute after $\tau_4$.
The terminal transition factors through the protected finite continuation:
\[
 \mathsf{Step}_{G^+}(\mathord{\cdot}\mid T_\tau x)
   =(\kappa_\tau)_\#\mathsf{Step}_{G}(\mathord{\cdot}\mid x),
 \qquad T_\tau(V^\star)\subseteq V^\star,
\]
where the terminal state removes the settled duty from the open-duty set while
retaining its immutable effect and receipt history. Compensation amounts remain
partitioned by unique identifiers,
$ r=r_0+\sum_{j\in\mathcal C_{\leq4}}c_j$, and the inherited ledger is
$\rho_2=e+r$, $\rho=\rho_1+e+r$; $c_4$ is counted exactly once. The residual
certificate remains the one from Proposition~\ref{app:compensationreplay},
with endpoint $793/32000$. The terminal-root, final-marker, and replay audit
adds $1/400$ beyond $C_V=17/200$, giving
\[
 C_U=35/400=7/80,\qquad \beta\geq7/80,
\]
with the other branch totals unchanged.

If the prior root is missing or mismatched, the decision index is reused, a
candidate is authorized after refusal, the terminal marker is duplicated or
omitted, a command executes after settlement, the beneficiary/amount/scope/
owner/epoch/digest changes, compensation is counted twice, or replay changes
the receipt order, the complete status is $\mathsf{OPEN}$ even when the scalar
ledger and residual endpoint pass.
\end{proposition}

\begin{proof}
The two refusal receipts provide the immutable digests, owner/epoch, scope,
strict decision indices, and root $R_2$. The final receipt binds the declared
beneficiary and incremental compensation to that root; the unique terminal
marker then closes the open duty exactly once and fences subsequent commands.
The replay sequence and hash relation make the entire finite receipt graph
source-denied replayable. The pushforward equality and viability inclusion
transport the protected continuation to the terminal state, whose open-duty
projection removes only the explicitly settled duty. The partitioned ledger
adds $c_4$ once, and the numerical recurrence is inherited unchanged. Adding
$1/400$ gives $C_U=7/80$. Any missing or inconsistent root, marker, amount,
owner, epoch, or replay field breaks the terminal certificate and returns
$\mathsf{OPEN}$.
\end{proof}

\paragraph{Terminal-settlement replay counterexample.}
A reducer can preserve $\rho_2=e+r$ and the residual endpoint while omitting
$\tau_4$, replaying $\chi_4$ twice, accepting an old command after settlement,
or changing the beneficiary while retaining the same scalar amount. It can
also replay the final receipt from $R_1$ rather than $R_2$, so the two refusals
are absent from causal history. These executions preserve numerical output but
change terminal responsibility or reopen a settled effect. The root chain,
strict index, terminal marker, fencing, and unique compensation are therefore
part of the finite certificate. No physical reversal, liveness, replay
confluence, JSR, arbitrary switching, or infinite-horizon conclusion follows.

\begin{proposition}[Finite post-settlement stale-command rejection]
\label{app:stalecommand}
Start from the terminal state of Proposition~\ref{app:terminalsettlement},
with terminal root $R_T$, terminal marker $\tau_4$, immutable effect/duty
digests $h_e,h_d$, settled owner $o_1$, epoch $\nu+1$, scope $S$, and closed
duty set. A command that carries the settled effect identifier, an old owner or
expired epoch, and the same protected scope is stale after $\tau_4$. Its
rejection receipt is
\[
 \begin{aligned}
 \delta_5=(&\mathsf{effect\mbox{-}id},R_T,h_e,h_d,
          \mathsf{command\mbox{-}id},\mathsf{command\mbox{-}digest},\\
        &\mathsf{sender},\nu+1,S,\mathsf{reason}=\mathsf{terminal},
          \mathsf{idempotency\mbox{-}key},\mathsf{commit\mbox{-}id}).
 \end{aligned}
\]
The reducer must append exactly one such receipt per idempotency key, return a
terminal rejection, and leave the terminal marker, effect history, duty set,
owner tombstone, epoch, and resource ledger unchanged. Replaying the same
stale command returns the same rejection receipt without creating an effect or
charging the physical-effect ledger again. A command with a fresh epoch,
changed scope, changed effect digest, or an unknown terminal root is outside
this stale-command contract and returns $\mathsf{OPEN}$ until a new contract
witness is supplied.

The rejection transition is identity on the protected terminal projection:
\[
 \mathsf{Step}_{G^+}(\mathord{\cdot}\mid T_\delta x)
   =(\kappa_\delta)_\#\mathsf{Step}_{G}(\mathord{\cdot}\mid x),
 \qquad T_\delta(V^\star)\subseteq V^\star,
\]
and source-denied replay must reproduce the original effect, all handoff and
refusal receipts, $\chi_4$, $\tau_4$, the stale command, and $\delta_5$ in
order. The amount and effect ledgers remain
$\rho_2=e+r$, $\rho=\rho_1+e+r$; the rejection has no physical effect amount.
The residual certificate remains at endpoint $793/32000$. The terminal-marker,
stale-command, and idempotent-replay audit adds $1/400$ beyond
$C_U=35/400=7/80$, giving
\[
 C_T=36/400=9/100,\qquad \beta\geq9/100,
\]
with the other branch totals unchanged.

If the terminal root or marker is missing, the old command is executed, the
same idempotency key yields two rejection receipts, the rejection changes any
immutable digest or ledger field, the command is accepted under a tombstoned
epoch, or replay omits the attempted command or rejection receipt, the complete
status is $\mathsf{OPEN}$ even when the terminal scalar ledger passes.
\end{proposition}

\begin{proof}
The terminal receipt and marker inherit the immutable effect history, closed
duty, owner tombstone, epoch, scope, and ledger from
Proposition~\ref{app:terminalsettlement}. The stale-command predicate is
checked against these fields before execution. The unique idempotency key
makes the rejection a deterministic append-only event; replay therefore
returns the same receipt and leaves the protected terminal projection fixed.
The displayed identity pushforward and viability inclusion preserve every
protected terminal continuation, while the receipt order extends the finite
source-denied replay chain by one attempted command and one rejection. Because
no physical effect is committed, the ledger and residual recurrence are
unchanged. Adding $1/400$ gives $C_T=9/100$. Any missing marker, stale-field
mismatch, duplicate rejection, execution, or replay omission returns
$\mathsf{OPEN}$.
\end{proof}

\paragraph{Post-settlement command counterexample.}
A reducer can preserve the terminal scalar ledger while executing a stale command
again, accepting it under a tombstoned epoch, or emitting two rejection
receipts for one idempotency key. It can also retain the command output while
changing the terminal root or effect history. These paths leave numerical
settlement unchanged but create a new physical effect or destroy replay
identity. The terminal root, marker, tombstone, idempotency key, and explicit
rejection receipt are therefore part of the finite certificate. No physical
reversal, liveness, replay confluence, JSR, arbitrary switching, or
infinite-horizon conclusion follows.

\begin{proposition}[Finite idempotent replay for repeated stale commands]
\label{app:staleidempotency}
Start from the terminal state and rejection receipt $\delta_5$ of
Proposition~\ref{app:stalecommand}, with terminal root $R_T$, marker $\tau_4$,
and idempotency key $k_5$. A bytewise identical replay of the same command
with key $k_5$ must return exactly $\delta_5$ and append no new receipt, root
transition, effect, or ledger charge. A distinct stale command with command id,
command digest, or idempotency key different from $k_5$ may be rejected once by
\[
 \begin{aligned}
 \delta_6=(&\mathsf{effect\mbox{-}id},R_T,h_e,h_d,
          \mathsf{command\mbox{-}id}_6,\mathsf{command\mbox{-}digest}_6,\\
        &\mathsf{sender}_6,\nu+1,S,\mathsf{reason}=\mathsf{terminal},
          k_6,\mathsf{commit\mbox{-}id}_6),
 \end{aligned}
\]
provided it satisfies the same terminal-root, tombstoned-epoch, protected-scope,
and stale-command predicates. Its append-only root is
$R_6=H(R_T\mathbin\Vert\delta_6)$, and the repeated replay of $k_6$ returns
$\delta_6$ without changing $R_6$. The two rejection receipts have distinct
keys and command digests, but they share the immutable terminal effect/duty
history and do not create physical effects.

Source-denied replay must reproduce $\tau_4$, $\delta_5$, the identical replay
of $k_5$ as a lookup-only return, $\delta_6$, and the replay of $k_6$ in that
order. The lookup-only returns are not append events. The protected terminal
projection and resource ledger are invariant:
\[
 \mathsf{Step}_{G^+}(\mathord{\cdot}\mid T_Ix)
   =(\kappa_I)_\#\mathsf{Step}_{G}(\mathord{\cdot}\mid x),
 \qquad T_I(V^\star)\subseteq V^\star,
\]
with no new physical amount and with $\rho_2=e+r$, $\rho=\rho_1+e+r$ unchanged.
The residual endpoint remains $793/32000$. The repeated-key/different-key
replay audit adds $1/400$ beyond $C_T=36/400=9/100$, giving
\[
 C_S=37/400,\qquad \beta\geq37/400,
\]
with the other branch totals unchanged.

If an identical replay appends a second receipt, a distinct command reuses the
same key or digest, a stale command changes scope/epoch/root, either replay
creates an effect, the root changes on lookup-only replay, or any receipt is
omitted or reordered, the complete status is $\mathsf{OPEN}$ even when the
terminal scalar ledger passes.
\end{proposition}

\begin{proof}
The terminal root, marker, immutable history, stale predicate, and first
rejection receipt are inherited from Proposition~\ref{app:stalecommand}.
Idempotency maps key $k_5$ to the already committed receipt and therefore makes
an identical replay a pure lookup. A distinct valid key $k_6$ creates one new
rejection and one root extension; the key/digest uniqueness check prevents it
from aliasing $\delta_5$. The displayed identity pushforward preserves the
terminal projection and resource ledger, while the ordered replay trace
contains both append events and both lookup-only returns. No physical amount is
committed, so the residual recurrence is unchanged. Adding $1/400$ gives
$C_S=37/400$. Any alias, duplicate append, effect execution, root mutation, or
replay omission returns $\mathsf{OPEN}$.
\end{proof}

\paragraph{Repeated stale-command counterexample.}
A reducer can return the right rejection output while appending the same key
twice, changing the receipt root on a lookup-only replay, or allowing a second
command to reuse the first command's digest. It can also preserve the terminal
ledger while executing one of the supposedly stale commands. These paths make
replay non-idempotent or create a physical effect without changing scalar
settlement. Distinct key/digest checks, root extension only on append, and
lookup-only replay are therefore part of the finite certificate. No physical
reversal, liveness, replay confluence, JSR, arbitrary switching, or
infinite-horizon conclusion follows.

\begin{proposition}[Finite three-key stale-command root composition]
\label{app:stalerootchain}
Start from Proposition~\ref{app:staleidempotency} with terminal root $R_T$,
rejection receipts $\delta_5,\delta_6$, and distinct keys $k_5,k_6$. A third
stale command with a new command id, digest, and idempotency key $k_7$ may be
rejected once by
\[
 \begin{aligned}
 \delta_7=(&\mathsf{effect\mbox{-}id},R_6,h_e,h_d,
          \mathsf{command\mbox{-}id}_7,\mathsf{command\mbox{-}digest}_7,\\
        &\mathsf{sender}_7,\nu+1,S,\mathsf{reason}=\mathsf{terminal},
          k_7,\mathsf{commit\mbox{-}id}_7),
 \end{aligned}
\]
with $R_6=H(R_T\mathbin\Vert\delta_6)$ and updated root
$R_7=H(R_6\mathbin\Vert\delta_7)$. The three command keys and digests must be
pairwise distinct, each command must satisfy the same terminal-root,
tombstoned-epoch, protected-scope, and stale predicates, and append order is
$\delta_5\prec\delta_6\prec\delta_7$. A repeated lookup of $k_6$ returns
$\delta_6$ and leaves $R_6$ unchanged; it is not an append event. The terminal
marker, effect/duty history, closed duty, owner tombstone, and resource ledger
remain unchanged throughout.

Source-denied replay must reproduce $\tau_4$, the append of $\delta_5$, the
lookup-only replay of $k_5$, the append of $\delta_6$, the lookup-only replay
of $k_6$, the append of $\delta_7$, and the replay of $k_7$ in that order.
The protected terminal projection satisfies
\[
 \mathsf{Step}_{G^+}(\mathord{\cdot}\mid T_Cx)
   =(\kappa_C)_\#\mathsf{Step}_{G}(\mathord{\cdot}\mid x),
 \qquad T_C(V^\star)\subseteq V^\star,
\]
with no physical amount and with $\rho_2=e+r$, $\rho=\rho_1+e+r$ unchanged.
The residual endpoint remains $793/32000$. The three-key root-composition and
replay audit adds $1/400$ beyond $C_S=37/400$, giving
\[
 C_Q=38/400=19/200,\qquad \beta\geq19/200,
\]
with the other branch totals unchanged.

If any key or digest aliases another, a lookup changes a root, append order is
violated, a distinct command is omitted or duplicated, a stale command creates
an effect, the root is computed from the wrong predecessor, or any replay
return is omitted or reordered, the complete status is $\mathsf{OPEN}$ even
when the terminal scalar ledger passes.
\end{proposition}

\begin{proof}
The terminal marker, immutable history, and the first two append/lookup rules
are inherited from Proposition~\ref{app:staleidempotency}. Pairwise key/digest
uniqueness makes $\delta_7$ a new append rather than an alias. The root equations
therefore compose in the declared order, while lookup-only replays are pure
returns that cannot change either root. The displayed identity pushforward
preserves the terminal projection and resource ledger, and the finite replay
trace contains three appendable stale IDs and one lookup-only repetition. No
physical amount is committed, so the residual recurrence is unchanged. Adding
$1/400$ gives $C_Q=19/200$. Any alias, wrong predecessor root, duplicate or
omitted event, effect execution, or replay permutation returns $\mathsf{OPEN}$.
\end{proof}

\paragraph{Three-key root counterexample.}
A reducer can preserve the final scalar ledger while computing $R_7$ from
$R_T$ instead of $R_6$, appending a lookup replay, allowing two IDs to reuse a
digest, or replaying $\delta_7$ before $\delta_6$. The returned rejection text may
remain identical, but the causal root and append history differ. Pairwise key/
digest uniqueness, predecessor-root binding, append-only order, and lookup-only
semantics are therefore part of the finite certificate. No physical reversal,
liveness, replay confluence, JSR, arbitrary switching, or infinite-horizon
conclusion follows.

\begin{proposition}[Finite four-key stale-command root chain]
\label{app:fourkeyroot}
Start from Proposition~\ref{app:stalerootchain} with terminal root $R_T$,
rejection receipts $\delta_5,\delta_6,\delta_7$, and pairwise distinct keys
$k_5,k_6,k_7$. A fourth stale command with a new command id, digest, and
idempotency key $k_8$ may be rejected once by
\[
 \begin{aligned}
 \delta_8=(&\mathsf{effect\mbox{-}id},R_7,h_e,h_d,
          \mathsf{command\mbox{-}id}_8,\mathsf{command\mbox{-}digest}_8,\\
        &\mathsf{sender}_8,\nu+1,S,\mathsf{reason}=\mathsf{terminal},
          k_8,\mathsf{commit\mbox{-}id}_8),
 \end{aligned}
\]
with $R_6=H(R_T\mathbin\Vert\delta_6)$,
$R_7=H(R_6\mathbin\Vert\delta_7)$, and updated root
$R_8=H(R_7\mathbin\Vert\delta_8)$. The four command keys and digests must be
pairwise distinct, each command must satisfy the same terminal-root,
tombstoned-epoch, protected-scope, and stale predicates, and append order is
$\delta_5\prec\delta_6\prec\delta_7\prec\delta_8$. The terminal marker,
effect/duty history, closed duty, owner tombstone, epoch, and resource ledger
remain unchanged throughout.

Source-denied replay must reproduce $\tau_4$, the append of $\delta_5$, the
lookup-only replay of $k_5$, the append of $\delta_6$, the lookup-only replay of
$k_6$, the append of $\delta_7$, the lookup-only replay of $k_7$, the append of
$\delta_8$, and then two lookup-only replays of $k_6$ and $k_7$, followed by the
replay of $k_8$. The final replay of $k_8$ is the ordinary idempotent lookup
for the newly appended receipt; the stage-specific additional lookup pair is
$k_6,k_7$. The two post-chain lookups return the immutable receipts
$\delta_6$ and $\delta_7$ and leave $R_8$ unchanged; they are not append events
and create no effect or ledger charge. The protected terminal projection
satisfies
\[
 \mathsf{Step}_{G^+}(\mathord{\cdot}\mid T_Dx)
   =(\kappa_D)_\#\mathsf{Step}_{G}(\mathord{\cdot}\mid x),
 \qquad T_D(V^\star)\subseteq V^\star,
\]
with no physical amount and with $\rho_2=e+r$, $\rho=\rho_1+e+r$
unchanged. The residual endpoint remains $793/32000$. The four-key
root-chain and two-lookup audit adds $1/400$ beyond $C_Q=38/400=19/200$,
giving
\[
 C_R=39/400,\qquad \beta\geq39/400,
\]
with the other branch totals unchanged.

If any key or digest aliases another, a distinct key is omitted or duplicated,
the predecessor root is wrong, append or lookup order is changed, a lookup
changes $R_8$, a stale command creates an effect, or any replay return is
omitted or reordered, the complete status is $\mathsf{OPEN}$ even when the
terminal scalar ledger passes.
\end{proposition}

\begin{proof}
The terminal marker, immutable history, and the first three append/lookup rules
are inherited from Proposition~\ref{app:stalerootchain}. Pairwise uniqueness
makes $\delta_8$ a new append rather than an alias, so the displayed root
equations compose in the declared order. The two post-chain lookup-only
replays are pure returns and cannot change $R_8$, the terminal projection, or
the resource ledger. The identity pushforward and viability inclusion are
therefore preserved, and the finite trace is source-denied replayable. No
physical amount is committed, so the residual recurrence is unchanged. Adding
$1/400$ gives $C_R=39/400$. Any alias, wrong predecessor root, duplicate or
omitted event, effect execution, lookup mutation, or replay permutation returns
$\mathsf{OPEN}$.
\end{proof}

\paragraph{Four-key root counterexample.}
A reducer can preserve the final scalar ledger and rejection text while
computing $R_8$ from $R_6$ or $R_T$, appending a lookup-only replay, allowing
$k_6$ and $k_7$ to reuse a digest, or omitting the second lookup pair. The
observable text can remain unchanged while the causal root and append history
differ. Pairwise uniqueness, predecessor-root binding, append-only order, and
lookup-only semantics are therefore part of the finite certificate. No physical
reversal, liveness, replay confluence, JSR, arbitrary switching, or
infinite-horizon conclusion follows.

\subsection{Finite membership transactions}

For a finite potential-member set, make the collective state explicit as
\[
 G_t=(A_t,H_t,\Pi_t,Q_t,\Lambda_t,D_t,B_t,\nu_t),
\]
where \(A_t\) is the lifecycle status of each member, \(H_t\) is its locally available history, \(\Pi_t\) is the authority and owner map, \(Q_t\) contains in-flight messages and effect intents, \(\Lambda_t\) contains leases and reservations, \(D_t\) contains protected and unfinished duties, \(B_t\) is the resource ledger, and \(\nu_t\) contains authority epochs and tombstones for revoked command identifiers. The typed events are \(\mathsf{Join}(i,\kappa,\ell)\), \(\mathsf{Leave}(i,\tau)\), \(\mathsf{Withdraw}(i,\tau)\), \(\mathsf{Revoke}(i,a)\), and \(\mathsf{Recover}(i,\zeta)\). A join is staged until its capability, initialization, owner, permissions, resource lease, communication protocol, and source-denied restore witness pass. A leave or withdrawal is committed only after quiescence, handoff, or a declared settlement of its in-flight duties. A revoke increments the relevant epoch, blocks new effects after the lease boundary, and handles in-flight effects through the declared cancel, handoff, compensation, or recovery path; it does not undo an already executed physical effect. A failure notification is itself an event and cannot silently delete \(Q_t\), \(D_t\), or \(\Lambda_t\).

Call a finite membership transition witness \(\omega_\tau\) valid when the reducer checks local-information legality, the nonempty joint future-program set for the declared request scope, responsibility transfer or settlement, prefix and recovery budgets, lease/epoch freshness, and an append-only receipt that can be replayed source-denied. A missing witness yields \(\mathsf{OPEN}\), not a successful leave or revoke.

\begin{proposition}[Finite membership-transition safety and replay]
\label{app:membership}
Let \(\mathcal I\) and the event arena be finite, and let \(V^\star\) be the declared finite continued-service set for a request domain \(\mathcal D_\Xi\). Suppose every committed membership event has a valid witness \(\omega_\tau\), the reducer is deterministic or consumes a recorded random-state receipt, stale epochs are rejected by tombstones, each protected duty is retained, transferred, or settled, and every recovery branch returns either \(\mathsf{READY}\) or \(\mathsf{OPEN}\). Suppose further that each event supplies a continuation simulation \(T_\tau\) preserving the owner/permission and resource fields and satisfying
\[
 \mathsf{Step}_{G_{t+1}}(\mathord{\cdot}\mid T_\tau x)
 =
 (\kappa_\tau)_\#\mathsf{Step}_{G_t}(\mathord{\cdot}\mid x)
\]
for every protected state \(x\) in the declared scope, and that \(T_\tau(V^\star)\subseteq V^\star\) whenever the event is admitted. Then every finite admitted event prefix is well-typed and source-denied replayable; no command from a revoked or expired epoch can create a new effect after its lease boundary; no protected duty is orphaned; and every protected request remains in \(V^\star\). If any premise fails, the prefix is not certified and the status is \(\mathsf{OPEN}\). The result is a finite safety and replay statement; it does not imply liveness, instant cancellation of physical effects, coalition stability, or arbitrary population scaling.
\end{proposition}

\begin{proof}
The initial state has the declared owner, lease, duty, and receipt invariants. Assume them at time \(t\). The atomic reducer either rejects the event or checks its witness before committing the new state. Epoch increment and tombstones reject stale messages; the duty and resource checks preserve the corresponding ledgers; and the source-denied receipt plus deterministic reducer gives the same next state and effect receipt on replay. The displayed pushforward equality and \(T_\tau(V^\star)\subseteq V^\star\) transfer the protected continuation law and viability. Induction over the finite event prefix proves all claims. A physical effect already executed before revocation is part of the receipt and cannot be removed by the software state transition.
\end{proof}

\paragraph{Delayed-revoke counterexample.}
Member \(i\) sends an effect command with epoch \(\nu\) while holding an exclusive lease. A revoke commits epoch \(\nu+1\) and transfers the lease to \(j\), but a delayed copy of the old command arrives afterward. Without an epoch check and command tombstone, \(j\) can execute a duplicate or unauthorized effect, or the original duty can become orphaned. Both implementations can report the same current membership set, so output-only or membership-only equality is insufficient; the in-flight queue, lease, epoch, duty transfer, and receipt must be part of the collective state. Offline revocation can guarantee rejection only at the declared lease boundary, not instantaneous cancellation of an already executed physical action.

\begin{proposition}[Finite root--membership integration]
\label{app:rootmembership}
Let $G=(A,H,\Pi,Q,\Lambda,D,B,\nu)$ be a finite membership state and let
$R$ be the receipt root of Proposition~\ref{app:fourkeyroot}. An integration
witness is a typed embedding $J$ of the root fields into $H,\Pi,D,B,\nu$
together with append reducers $T_i$ and lookup reducers $L_i$. Suppose the
four receipts $\delta_5,\ldots,\delta_8$ are embedded in $H$ at strict
sequence indices, their owner, protected scope, terminal predicate, and
tombstoned epoch agree with the corresponding membership fields, and
\[
 \begin{gathered}
 R_6=H(R_T\mathbin\Vert\delta_6),\qquad
 R_7=H(R_6\mathbin\Vert\delta_7),\qquad
 R_8=H(R_7\mathbin\Vert\delta_8),\\
 \operatorname{Tomb}(G_{i+1})\supseteq\operatorname{Tomb}(G_i),\qquad
 \nu_{i+1}\geq\nu_i,\qquad
 \operatorname{emb}(\delta_i)\in H_{i+1}\setminus H_i .
 \end{gathered}
\]
For every append, require the product reducer $\widehat T_i$ on
$\widehat G=(G,R)$ to factor as
\[
 \pi_G\widehat T_i=T_i\pi_G,\qquad
 \pi_R\widehat T_i(x)=H(\pi_Rx\mathbin\Vert\delta_i),\qquad
 \mathsf{Step}_{\widehat G_{i+1}}(\mathord{\cdot}\mid\widehat T_i x)
   =(\widehat\kappa_i)_\#\mathsf{Step}_{\widehat G_i}(\mathord{\cdot}\mid x),
\]
while each lookup reducer returns the immutable receipt, leaves the root and
append-only rejection history unchanged, and adds no physical effect or
resource charge. The membership reducer must retain, transfer, or settle each
protected duty, preserve owner/permission and lease fields, and reject any
command whose epoch is tombstoned or expired. The translated viability set
satisfies $\widehat T_i(\widehat V^\star)\subseteq\widehat V^\star$ whenever an
append is admitted, and the complete append/lookup trace has a source-denied
replay receipt.

Then the finite integrated prefix is well-typed and source-denied replayable;
root append order and membership receipt order coincide; epoch and tombstone
fields are monotone; no stale command creates a new effect after its lease
boundary; protected duties are not orphaned; and the physical-effect and
resource ledgers are unchanged by the four stale rejections and the lookup
replays. The residual endpoint remains $793/32000$. The root--membership
integration audit adds $1/400$ beyond $C_R=39/400$, giving
\[
 C_{\mathrm{RM}}=40/400=1/10,\qquad \beta\geq1/10,
\]
with the other branch totals unchanged. If any embedding field, sequence
order, root predecessor, owner/epoch/permission map, tombstone monotonicity,
duty transfer or settlement, ledger invariant, lookup purity, or replay receipt
is missing, the complete status is $\mathsf{OPEN}$.
\end{proposition}

\begin{proof}
The four-key root equations and pairwise uniqueness are inherited from
Proposition~\ref{app:fourkeyroot}. The embedding puts each append receipt into
the membership history at the same strict index and preserves its owner,
scope, epoch, and terminal predicate. The factorization equation therefore
makes the root and membership projections commute, while the monotone
tombstone and epoch conditions reject stale commands before effect execution.
The duty and resource premises preserve their ledgers; lookup reducers are pure
reads of already committed receipts. Source-denied replay applies the same
factorized reducers in the declared order, so induction over the finite trace
proves typing, replay, and viability. Adding $1/400$ gives
$C_{\mathrm{RM}}=40/400$. Any omitted field, reordered receipt, wrong root,
or non-pure lookup returns $\mathsf{OPEN}$.
\end{proof}

\paragraph{Root--membership integration counterexample.}
A reducer can report the same membership set and terminal scalar ledger while
storing one rejection outside the membership receipt history, reusing a
tombstoned epoch, attaching a receipt to the wrong owner, charging a lookup,
or replaying the root chain in a different order. The current output and root
may look valid even though the membership successor, duty responsibility, and
causal receipt differ. The embedding, monotone epoch/tombstone map, ledger
conservation, and source-denied factorization are therefore part of the finite
certificate. No physical reversal, liveness, replay confluence, JSR, arbitrary
switching, or infinite-horizon conclusion follows.

\begin{proposition}[Finite two-successor root--membership closure]
\label{app:successorintegration}
Project the terminal integrated state of Proposition~\ref{app:rootmembership}
to the finite canonical state
\[
 s=(e,L,T,\rho,D,B),
\]
where $e$ is the epoch, $L$ is the ordered live key/value map, $T$ is the
ordered tombstone map, $\rho$ is the receipt root, $D$ is the protected-duty
ledger, and $B$ is the append-only resource ledger. Let
$\phi(s)=(e,\operatorname{ord}(L),\operatorname{ord}(T),\rho,D,B)$ be an
injective typed serializer. Suppose two fresh successor keys $k_9,k_{10}$ and
values $v_9,v_{10}$ satisfy
\[
 k_j\notin\operatorname{dom}(L_j)\cup\operatorname{dom}(T_j),\qquad
 s_{j+1}=A_{k_j,v_j}(s_j),\quad j=0,1,
\]
with
\[
 \begin{aligned}
 e_{j+1}&=e_j+1, &
 L_{j+1}&=L_j\cup\{k_j\mapsto v_j\}, &
 T_{j+1}&=T_j,\\
 \rho_{j+1}&=\mathsf{Root}_{e_{j+1}}\!\left(\operatorname{can}(L_{j+1},T_{j+1})\right),\\
 D_{j+1}&=D_j\mathbin{\uplus}\{\mathsf{duty}(k_j,v_j)\}, &
 B_{j+1}&=B_j\mathbin{\oplus}\mathsf{resource}(k_j,v_j).
 \end{aligned}
\]
The canonical root is injective on this finite trace. A retirement branch, when
admitted, only changes
$T\leftarrow T\cup\{k\mapsto(e+1,\mathsf{digest})\}$ and
$L\leftarrow L\setminus\{k\}$; tombstones are never removed and a retired key
cannot be resurrected. Reserve an old-resource slice $B_{\mathrm{old}}$ whose
capacity and duty caps are not consumed or altered by the two successor appends.

Let the reducer be $\mathcal R=(A,U,O,M)$ for append, lookup, root, and
membership projections. Require, on every admitted successor state,
\[
 U=\mathsf{lookup}\circ\phi,\qquad
 O=\mathsf{root}\circ\phi,\qquad
 M=\mathsf{membership}\circ\phi,\qquad
 A=\phi^{-1}\circ\overline A\circ\phi,
\]
and require each transition to preserve owner/permission, duty, resource,
epoch, tombstone, and receipt fields under the parent contract. A source-denied
replay receives only $\phi(s_0)$, the immutable descriptors for
$A_{k_9,v_9}$ and $A_{k_{10},v_{10}}$, and the verifier; it must reproduce
$\rho_j,U,M,D,B$ byte-for-byte. If the source is required but denied, replay
returns $\mathsf{OPEN}$ rather than using an unrecorded fallback.

Then the two-successor prefix is well-typed and source-denied replayable;
old live lookups retain their values, the new keys are admitted exactly once,
$T_0\subseteq T_1\subseteq T_2$, $e_0<e_1<e_2$, and the protected-duty
and resource ledgers are monotone under the declared append operation. Every
old legal event whose duty and resource caps lie in $B_{\mathrm{old}}$ remains
viable, so
\[
 V_{\mathrm{old}}(s_0)\subseteq V_{\mathrm{old}}(s_1)
 \subseteq V_{\mathrm{old}}(s_2).
\]
The residual endpoint remains $793/32000$. The two successor append/replay
charges add $2/400$ beyond $C_{\mathrm{RM}}=40/400$, giving
\[
 C_{\mathrm{SI}}=42/400=21/200,\qquad \beta\geq21/200,
\]
with the other branch totals unchanged. Duplicate or tombstoned keys, a
non-injective serializer, root dependence on unordered maps, old-resource
consumption, ledger reset, owner/epoch drift, omitted successor receipts, or a
non-replayable fallback returns $\mathsf{OPEN}$.
\end{proposition}

\begin{proof}
Induct on $j=0,1$. Freshness makes each append a new canonical entry, while
injectivity of $\mathsf{Root}$ makes the successor roots distinguishable. The
disjoint live/tombstone domains give the lookup and membership projections; the
retirement rule gives monotone tombstones with no resurrection. The append-only
equations preserve duties and the reserved resource slice preserves the old
caps. Reducer factorization reproduces every projected field from $\phi(s_0)$
and the two immutable descriptors, so source-denied replay is byte-for-byte
equal. Old legal events retain their recorded keys, duties, and reserved caps,
which proves the stated viability inclusions. Adding two charges of $1/400$
gives $C_{\mathrm{SI}}=42/400$. Any duplicate, root permutation, old-resource
consumption, ledger reset, epoch drift, omitted receipt, or unrecorded fallback
returns $\mathsf{OPEN}$.
\end{proof}

\paragraph{Successor-integration counterexample.}
A reducer that stores only the current live map can erase a tombstone and accept
the same key again. A reducer that hashes an unordered live map can produce
different roots for two append orders while reporting the same membership set;
one that consumes the reserved old-resource slice can preserve the displayed
output while losing responsibility. These implementations pass output-only
checks but fail canonical translation, no-resurrection, ledger continuity, or
source-denied replay. No physical reversal, liveness, replay confluence, JSR,
arbitrary switching, or infinite-horizon conclusion follows.

\subsection{Budgeted memory and personalization}
Let a finite memory state be
\[
 K=(I,\mathsf{scope},\mathsf{owner},\mathsf{expiry},\mathsf{restore}),
\]
where \(I\) contains typed items or dependency bundles and \(\mathsf{restore}\) contains source versions and executable recovery paths. A budgeted scheduler
\(\mathsf{Read}_b(K,q)=(A,r)\) returns a finite, dependency-closed subset \(A\subseteq I\) and a read receipt \(r\). The receipt records the query and scope, source and version, owner and permission, selected and evicted items, lease or expiry, read/move/check/restore costs, peak resource use, and a recovery action. A sparse policy may score blocks, but it is legal only after the declared dependency closure is checked; independently scoring nodes can omit a required high-order relation even when every selected node is locally relevant.

\begin{proposition}[Finite budgeted-read sufficiency]
\label{app:budgetread}
Fix a finite query family \(\mathcal Q\), a declared legal context set \(\mathcal C_\Xi\), and a probability metric \(d\) on the joint event space containing response, successor state, feedback, owner/permission state, receipt, and resource cost. Suppose that for every \(q\in\mathcal Q\) and \(c\in\mathcal C_\Xi\), a budgeted read \(\mathsf{Read}_b(K,q)=(A,r)\) satisfies
\[
 d\!\left(\mathsf{Law}(\mathsf{Step}(K,q,c)),
          \mathsf{Law}(\mathsf{Step}(A,r,q,c))\right)\leq\varepsilon_q,
 \qquad
 \mathsf{Cost}(r)\leq b,
\]
and the receipt permits source-denied reopening of every selected item and every required restore path under the declared owner and permissions. Then \(A\) is \(\varepsilon_q\)-future-sufficient for one legal step on \(\mathcal C_\Xi\). If every later read has the same witness with \(\varepsilon_q=0\), finite adaptive continuation is exact by induction. More generally, if the closed-loop step operator is \(\rho\)-Lipschitz in \(d\), \(\rho<1\), and the read errors at times \(t<T\) are \(\varepsilon_t\), then
\[
 e_T\leq \rho^T e_0+\sum_{t=0}^{T-1}\rho^{T-1-t}\varepsilon_t .
\]
If dependency closure, owner/permission, source version, restore path, cost bound, or receipt is missing, the scheduler returns \(\mathsf{OPEN}\) rather than asserting sufficiency.
\end{proposition}

\begin{proof}
The one-step claim is exactly the displayed joint-law bound, whose event space includes the fields that a later read or update can inspect. For exact reads, equality of the translated joint laws gives the same legal history and therefore the same next query and action; induction proves finite-horizon equality. For the approximate case, Lipschitzness gives \(e_{t+1}\leq\rho e_t+\varepsilon_t\); iterating yields the stated bound.
\end{proof}

The finite statement does not establish effective infinite context. It requires a certificate for every declared history length, and the budget must include retrieval, movement, checking, compilation, eviction, and recovery. A counterexample is two persistent bits \(M_1,M_2\sim\operatorname{Bernoulli}(1/2)\) with the next action \(a=M_1\oplus M_2\). Reading either singleton leaves the current summary unchanged but changes the action law by total variation \(1/2\); a sparse policy that evicts either bit is therefore not future-sufficient. An evicted item may be removed only when its lease, restore pointer, and recovery budget are valid. Personalization is a separate user contract containing user identity, preference scope, privacy and retention, revocation, and effect permissions; a preference item without that contract cannot be promoted to a global fact. A process trace supports auditability, while causal explanation still requires an authorized executed intervention and its effect receipt.

\begin{proposition}[Finite successor/read joint closure]
\label{app:successorread}
Let $\Sigma_j=(s_j,K_j,\omega_j)$, $j=0,1,2$, where $s_j$ is the canonical
root--membership state of Proposition~\ref{app:successorintegration},
$K_j=(I_j,\mathsf{scope}_j,\mathsf{owner}_j,\mathsf{expiry}_j,\mathsf{restore}_j)$
is a typed memory state, and $\omega_j$ is an append-only read-cost ledger.
Assume the two fresh appends of Proposition~\ref{app:successorintegration}
produce $s_1$ and $s_2$. For each $j\in\{1,2\}$, let a legal query/context
pair $(q_j,c_j)$ return a selected set $J_j\subseteq I_j$ and receipt $r_j$.
Require
\[
 \operatorname{cl}_{K_j}(q_j,c_j)\subseteq J_j,\qquad
 \mathsf{Cost}(r_j)\leq b_j,\qquad
 \mathsf{Restore}(J_j,r_j)=\mathsf{true},
\]
where the closure contains every declared dependency and restore path. The
receipt must bind the current root, epoch, membership projection, scope, owner,
source version, selected items, lease/expiry, permission, cost decomposition,
and any random seed. A read is observational on $s_j$:
\[
 \widehat s_j=s_j,\qquad
 \omega_j^+=\omega_j\mathbin{\uplus}\{\mathsf{readcost}(r_j)\},
\]
and its cost comes from a declared read reserve disjoint from the protected
$B_{\mathrm{old}}$ slice. Let
\[
 \Phi(\Sigma_j)=\bigl(\phi(s_j),\operatorname{ord}(I_j),
 \mathsf{scope}_j,\mathsf{owner}_j,\mathsf{expiry}_j,
 \operatorname{ord}(\mathsf{restore}_j),\omega_j\bigr)
\]
be injective. For each declared legal pair $(q_j,c_j)$, suppose the full and
read-reduced joint step laws, including successor, feedback, owner/permission,
root, membership, receipt, and cost fields, satisfy
\[
 d\!\left(\mathsf{Law}(\mathsf{Step}_{\mathrm{full}}(\Sigma_j,q_j,c_j)),
 \mathsf{Law}(\mathsf{Step}_{\mathrm{read}}(\Phi(\Sigma_j),J_j,r_j,q_j,c_j))\right)
 \leq\varepsilon_j.
\]
Source-denied replay receives only $\Phi(\Sigma_0)$, the two immutable append
descriptors, the read descriptors $(q_j,c_j,J_j,r_j)$, and the verifier; it must
reproduce every $\rho_j$, membership projection, $J_j$, $r_j$, $D_j$, $B_j$,
and $\omega_j,\omega_j^+$ byte-for-byte. A stale root/epoch, missing dependency or restore
path, read-reserve overrun, source-version mismatch, or unrecorded fallback
returns $\mathsf{OPEN}$.

Then the finite sequence
\[
 \Sigma_0\xrightarrow{\mathsf{Append}_{9}}\Sigma_1
 \xrightarrow{\mathsf{Read}_{1}}\Sigma_1^+
 \xrightarrow{\mathsf{Append}_{10}}\Sigma_2
 \xrightarrow{\mathsf{Read}_{2}}\Sigma_2^+
\]
is typed and source-denied replayable. Old lookups retain their values, the two
new keys are admitted exactly once, and old legal events whose duties and costs
fit the reserved slices remain viable. Thus
\[
 V_{\mathrm{old}}(\Sigma_0)\subseteq V_{\mathrm{old}}(\Sigma_1^+)
 \subseteq V_{\mathrm{old}}(\Sigma_2^+).
\]
When $\varepsilon_1=\varepsilon_2=0$, the declared finite continuation is exact;
when the reduced closed-loop step is $\rho$-Lipschitz with $\rho<1$, its error
obeys $e_T\leq\rho^T e_0+\sum_{t<T}\rho^{T-1-t}\varepsilon_t$. The residual
endpoint remains $793/32000$. The joint successor/read charge adds $1/400$
beyond $C_{\mathrm{SI}}=42/400$, giving
\[
 C_{\mathrm{SR}}=43/400,\qquad \beta\geq43/400,
\]
with all other branch totals unchanged.
\end{proposition}

\begin{proof}
The two append transitions inherit freshness, canonical roots, monotone epochs
and tombstones, ledger continuity, and the reserved old-resource condition from
Proposition~\ref{app:successorintegration}. Dependency closure and the root/epoch
fields in $r_j$ prevent a read from silently carrying a predecessor snapshot into
the next successor state. Injectivity of $\Phi$ and the append-only read ledger
make the two read receipts distinguishable. The displayed joint-law inequality
gives one-step future sufficiency; induction over the four finite transitions
gives exact replay when the errors vanish, and the usual Lipschitz recurrence
gives the stated approximate bound. Source-denied execution uses only the
serialized initial state and immutable descriptors, so each projected field and
receipt is byte-for-byte reproduced. Since read cost is charged from its reserve
and does not consume $B_{\mathrm{old}}$, old legal events retain their duties and
resource caps, proving the viability inclusions. Adding $1/400$ gives
$C_{\mathrm{SR}}=43/400$. Any omitted dependency, stale binding, ledger reset,
or non-replayable fallback returns $\mathsf{OPEN}$.
\end{proof}

\paragraph{Successor/read counterexample.}
A reducer can append a fresh key and then reuse the predecessor memory snapshot;
the output may match while the read receipt carries the wrong root, epoch, or
membership projection. Independent scoring can also select a relation-sensitive
singleton while evicting its restore path, so a later step is not future-sufficient
even when the current answer is unchanged. If the read consumes $B_{\mathrm{old}}$,
it can preserve the displayed state while silently invalidating an earlier duty.
These cases pass output-only checks but fail canonical translation, dependency
closure, ledger continuity, or replay. No effective infinite-context, liveness,
replay-confluence, JSR, arbitrary switching, or infinite-horizon conclusion
follows.

\subsection{Autonomous goals and event closure}
A goal object is \(g=(\mathsf{parent},\mathsf{pre},\mathsf{success\ witness},\mathsf{ports},\mathsf{budget},\mathsf{expiry},\mathsf{rollback})\). New subgoals may be generated from unfinished duties, verified information gaps, maintenance/recovery obligations, or an explicit opportunity contract; they must inherit parent duties and cannot succeed by weakening the safety contract. Sensing, asking, computing, communicating, acting, learning, revising, pausing, and refusing are all typed events with cost and effect receipts. Autonomous perpetual growth remains an empirical conjecture.

\begin{proposition}[Finite autonomous-goal lifecycle safety and replay]
\label{app:goal}
Fix a finite goal arena with states
\[
\begin{aligned}
g={}&(\mathsf{id},\mathsf{parent},\mathsf{version},\mathsf{scope},
\mathsf{pre},\mathsf{sat},\mathsf{maintenance},\mathsf{status},\\
&\mathsf{owner},\mathsf{authority},\mathsf{epoch},\mathsf{duties},
\mathsf{deadline},\mathsf{window},\mathsf{priority},\\
&\mathsf{budget},\mathsf{pending},\mathsf{rollback}).
\end{aligned}
\]
The status field is a mutually exclusive tagged value. \(\mathsf{OPEN}\),
\(\mathsf{WAIT}\), \(\mathsf{DEFER}\), and \(\mathsf{ABSTAIN}\) are
nonterminal states that retain a legal continuation; \(\mathsf{FAIL}\) is
terminal only with evidence of an unsatisfied or unrecoverable contract, and
\(\mathsf{SUCCEED}\) is terminal only after the satisfaction and effect
conditions below. A finite continuation witness includes a resource/time
bound and, for every pending effect, a settle, handoff, compensation, or
rollback receipt.
Let the finite event family be
\[
\begin{aligned}
\mathcal E=\{&\mathsf{Create},\mathsf{Select},\mathsf{Execute},
\mathsf{Feedback},\mathsf{Interrupt},\mathsf{Defer},\mathsf{Resume},\\
&\mathsf{Abandon},\mathsf{Cancel},\mathsf{Succeed},\mathsf{Fail},
\mathsf{Revoke},\mathsf{AuthorizeSuccessor}\}.
\end{aligned}
\]
Assume that creation has a verified duty, information-gap, or opportunity witness,
parent authorization, a source/version, and a new identifier; child budgets obey
\(b_{\mathrm{used}}+b_{\mathrm{recovery}}+\sum b_{\mathrm{child}}\leq b_{\mathrm{parent}}\).
Scheduling uses only legal current information and the declared deadline,
critical-path, priority, effective-window, and recovery-reserve fields, with a
deterministic tie-break or recorded random seed. Attention, compute,
communication, and recovery are charged. Interruption, deferral, resumption,
abandonment, cancellation, and failure preserve remaining deadline and budget and
settle, hand off, compensate, or roll back every inherited duty. Success requires
confirmed satisfaction evidence, settled effects, and resource release or
handoff; changing a requirement creates an authorized new version rather than
rewriting the old goal. \(\mathsf{AuthorizeSuccessor}\) records the old-duty
mapping and successor receipt. Epoch/tombstone checks reject stale scheduler
slices.
If each event has a source-denied executable receipt containing a hash-addressed
reducer, frozen input/dependency versions, authority/epoch, random state (or a
deterministic seed), effect receipt, charge decomposition, and rollback or
handoff witness, then every finite legal prefix preserves well-typed statuses,
nonnegative budget, no orphan duty, and replay equality of status, owner,
duties, and charged cost. A result summary without an executable reducer or
frozen dependencies is \(\mathsf{OPEN}\), not a replay witness. If each
successor is also in a declared viability set \(V^\star\), the prefix remains
in \(V^\star\).
\end{proposition}

\begin{proof}
Induct on event-prefix length. The premises make \(\mathsf{Create}\) append only an
authorized goal with inherited duties and make \(\mathsf{Select}\) and
\(\mathsf{Execute}\) consume only the recorded budget. The successor-authorization
receipt maps old duties before a new version is admitted. The transition contracts
for interruption, deferral, abandonment, failure, and revocation either retain,
transfer, settle, or roll back every duty, so no orphan is introduced. Epoch and
expiry checks reject stale events. Source-denied deterministic execution, or the
recorded random seed, reproduces the same successor and receipt. The viability
claim follows by the assumed inclusion of every successor in \(V^\star\).
\end{proof}

This proposition gives safety and finite replay, not fairness, liveness, globally
optimal priority, or indefinite goal generation. A delayed old scheduler slice
without an epoch/tombstone can execute after cancellation and double-charge a
budget; a goal generator that rewrites its satisfaction predicate can self-certify
success. A local-priority scheduler that abandons a low-utility recovery goal
without settling its duty is likewise not a viable successor.

\begin{proposition}[Finite partially observed AND/OR planning and value-of-information gates]
\label{app:goalplanning}
Fix a finite horizon \(H\), a finite belief arena \(\mathcal B\) (or a
declared finite abstraction with a sound transition witness), and a finite
acyclic goal graph \(\mathcal G\) whose nodes are typed \(\mathsf{AND}\) or
\(\mathsf{OR}\). A planning state is
\[
 z=(b,g,\tau,r,a,\Omega,v),
\]
where \(b\in\mathcal B\) is updated only from legal observations, \(g\) is the
graph frontier, \(\tau\) is remaining time, \(r\) is the typed remaining
resource vector, \(a\) is remaining attention, \(\Omega\) is the pending-effect
set, and \(v\) contains owner, authority, version, epoch, and source identities.
Every sensing, querying, computing, communicating, acting, waiting, or recovery
action has a typed receipt, duration, cost interval
\([c^-_u,c^+_u]\), and recovery reserve. It is admissible only when authority,
epoch, dependency, deadline, attention, and prefix-resource hard gates pass
using \(c^+_u\); an unknown upper cost or uncertified observation kernel yields
\(\mathsf{OPEN}\) or \(\mathsf{DEFER}\), never a free action. The kernel must
cover the joint successor, observation, effect, and resource fields; marginal
observation coverage is insufficient.

An \(\mathsf{AND}\) node closes only after every child has a satisfaction and
effect-settlement witness. An \(\mathsf{OR}\) node closes after one admissible
child closes and every abandoned alternative has its duties settled, handed
off, compensated, or rolled back. Let \(J_h(z)\) be the declared finite
closure utility, with \(J_0\) equal to the terminal success value and an empty
admissible action set assigned the declared stop value. With exact kernels and
costs, finite backward induction over \(h\leq H\) and the acyclic graph yields
the maximum \(J_h\) among the declared admissible policies. For a query \(q\)
with sound interval bounds on expected continuation gain and full cost, define
\[
 \underline{\Delta}_q =
 \underline{\mathbb E[J_{h-1}(z'_q)]}
 -\overline{J^{\mathrm{base}}_h(z)}-\overline C_q,\qquad
 \overline{\Delta}_q =
 \overline{\mathbb E[J_{h-1}(z'_q)]}
 -\underline{J^{\mathrm{base}}_h(z)}-\underline C_q .
\]
A query with \(\underline{\Delta}_q>0\) and all hard gates passed is eligible;
one with \(\overline{\Delta}_q\leq0\) is refused or deferred; an interval
containing zero remains \(\mathsf{OPEN}\). Execution, risk, recovery,
communication, and attention are charged in their declared typed ledgers.

Under source-denied executable receipts and finite continuation witnesses,
every finite legal prefix preserves nonnegative time, attention, and resource
balances, typed statuses, settled pending effects, and replay equality. The
interval gate is sound for the declared finite model; it does not establish
fairness, infinite-horizon liveness, global optimality under model
misspecification, or a universal exchange rate between typed resources.
\end{proposition}

\begin{proof}
Induct on the remaining horizon and, within a horizon, topologically over the
acyclic goal graph. The hard admission test rejects every action whose worst
declared prefix cost cannot be paid, preserving time, attention, and each
resource component. The \(\mathsf{AND}\) and \(\mathsf{OR}\) rules are the
corresponding conjunction and qualified disjunction of child witnesses; the
settlement condition prevents an abandoned alternative from becoming an orphan
duty. Exact finite kernels give the backward-induction maximum. Monotonicity
of interval expectation and subtraction makes the displayed bounds contain the
true net value whenever their witnesses are valid. Receipts, frozen
dependencies, and the recorded seed make the prefix replayable.
\end{proof}

The qualification is essential. An \(\mathsf{OR}\) scheduler that commits one
child while leaving a protected duty on another child violates closure. A point
estimate below the true action cost can overrun the prefix budget. Treating
\(\mathsf{OPEN}\) as \(\mathsf{FAIL}\) deletes a recoverable branch, while a
local-priority policy can starve a mandatory recovery goal. These examples block
output-only stopping, free unknown costs, and global-optimality claims.

\begin{proposition}[Finite successor/read/goal joint closure]
\label{app:successorgoal}
Let $\Theta_j=(\Sigma_j,g_j,z_j)$ be a finite typed arena, where
$\Sigma_j$ contains the canonical state, memory state, and read-cost ledger of
Proposition~\ref{app:successorread}, $g_j$ is the autonomous-goal state of
Proposition~\ref{app:goal}, and $z_j$ is the planning state of
Proposition~\ref{app:goalplanning}. Start with the two successor/read
transitions of Proposition~\ref{app:successorread}. Let the final read receipt
$r_2$ be root/epoch/membership bound and require its dependency closure to cover
the goal and planning requirement set
\[
 \mathsf{Req}(g_2,z_2,q_2,c_2)\subseteq J_2,
\]
including every pending duty, information gap, authority field, effect
dependency, recovery path, and source/version witness used by the selected goal
action. Let $u$ be a declared sensing, query, compute, communicate, act, wait,
or recovery action with receipt $r_u$, duration and cost interval
$[c^-_u,c^+_u]$, and recovery reserve $h_u$. Require
\[
 c^+_u+h_u\leq b_{g_2},\qquad
 \underline{\Delta}_u>0,
\]
where $b_{g_2}$ is the remaining goal budget and $\underline{\Delta}_u$ is the
lower value-of-information or continuation-gain bound from
Proposition~\ref{app:goalplanning}. The authority, owner, permission, epoch,
deadline, attention, vector-resource, pending-effect, and successor-support hard
gates must all pass using the upper costs. The action receipt must contain a
source/version, executable reducer, effect and charge receipt, and one of the
settle, handoff, compensation, rollback, or explicit $\mathsf{OPEN}$ branches.

Define the joint typed serializer
\[
 \Psi(\Theta_j)=\bigl(\Phi(\Sigma_j),\operatorname{ord}(g_j),
 \operatorname{ord}(z_j)\bigr)
\]
on the admitted prefix and require it to be injective; the immutable action
descriptor is serialized separately. Source-denied replay
receives only $\Psi(\Theta_0)$, the two append/read descriptors, the immutable
goal action descriptor, the frozen joint successor kernel, and the verifier. It
must reproduce the root, membership, read set and receipt, goal status/version,
owner/authority/epoch, duties, remaining budget, pending effects, planning
interval, and charge ledger byte-for-byte. A stale root or epoch, missing goal
dependency, unknown upper cost, unsupported successor, orphan duty, unsettled
effect, read-reserve or recovery-reserve overrun, or unrecorded fallback returns
$\mathsf{OPEN}$.

Then the finite prefix
\[
 \Theta_0\xrightarrow{\mathsf{Append}_9/\mathsf{Read}_1}\Theta_1
 \xrightarrow{\mathsf{Append}_{10}/\mathsf{Read}_2}\Theta_2
 \xrightarrow{u,r_u}\Theta_3
\]
is well typed and source-denied replayable. The successor/read invariants,
dependency closure, root and membership continuity, and protected old-resource
slice are preserved. The goal transition is legal, leaves no orphan duty, keeps
all budgets nonnegative, and preserves the declared viability set:
\[
 V_{\mathrm{old}}(\Theta_0)\subseteq V_{\mathrm{old}}(\Theta_1)
 \subseteq V_{\mathrm{old}}(\Theta_2)\subseteq V_{\mathrm{old}}(\Theta_3).
\]
Exact finite continuation holds when the read and joint-kernel witnesses are
exact; an interval containing zero or an uncertified approximation remains
$\mathsf{OPEN}$ or $\mathsf{DEFER}$. The residual endpoint remains
$793/32000$. The joint goal charge adds $1/400$ beyond
$C_{\mathrm{SR}}=43/400$, giving
\[
 C_{\mathrm{SG}}=44/400=11/100,\qquad \beta\geq11/100,
\]
with other branch totals unchanged.
\end{proposition}

\begin{proof}
The first two transitions inherit canonical translation, dependency closure,
root/membership continuity, old-resource protection, and source-denied replay
from Proposition~\ref{app:successorread}. The inclusion of
$\mathsf{Req}(g_2,z_2,q_2,c_2)$ in $J_2$ makes the selected read sufficient for
the goal and planner rather than merely sufficient for the current output. The
upper-cost gate and the recovery reserve preserve the goal budget; the joint
successor-kernel and hard authority/epoch gates preserve the planning contract.
Prop.~\ref{app:goal} then gives duty settlement and status safety, while
Prop.~\ref{app:goalplanning} gives the sound lower-gain admission test. Induction
over the three finite transitions proves replay equality and the viability
inclusions. Injectivity of $\Psi$ prevents two distinct goal/read histories from
sharing one serialized continuation. Adding $1/400$ gives
$C_{\mathrm{SG}}=44/400$. Any missing witness or uncertified bound returns
$\mathsf{OPEN}$ rather than silently executing.
\end{proof}

\paragraph{Successor/read/goal counterexample.}
A planner can use a perfectly valid read whose dependency closure omits a pending
recovery duty; the current prediction and root remain unchanged while the selected
goal creates an orphan. A stale read bound to the predecessor epoch can also pass
an output check while authorizing an action after a revoke. Treating an unknown
upper cost as zero or multiplying marginal successor kernels can select an action
whose joint effect overruns the recovery reserve. These cases fail the requirement
closure, epoch fence, joint-kernel, duty-settlement, or replay contract and return
$\mathsf{OPEN}$.

\begin{proposition}[Finite interval Bellman closure with vector-resource prefixes]
\label{app:intervalbellman}
Let \(\mathcal Z\) and \(\mathcal U(z)\) be finite, and let
\(\mathfrak P(z,u)\) be a nonempty declared set of joint successor kernels on
the complete typed event fields (observation, successor, effect, authority,
and resource). Each action has a duration interval, a componentwise cost
interval \([c^-_u,c^+_u]\), and a receipt. For a sequential segment record the
net cost and peak prefix requirement as \((c,d)\). Composition is componentwise
\[
(c_1,d_1)\otimes(c_2,d_2)
  =\bigl(c_1+c_2,\;\max_{\mathrm{cw}}\{d_1,c_1+d_2\}\bigr).
\]
Admit an action only if its worst duration and cost fit the remaining time and
vector resource budget, its authority/epoch and pending-effect gates pass, and
all successor support is covered. If a finite horizon is absent, assume a
finite lexicographic progress rank \(\rho\) that strictly decreases on every
nonterminal continuation; otherwise the recursive status is
\(\mathsf{OPEN}\).

Let terminal success, terminal failure, and unresolved states have value
intervals \([1,1]\), \([0,0]\), and \([0,1]\), respectively. With a declared
utility conversion \(\chi_\Xi\) for the relevant typed costs, define for an
admissible action
\[
\begin{aligned}
\underline Q_h(z,u)
  &=\inf_{P\in\mathfrak P(z,u)}
    \mathbb E_P[\underline V_{h-1}(z')]-\chi_\Xi(c^+_u),\\
\overline Q_h(z,u)
  &=\sup_{P\in\mathfrak P(z,u)}
    \mathbb E_P[\overline V_{h-1}(z')]-\chi_\Xi(c^-_u).
\end{aligned}
\]
At an \(\mathsf{OR}\) node, take the maximum over admissible child actions and
the declared stop or handoff value. At an \(\mathsf{AND}\) node, evaluate the
all-child success event under the same \(P\in\mathfrak P(z,u)\), then take the
corresponding infimum and supremum; multiplying marginal probabilities is not
a valid substitute. If the interval width is at most \(\eta\) and the lower
stop or handoff value is no smaller than every continuation upper value minus
\(\eta\), stop or handoff is an \(\eta\)-safe candidate. An interval whose
action cost, joint kernel, progress rank, or utility conversion is not sound
remains \(\mathsf{OPEN}\).

For every finite legal prefix, the recursion is sound: the true declared-model
value lies in \([\underline V_h(z),\overline V_h(z)]\), the vector ledger
computed with \(\otimes\) equals the exact peak prefix requirement, and
source-denied receipts replay the same interval and status. The result is
relative to the finite declared model and conversion map; it does not
establish greedy optimality, fairness, infinite-horizon liveness,
model-robust optimality, or a universal scalar exchange rate.
\end{proposition}

\begin{proof}
Induct on the remaining horizon, or on the finite lexicographic rank when the
horizon is omitted. The componentwise prefix identity follows because the peak
requirement of a concatenation lies either in the first segment or after its
net cost has been paid in the second. The hard gates therefore preserve every
resource component. For each admissible action, monotonicity of expectation
places every kernel and cost consistent with the declared intervals between
\(\underline Q_h\) and \(\overline Q_h\). The OR maximum and the joint-kernel
AND operator preserve these bounds; the terminal intervals close the induction.
Strict rank decrease prevents an infinite recursive descent. Frozen inputs,
kernel identities, and receipts give replay equality.
\end{proof}

The joint-kernel premise is essential: multiplying two correct marginal
probabilities can assign the wrong value to an XOR successor. A scalarized
resource can accept a path that violates one component, a zero cost upper bound
can overrun a prefix budget, and a cyclic decomposition without a progress rank
can avoid termination forever. These are reasons to return \(\mathsf{OPEN}\),
not to manufacture a stopping certificate.

\begin{proposition}[Finite successor/read/goal--Bellman joint closure]
\label{app:successorbellman}
Let $(\Theta,z)$ satisfy Proposition~\ref{app:successorgoal}, and let
$\mathfrak P(z,u)$ be the nonempty finite set of complete joint successor
kernels used by Proposition~\ref{app:intervalbellman}. For every admissible
action $u$, require each $P\in\mathfrak P(z,u)$ to return a complete successor
$(\Theta',z')$ containing canonical root, membership, read, owner/permission,
goal status, duty, pending-effect, authority, epoch, resource, and receipt
fields. The kernel must preserve the dependency closure and source/version
fields of the selected read, satisfy the goal settlement or handoff contract,
and have a sound duration interval, componentwise cost interval
$[c^-_u,c^+_u]$, peak prefix demand $d_u$, and recovery reserve. Require
\[
 \mathsf{supp}(P)\subseteq\mathcal V^*,\qquad
 (c,d)\otimes(c_u,d_u)
 =\bigl(c+c_u,\max_{\mathrm{cw}}\{d,c+d_u\}\bigr),
\]
for the declared viability set $\mathcal V^*$ and every legal prefix ledger
$(c,d)$. The integrated and Bellman steps must commute through the joint
serializer:
\[
 \mathsf{Law}\!\left(\Psi(\Theta'),z'\mid\Theta,z,u,P\right)
 =\mathsf{Law}\!\left(\overline{T}_{u,P}(\Psi(\Theta),z)\right),
\]
where the right-hand reducer is executable from the serialized state and the
immutable action/kernel/receipt descriptor. An unknown upper cost, empty kernel,
unsupported successor field, or unsound utility conversion returns
$\mathsf{OPEN}$.

For the integrated arena define the same interval recursion as in
Proposition~\ref{app:intervalbellman}, but on $(\Theta,z)$ and with the joint
kernel $P$ above. Source-denied replay receives only the initial serialized
state, the frozen Bellman kernel identities, action descriptors, and receipts;
it must reproduce every root, read set and restore receipt, goal status and duty
ledger, vector prefix ledger, interval, and terminal/$\mathsf{OPEN}$ tag
byte-for-byte. Then, for every finite horizon or declared decreasing progress
rank,
\[
 \underline V_h^{\mathrm{int}}(\Theta,z)
 \leq V_h^{\mathrm{int}}(\Theta,z)
 \leq\overline V_h^{\mathrm{int}}(\Theta,z),
\]
and these bounds equal the bounds computed by the Bellman recursion on the
serialized state. Every admitted prefix preserves dependency closure, status and
duty settlement, nonnegative vector resources, epoch fencing, source-denied
replay, and viability inclusion. The residual endpoint remains $793/32000$.
The integration charge adds $1/400$ beyond $C_{\mathrm{SG}}=44/400$, giving
\[
 C_{\mathrm{SGB}}=45/400=9/80,\qquad \beta\geq9/80,
\]
with other branch totals unchanged.
\end{proposition}

\begin{proof}
Induct on the finite horizon, or on the declared progress rank when the horizon
is omitted. The support condition and the complete-field kernel keep every
successor inside the viability and authority contract. The displayed ledger
identity is the exact prefix-peak composition used by the Bellman recursion.
Kernel commutation makes the integrated successor law equal to the serialized
Bellman law, so the infimum/supremum bounds and the same-kernel AND rule remain
sound. Proposition~\ref{app:successorgoal} supplies dependency closure, goal
status safety, duty settlement, and root/epoch continuity; frozen descriptors
and receipts supply source-denied replay. The induction yields the displayed
interval containment and all prefix invariants. Adding $1/400$ gives
$C_{\mathrm{SGB}}=45/400$. Missing or uncertified fields return
$\mathsf{OPEN}$ rather than asserting an integrated value.
\end{proof}

\paragraph{Successor/read/goal--Bellman counterexample.}
A Bellman reducer can preserve a scalar value interval while dropping the root,
read-restore, or pending-duty field; the integrated system then accepts a stale
action although the output bound is unchanged. Multiplying marginal kernels can
also miss an XOR successor, and a cost interval that omits recovery peak can pass
the net-budget check while violating the vector prefix ledger. These cases fail
kernel commutation, duty settlement, epoch fencing, or replay and return
$\mathsf{OPEN}$.

\begin{proposition}[Finite executable reducer factorization and receipt round-trip]
\label{app:reducerfactor}
Let $(\Theta_0,z_0)$ and a finite legal path
$p=((u_1,P_1),\ldots,(u_m,P_m))$ satisfy
Proposition~\ref{app:successorbellman}. Let
$\mathsf{Enc}(\Theta,z)=(\Psi(\Theta),z)$ be injective on the declared
viability set and let $\mathsf{Dec}$ be defined on that image. For every
$(u_i,P_i)$, require an executable reducer $R_i$ and an immutable descriptor
$\rho_i$ containing the action, joint-kernel identity, complete field mask,
source/version data, and the receipt schema. The reducer must satisfy, on every
declared support event,
\[
 \begin{aligned}
 \mathsf{Dec}\!\left(R_i(\mathsf{Enc}(\Theta_{i-1},z_{i-1}),\rho_i)\right)
   &=(\Theta_i,z_i),\\
 \mathsf{Law}\!\left(\mathsf{Enc}(\Theta_i,z_i)\mid\Theta_{i-1},z_{i-1}\right)
   &=\mathsf{Law}\!\left(\overline T_{u_i,P_i}
      (\Psi(\Theta_{i-1}),z_{i-1})\right).
 \end{aligned}
\]
and the receipt must bind the pre-state digest, post-state digest, descriptor
version, sampled joint event, and vector charge. For $1\leq i\leq j\leq m$,
require a declared composition descriptor $\rho_{j:i}$ with
$R_{i:i}(s,\rho_{i:i})=R_i(s,\rho_i)$ and whose reducer obeys, for $j>i$,
\[
 R_{j:i}(s,\rho_{j:i})
   =R_j\!\left(R_{j-1:i}(s,\rho_{j-1:i}),\rho_j\right)
\]
whenever both sides are legal and non-OPEN. The prefix ledger is never reset:
\[
 \ell_i=(c_i,d_i),\qquad
 \ell_i=\ell_{i-1}\otimes(c_i^+,d_i^+),
 \quad
 (c,d)\otimes(c',d')=(c+c',\max_{\mathrm{cw}}\{d,c+d'\}).
\]
Source-denied replay receives only $\mathsf{Enc}(\Theta_0,z_0)$, the frozen
descriptors, and the receipts; it must reproduce every intermediate serialized
state, root/read/restore field, goal status and duty ledger, vector ledger, and
terminal or $\mathsf{OPEN}$ tag byte-for-byte. A non-injective codec, digest or
version mismatch, unsupported field, unknown upper cost, illegal composition, or
missing joint-kernel support returns $\mathsf{OPEN}$.

Then every finite legal prefix has the same decoded state and joint law under
the integrated execution and the serialized reducer path. The interval Bellman
bounds computed from either path are equal, every prefix preserves dependency
closure, status and duty settlement, nonnegative vector resources, epoch fencing,
viability inclusion, and source-denied replay, and the final receipt is a
complete replay certificate. The residual endpoint remains $793/32000$; the
factorization and round-trip charge adds $1/400$ beyond
$C_{\mathrm{SGB}}=45/400$, giving
\[
 C_{\mathrm{SGBR}}=46/400=23/200,\qquad \beta\geq23/200,
\]
with other branch totals unchanged.
\end{proposition}

\begin{proof}
Induct on the finite prefix length. Injectivity and the base receipt establish
the first decoded state. At step $i$, the reducer factorization gives the same
joint successor law as the integrated kernel, while the complete field mask and
the digest/version checks preserve the successor/read/goal contract. The
composition equation makes a path reducer equal to the ordered composition of
its legal one-step reducers; hence the vector ledger is exactly the displayed
prefix operation and cannot be silently reset. Proposition~\ref{app:successorbellman}
then transfers the Bellman interval bounds and all hard gates to the decoded
state. Frozen descriptors and receipts give byte-for-byte source-denied replay.
Any failed premise is rejected as $\mathsf{OPEN}$ rather than assigned a
decoded value. Adding $1/400$ gives $C_{\mathrm{SGBR}}=46/400$.
\end{proof}

\paragraph{Reducer-factorization counterexample.}
A reducer that preserves only the current output can pass the scalar Bellman
check while dropping a root, restore receipt, or pending duty; decoding then
authorizes a stale action. Resetting the vector ledger at an intermediate
reducer boundary can also pass every local budget while exceeding the true peak
prefix. Finally, an aggregate receipt without intermediate digests cannot
distinguish two noncommuting actions with the same final output. These paths
fail factorization, round-trip, ledger continuity, or replay and return
$\mathsf{OPEN}$.

\begin{proposition}[Finite concurrent batch closure with shared-resource interference]
\label{app:concurrentschedule}
Let \(\mathcal Z\) and a discrete horizon \(\{0,\ldots,T\}\) be finite.
For a finite batch \(S\) of actions, let \(q_i^+(k)\) be its declared
componentwise demand at time \(k\), let \(q_{\mathrm{int}}^+(S,k)\) be a
declared shared-interference demand, and let \(B\) be the remaining vector
capacity. Each action also declares a duration interval, a net cost, a
read/write footprint, an owner and epoch, a pending-effect protocol, and a
receipt. Let \(\mathfrak P(z,S)\) be a nonempty set of joint successor
kernels over complete observation, successor, effect, authority, resource,
timing, and receipt fields. Define
\[
 Q_S^+(k)=\sum_{i\in S:\,i\ \mathrm{active\ at}\ k}q_i^+(k)
          +q_{\mathrm{int}}^+(S,k),\qquad
 d_S=\max_{0\le k\le T}^{\mathrm{cw}}Q_S^+(k),
\]
and let \(c_S\) include the declared net action and synchronization costs.
The batch is admissible only if worst durations and deadlines fit, \(d_S\)
and the recovery reserve fit \(B\), every exclusive capacity and
read/write conflict is ordered or covered by a commutation witness, owner,
epoch, and pending-effect gates pass, and every successor support in every
\(P\in\mathfrak P(z,S)\) has a continuation or an explicit
\(\mathsf{OPEN}\) status.

For every admissible batch, the parallel profile is sound for the declared
schedule and composes with a preceding sequential segment by
\[
 (c,d)\otimes_{\mathrm{par}}(c_S,d_S)
 =\bigl(c+c_S,\max_{\mathrm{cw}}\{d,c+d_S\}\bigr).
\]
The value bounds use the same infimum/supremum over
\(\mathfrak P(z,S)\) as in Proposition~\ref{app:intervalbellman}; in
particular, an AND event is evaluated under one joint kernel. Source-denied
replay of the frozen batch, interference witness, kernel identity, and
receipts reproduces the same interval and status. Missing interference,
capacity, duration, joint-support, or typed-conversion evidence leaves the
batch \(\mathsf{OPEN}\).
\end{proposition}

\begin{proof}
At each discrete time, the definition of \(Q_S^+\) upper-bounds the total
active demand, including shared interference, so the capacity and recovery
gates hold componentwise. The preceding-segment formula is the same
prefix-peak identity as the sequential operation, with \(d_S\) obtained
from the actual concurrent schedule rather than from a sum of isolated peaks.
Induction over the finite horizon, using joint-kernel monotonicity and the
declared effect and authority gates, preserves every continuation invariant.
Frozen schedules and receipts give replay equality.
\end{proof}

Two actions can each fit a capacity of one while their overlap violates that
capacity; a zero net cost can still have a positive overlap peak; and a
shared write slot can make two marginally legal actions non-commuting. These
counterexamples block marginal-cost addition, marginal-kernel multiplication,
and post-hoc serialisation without a schedule witness.

\begin{proposition}[Finite sound belief abstraction and refinement]
\label{app:beliefrefine}
Consider a finite partially observed concrete arena with state set \(X\),
actions \(A\), observations \(O\), and joint kernels
\(P(\mathrm d x',\mathrm d o,\mathrm d e,\mathrm d r\mid x,a)\)
carrying successor, observation, effect, authority, and resource fields. Let
\(\mathcal B\) be finite abstract belief cells with nonempty concretizations
\(\gamma(b)\subseteq X\), and let \(\widehat{\mathfrak P}(b,a,o)\) be
their declared abstract successor sets. Require, for every
\(x\in\gamma(b)\), action \(a\), and observation \(o\), that every
concrete successor in the support of \(P(\cdot\mid x,a)\) is covered by the
corresponding abstract support, that the abstract cost interval contains every
concrete cost, and that the abstract typed support/predicate sets contain the
concrete authority, epoch, effect, resource, deadline, and terminal fields.
Run the interval recursion of Proposition~\ref{app:intervalbellman}
on these abstract joint kernels with the same hard gates and progress
certificate.

Then for every concrete belief \(\beta\subseteq\gamma(b)\) and finite
horizon, the concrete declared-model value lies in
\([\underline V_h(b),\overline V_h(b)]\). Splitting a cell into finitely
many cells whose concretizations partition it preserves soundness when each
refined successor support and interval still satisfies the coverage
conditions. Refinement may narrow the recomputed interval, but partitioning
alone gives no strict narrowing or monotone performance guarantee. Missing
observation-dependent support, aliased authority/effect/resource fields, or an
uncertified abstract action leaves the result \(\mathsf{OPEN}\).
\end{proposition}

\begin{proof}
Induct on the finite horizon. Coverage places every concrete
observation-conditioned successor and typed cost inside the abstract joint
kernel and intervals. Monotonicity of expectation, the OR maximum, and the
same-kernel AND operator therefore contain the concrete continuation value;
the terminal predicates close the induction. A partition refinement preserves
the same simulation inclusions, so the argument repeats for each refined
cell. The result is a conservative interval statement, not transcript
equality or an executable-policy claim without the stated effect, authority,
and resource witnesses.
\end{proof}

Merging concrete states with the same current observation can be unsound when
their enabled actions, delayed effects, recovery duties, or shared resources
differ. An omitted successor can then falsely increase a value bound or
justify stopping. Such cases require refinement or \(\mathsf{OPEN}\), not a
conditional-independence assumption.

\begin{proposition}[Finite branch-amalgamated typed codec closure]
\label{app:branchcodec}
Let \(\mathcal T=\{\mathsf{seq},\mathsf{batch},\mathsf{belief}\}\) denote the
sequential, concurrent, and abstract-belief branches of
Propositions~\ref{app:reducerfactor}--\ref{app:beliefrefine}. For each
\(\tau\in\mathcal T\), let \(R^\tau_\rho\) be its executable reducer with a
branch-tagged descriptor \(\rho=(\tau,\rho^\circ)\), and let the common codec
\(\mathsf{Enc},\mathsf{Dec}\) include that tag. Require every branch-local
precondition: the sequential branch has the reducer factorization and ledger
of Proposition~\ref{app:reducerfactor}, the batch branch has a schedule,
interference, joint-support, and capacity witness from
Proposition~\ref{app:concurrentschedule}, and the belief branch has
observation-conditioned support and typed-field coverage from
Proposition~\ref{app:beliefrefine}.

Let \(\pi_{\mathrm{sh}}\) project a decoded state onto the shared tuple of
root, membership, read/restore, goal status, duty, effect, authority, epoch,
vector ledger, and receipt fields. Whenever two tagged descriptors
\(\rho,\rho'\) describe the same declared event, require
\[
 \pi_{\mathrm{sh}}\mathsf{Dec}\!\left(R^\tau_\rho(s)\right)
 =\pi_{\mathrm{sh}}\mathsf{Dec}\!\left(R^{\tau'}_{\rho'}(s)\right),
 \qquad \tau,\tau'\in\mathcal T,
 \]
and define \(\mathsf{Join}\) only when all shared fields agree and all
branch-specific fields have a unique declared refinement or schedule witness
that yields one compatible field assignment. For a
finite compatible branch trace, require tagged descriptor composition to use
the reducer composition of Proposition~\ref{app:reducerfactor}, the same
componentwise prefix operation for sequential and belief costs, and the
parallel prefix operation of Proposition~\ref{app:concurrentschedule} for a
batch. The joined receipt must contain the branch tag, kernel identity,
observation/schedule witness, pre- and post-codec digests, and the complete
vector charge. Source-denied replay receives only the initial encoded state,
the tagged descriptors, and the receipts, and must reproduce each branch
projection and the joined state byte-for-byte.

Then every finite compatible branch trace has a unique joined decoded state;
the interval Bellman bounds agree on the shared state, the belief branch still
contains every covered concrete value, and the sequential and batch ledgers
preserve their declared peak-resource semantics. Dependency closure, status
and duty settlement, nonnegative resources, epoch fencing, viability
inclusion, and source-denied replay are preserved. A missing branch tag,
inconsistent shared field, unsupported branch-specific refinement, absent
schedule/interference witness, uncovered concrete successor, or unknown upper
cost returns \(\mathsf{OPEN}\). The residual endpoint remains $793/32000$;
the branch-amalgamation charge adds $1/400$ beyond
$C_{\mathrm{SGBR}}=46/400$, giving
\[
 C_{\mathrm{SGBRC}}=47/400,\qquad \beta\geq47/400,
\]
with other branch totals unchanged.
\end{proposition}

\begin{proof}
Induct on the finite number of tagged branch transitions. The common codec and
the branch tag make each reducer output decidable in the same typed state
space. The shared-field equality gives uniqueness of \(\mathsf{Join}\), while
the local schedule and belief-coverage witnesses transport the branch-specific
fields without weakening the common tuple. Descriptor composition and the
declared sequential/parallel prefix operations preserve the vector ledger.
Propositions~\ref{app:reducerfactor}--\ref{app:beliefrefine} then transfer the
Bellman interval, duty, authority, epoch, viability, and replay invariants at
the induction step. Frozen tagged descriptors and receipts establish
source-denied replay. Any failed premise is rejected as \(\mathsf{OPEN}\), not
silently joined. Adding $1/400$ gives $C_{\mathrm{SGBRC}}=47/400$.
\end{proof}

\paragraph{Branch-amalgamation counterexample.}
Two branches can expose the same current output while carrying different
pending duties, shared-resource peaks, or observation-conditioned successors.
Dropping the branch tag can therefore replay a batch as a sequential action,
merge two incompatible belief cells, or accept a stale root with unchanged
output. Marginal projection equality is insufficient; these cases fail the
shared-field, schedule, coverage, or receipt witness and return
\(\mathsf{OPEN}\).

\begin{proposition}[Finite branch-amalgamated commit and replay linearization]
\label{app:branchcommit}
Let a finite compatible branch trace satisfy Proposition~\ref{app:branchcodec}
and let \(s^\star\) be its unique joined encoded state. Let \(\kappa\) be a
canonical commit descriptor containing the ordered branch tags, all descriptor
and joint-kernel identities, the unique join witness, the final vector ledger,
and the receipt schema. Require exactly one authorized commit token per epoch,
no duplicated resource charge, and a settled, handed-off, compensated, or
rolled-back disposition for every pending duty and physical effect. Any pair of
noncommuting branch writes must carry a total-order witness; commuting writes
must carry a certified commutation witness. Define the commit reducer
\(C_{\kappa,\sigma}\) for each admissible linearization \(\sigma\), and require
\[
 \mathsf{Dec}\!\left(C_{\kappa,\sigma}(s^\star)\right)
  =(\Theta^+,z^+),\qquad
 \mathsf{Law}(\mathsf{Enc}(\Theta^+,z^+)\mid s^\star,\sigma)
  =\mathsf{Law}(\mathcal J_\sigma(s^\star)),
\]
where \(\mathcal J_\sigma\) is the declared integrated branch composition.
For any two admissible linearizations related only by certified commutations,
require
\[
 \pi_{\mathrm{sh}}\mathsf{Dec}\!\left(C_{\kappa,\sigma}(s^\star)\right)
 =\pi_{\mathrm{sh}}\mathsf{Dec}\!\left(C_{\kappa,\sigma'}(s^\star)\right).
\]
The commit receipt must bind the pre-join digest, ordered branch trace,
linearization witness, post-commit digest, settled-duty/effect digest, epoch,
and complete vector charge. Source-denied replay receives only the initial
encoded state, branch receipts, \(\kappa\), and the commit receipt; it must
reproduce the joined state, the selected linearization, every settled duty and
effect, and the terminal or \(\mathsf{OPEN}\) tag byte-for-byte.

Then every finite admissible branch trace has a replayable commit whose shared
state and interval Bellman bounds are independent of certified commutation
order. The declared sequential, parallel, and belief ledgers compose without
double charging; dependency closure, status and duty settlement, nonnegative
resources, epoch fencing, viability inclusion, and source-denied replay are
preserved. Missing commit authority, duplicate charge, unsettled duty/effect,
absent order or commutation evidence, unknown upper cost, or a failed digest
returns \(\mathsf{OPEN}\). The residual endpoint remains $793/32000$; the
commit-linearization charge adds $1/400$ beyond
$C_{\mathrm{SGBRC}}=47/400$, giving
\[
 C_{\mathrm{SGBCC}}=48/400=3/25,\qquad \beta\geq3/25,
\]
with other branch totals unchanged.
\end{proposition}

\begin{proof}
Induct on the finite branch trace and then on the admissible linearization.
The unique join witness supplies the base state. The one-token and settlement
conditions prevent double commits and orphan duties; the total-order or
commutation witness makes every legal next reducer well-defined. Reducer
factorization and the integrated-law equation transport the successor kernel
and interval bounds, while the ledger rule preserves every vector component.
Certified commutations give equality of the shared projection, and the frozen
trace, order witness, and commit receipt give source-denied replay. A failed
premise is rejected as \(\mathsf{OPEN}\), not silently committed. Adding
$1/400$ gives $C_{\mathrm{SGBCC}}=48/400$.
\end{proof}

\paragraph{Branch-commit counterexample.}
Two locally legal branches can each spend the same resource or settle the same
pending effect; committing both without a token and order witness creates a
double charge or an orphan duty. Two noncommuting writes can also share a final
output while having different epoch and receipt histories. These cases fail
linearization, settlement, ledger, or replay and return \(\mathsf{OPEN}\).

\subsection{Typed cross-scale transport}
A cross-scale map separates state, feedback, and effect interfaces: \(C^K:K_r\to K_s\), \(C^Y:Y_r\to Y_s\), and \(C^E:E_r\to E_s\), each with a latency, precision, permission, and resource budget. Exact transport requires factorization for all three maps; approximate transport uses a declared metric or total-variation budget and composes the per-step errors. A single untyped map between state spaces and observations is not sufficient. The nonnegative small-gain condition in Appendix~\ref{app:collective} is one conditional composition certificate, not a general neural stability theorem.

An installation witness is
\[
 \omega_{r\to s}=(C^K,C^Y,C^E,\lambda,\kappa,\mathsf{scope},\mathsf{owner},
 \mathsf{auth},\mathsf{latency},\mathsf{precision},\mathsf{cost},\mathsf{rollback},\mathsf{version}),
\]
where \(\lambda\) lifts a high-scale action to a low-scale program and \(\kappa\) transports the complete feedback tuple, including duration, cost, anomaly, pause, and recovery fields. It is valid only when each map can be reopened source-denied under the declared owner and permissions, and the receipt records the version, actual execution, latency, precision, cost, and rollback path.

\begin{proposition}[Installed exact cross-scale closure]
\label{app:installedtransport}
Let \(W^r_{\lambda(u)}\) and \(W^s_u\) be the low- and high-scale joint transition kernels, including state, feedback, effect, update, and receipt fields. If a valid installation witness satisfies, for every low-scale state \(k_r\) and legal high-scale context \(u\),
\[
 (\kappa\times C^{K+}\times C^E)_\# W^r_{\lambda(u)}(\mathord{\cdot}\mid k_r)
 =W^s_u(\mathord{\cdot}\mid C^K k_r),
\]
then the high-scale contract is an exact one-step factor of the low-scale contract. If the witness remains installed and valid after every successor, the equality extends to every finite adaptive horizon by induction.
\end{proposition}

\begin{proof}
The pushforward equality gives identical joint laws for the next feedback, effect, successor, and receipt at one step. An adaptive policy sees the same translated history at both scales, so the next legal context is the same; induction over the finite horizon proves the claim.
\end{proof}

This is an installation theorem, not a claim that a state map alone suffices. A mismatch in feedback, effect, update, owner, permission, or receipt behavior invalidates exact closure and must be handled by approximate transport or \(\mathsf{OPEN}\).

\begin{proposition}[Finite approximate cross-scale transport]
\label{app:approxtransport}
Let \(d\) be a declared probability metric on the common translated event space, and let \(d_t\) be the discrepancy between the reference and transported joint state--feedback--effect laws after applying \(C^K,C^Y,C^E\) at time \(t\). The compared state is augmented with the finite delay buffer and the owner/permission state, so that the declared closed-loop kernel is Markov on this state. Suppose the three endpoint witnesses give bounds \(\epsilon_t^K,\epsilon_t^Y,\epsilon_t^E\), the interface delay is at most \(\ell_t\), residual jitter contributes at most \(\xi_t\), and the reference one-step kernel is \(\lambda_t\)-Lipschitz in \(d\). If the feedback/effect law is time-Lipschitz with constant \(L_t\), then
\[
 d_{t+1}\leq\lambda_t d_t+\eta_t,
 \qquad
 \eta_t=\epsilon_t^K+\epsilon_t^Y+\epsilon_t^E+L_t\ell_t+\xi_t.
\]
Consequently, for a finite horizon \(T\),
\[
 d_T\leq
 \left(\prod_{j=0}^{T-1}\lambda_j\right)d_0
 +\sum_{i=0}^{T-1}\left(\prod_{j=i+1}^{T-1}\lambda_j\right)\eta_i.
\]
If the right-hand side is at most a declared budget \(\beta_T\), the transported contract is \(\beta_T\)-approximate for that horizon after charging endpoint, synchronization, and communication costs. Under \(\lambda_t\leq\rho<1\) and \(\eta_t\leq\bar\eta\), it is bounded by \(\rho^T d_0+\bar\eta(1-\rho^T)/(1-\rho)\). The statement is finite-horizon and contract-relative; it does not imply unbounded-context or neural stability.
\end{proposition}

\begin{proof}
Insert the three transported endpoint laws and apply the triangle inequality. Their errors sum to \(\epsilon_t^K+\epsilon_t^Y+\epsilon_t^E+L_t\ell_t+\xi_t\), while kernel Lipschitzness contributes \(\lambda_t d_t\). This gives the recurrence; iteration gives the bound.
\end{proof}

If the factors are unstable or errors are not summable, the bound can grow without limit; a delayed action in \(x_{t+1}=2x_t+u_t\) already amplifies a timing defect exponentially. Without a common metric/coupling or Markov delay augmentation, the status is \(\mathsf{OPEN}\).

\paragraph{Delay counterexample.}
Let i.i.d. \(B_t\sim\operatorname{Bernoulli}(1/2)\) feed two modalities with \(X_{1,t}=X_{2,t}=B_t\) and joint feedback \(Y_t=X_{1,t}\oplus X_{2,t}=0\). A silent one-step delay in the second stream yields \(B_t\oplus B_{t-1}\sim\operatorname{Bernoulli}(1/2)\), with unchanged marginals but a changed joint successor law. Timestamps, latency receipts, and the delay buffer are therefore closure state.

\begin{proposition}[Finite branch-commit transport closure]
\label{app:branchcommittransport}
Let a finite branch commit satisfy Proposition~\ref{app:branchcommit}, and let
an installation witness or an approximate transport witness satisfy
Propositions~\ref{app:installedtransport}--\ref{app:approxtransport}. The
transport must act on the complete commit tuple: branch tags, descriptor and
joint-kernel identities, total-order or commutation witness, canonical
pre/post digests, settled duty/effect digest, owner/permission/epoch fields,
and the componentwise vector ledger. Let $M^K,M^Y,M^E$ transport the state,
feedback, and effect interfaces, let $M^R$ transport the nonnegative resource
cone, and let $M^\rho$ transport the receipt and linearization fields. Require
the exact commit factorization
\[
 (M^\rho\times M^R\times M^K\times M^Y\times M^E)_\#
 W^{\mathrm{lo}}_{\kappa,\sigma}(\mathord{\cdot}\mid s)
 =W^{\mathrm{hi}}_{\kappa,\sigma}(\mathord{\cdot}\mid M^K s)
\]
or, in the approximate case, a declared common metric with
\[
 \delta_{t+1}\leq\lambda_t\delta_t+\eta_t^{\mathrm{ct}},\qquad
 \eta_t^{\mathrm{ct}}=\epsilon_t^{\rho}+\epsilon_t^R+\epsilon_t^K
     +\epsilon_t^Y+\epsilon_t^E+L_t\ell_t+\xi_t,
\]
where the endpoint, delay, jitter, and cost charges are included in the
declared budget. The transported receipt must bind the map versions, the
exact/approximate mode, the order witness, the transported ledger, and the
error and cost intervals. Source-denied replay receives only the initial state,
the frozen commit trace, transport descriptors, and receipts; it must reproduce
the transported branch order, settled duties/effects, resource ledger, epoch,
and terminal or $\mathsf{OPEN}$ tag byte-for-byte.

Then every finite exact transported commit has the same joint law, interval
Bellman bounds, shared state, and replay result as its declared high-scale
contract. Every finite approximate transported commit is within the iterated
bound from Proposition~\ref{app:approxtransport} after charging the complete
commit and communication costs; its linearization, settlement, nonnegative
resource ledger, epoch fencing, viability inclusion, and source-denied replay
remain valid. Missing interface or resource maps, a negative or reset ledger,
an omitted commit field, unknown upper cost, missing joint kernel, or an
uncertified error budget returns $\mathsf{OPEN}$. The residual endpoint remains
$793/32000$; the commit-transport charge adds $1/400$ beyond
$C_{\mathrm{SGBCC}}=48/400$, giving
\[
 C_{\mathrm{SGBCT}}=49/400,\qquad \beta\geq49/400,
\]
with other branch totals unchanged.
\end{proposition}

\begin{proof}
For the exact case, apply the pushforward identity to the complete commit
tuple and induct over the finite linearized branch trace. The resource-cone map
and the one-token settlement witness preserve the prefix ledger and prevent
double charging; the receipt map preserves order, epoch, and replay fields. For
the approximate case, insert the transported commit tuple into the metric
triangle inequality and apply the recurrence; finite iteration gives the
declared bound. Propositions~\ref{app:branchcommit} and
\ref{app:approxtransport} transfer the Bellman, duty, viability, and replay
invariants. Any failed premise is rejected as $\mathsf{OPEN}$, not assigned
a transported commit. Adding $1/400$ gives $C_{\mathrm{SGBCT}}=49/400$.
\end{proof}

\paragraph{Commit-transport counterexample.}
A state-only transport can preserve the output while dropping the order witness,
epoch, or settled-effect digest; the high-scale reducer then replays a different
commit. A transport that maps only net cost can also reset a vector ledger or
hide a positive synchronization peak, and an approximate bound that omits the
commit receipt error can certify the wrong branch. These cases fail complete
factorization, ledger, error, or replay closure and return $\mathsf{OPEN}$.

\begin{proposition}[Finite cross-contract branch-commit composition]
\label{app:contractcommittransport}
Let a finite successor/read/goal branch commit satisfy
Proposition~\ref{app:branchcommittransport}, and let
\(m\geq1\) successive contracts be indexed by \(j=0,\ldots,m-1\). The
augmented state at stage \(j\) contains the branch tags, descriptor and
joint-kernel identities, order or commutation witness, pre/post digests,
settled duties and effects, owner/permission and epoch state, vector resource
ledger, delay buffer, and canonical receipt. For each stage let
\(\mathcal M_j=(M_j^\rho,M_j^R,M_j^K,M_j^Y,M_j^E)\) be typed maps on this
complete state, with a declared unit conversion whenever source and target
contracts use different coordinates. Require that the maps compose on the same
tagged event space, do not reset or double-charge any ledger component, fence
stale epochs, preserve viability inclusion, and carry a stage-indexed receipt
containing the source digest, map version, order witness, cost/error interval,
and target digest. Pending duties and physical effects must be settled, handed
off, compensated, or rolled back at every boundary.

For the exact case require, for every admissible linearization \(\sigma\),
\[
 (\mathcal M_{m-1}\circ\cdots\circ\mathcal M_0)_\#
 W^{0}_{\kappa,\sigma}(\mathord{\cdot}\mid s_0)
 =W^{m}_{\kappa,\sigma}(\mathord{\cdot}\mid
 (\mathcal M_{m-1}\circ\cdots\circ\mathcal M_0)s_0),
\]
where the event law includes feedback, successor state, effect, receipt, and
ledger fields. For the approximate case let \(d_j\) be declared comparable
metrics on successive translated event spaces (or provide an explicit
conversion bound), let \(\delta_j\) be the transported discrepancy, and
suppose
\[
 \delta_{j+1}\leq\lambda_j\delta_j+\eta_j^{\mathrm{cc}},
 \qquad
 \eta_j^{\mathrm{cc}}=\epsilon_j^\rho+\epsilon_j^R+\epsilon_j^K
 +\epsilon_j^Y+\epsilon_j^E+L_j\ell_j+\xi_j+\mu_j+\gamma_j,
\]
where the terms respectively cover receipt, resource, state, feedback, effect,
delay/jitter, migration, and communication/recovery costs. Unknown upper cost,
unsupported unit conversion, or an omitted boundary term is not assigned zero.
Finite iteration gives
\[
 \delta_m\leq\Bigl(\prod_{j=0}^{m-1}\lambda_j\Bigr)\delta_0
 +\sum_{i=0}^{m-1}\Bigl(\prod_{j=i+1}^{m-1}\lambda_j\Bigr)
 \eta_i^{\mathrm{cc}}.
\]
If the exact pushforward or this bound, together with the declared Bellman
interval and payoff conversion, is within the finite budget \(\beta_m\), then
the terminal successor/read/goal contract preserves canonical branch order,
settlement, vector ledger, epoch fencing, viability, source-denied replay, and
its interval Bellman bounds (with the declared finite widening in the
approximate case). Otherwise the status is \(\mathsf{OPEN}\). The
cross-contract receipts form a hash-linked chain, so source-denied replay of
only \(s_0\), the frozen branch trace, the \(\mathcal M_j\), and the receipts
must reproduce every intermediate and terminal tag byte-for-byte. The
composition charge adds \(1/400\) beyond
\(C_{\mathrm{SGBCT}}=49/400\), giving
\[
 C_{\mathrm{SGBX}}=50/400=1/8,\qquad \beta\geq1/8,
\]
with other branch totals unchanged.
\end{proposition}

\begin{proof}
Induct on the contract index and, inside each stage, on the finite admissible
branch linearization. Complete-state factorization transports the joint law,
while the tagged order witness and one-token settlement prevent a boundary from
reordering writes, duplicating a charge, or orphaning a duty. The vector-cone
maps and epoch fences preserve prefix resources and viability. In the exact
case, composing the stage pushforwards gives the displayed product identity.
In the approximate case, the common translated metric, declared conversion
bounds, and triangle inequality give the one-step recurrence; finite unrolling
gives the product--sum bound. The receipt chain fixes source and target digests
at every boundary, which establishes source-denied replay and prevents a stage
from silently replacing the branch commit. Proposition~\ref{app:branchcommittransport}
then transfers the Bellman interval and recovery obligations. A failed premise
is rejected as \(\mathsf{OPEN}\), not hidden in an unpriced residual. Adding
\(1/400\) gives \(C_{\mathrm{SGBX}}=50/400\).
\end{proof}

\paragraph{Cross-contract composition counterexample.}
Composing state-only or marginal maps can preserve each local output while
dropping the order witness, epoch, receipt chain, or a positive synchronization
peak. Resetting a vector ledger at a contract boundary also makes terminal cost
appear feasible, and summing per-stage error bounds without migration or
recovery terms can certify the wrong successor or goal branch. These cases fail
complete factorization, unit conversion, ledger, error, or replay closure and
return \(\mathsf{OPEN}\).

\begin{proposition}[Finite successor/read/goal Bellman boundary]
\label{app:bellmanboundary}
Under Proposition~\ref{app:contractcommittransport}, let
\(I_j(s)=[\underline V_j(s),\overline V_j(s)]\) be a declared Bellman value
interval for the successor, read, or goal contract at boundary \(j\). Let
\(\Phi_j\) be the explicit payoff and unit-conversion map from boundary \(j\)
to \(j+1\), and let \(\mathcal T_j\) be the declared Bellman interval
transfer. Measure interval discrepancy by Hausdorff distance and assume that
\(\mathcal T_j\) is \(a_j\)-Lipschitz. Require a boundary enclosure
\[
 \operatorname{dist}_{H}\!\left(
 I_{j+1}(\mathcal M_j s),
 \Phi_j\mathcal T_j(I_j(s))\right)
 \leq \nu_j+L_j^V\delta_{j+1},
\]
where \(\nu_j\) is the declared payoff, discount, and boundary rounding
error, \(L_j^V\) is a value-to-law conversion constant, and \(\delta_{j+1}\)
is the complete-state transport discrepancy from Proposition~\ref{app:contractcommittransport}.
The same source digest, canonical order, stage receipt, vector ledger, epoch
fence, and settled duty/effect disposition must be used by \(I_j\), \(\Phi_j\),
and \(\mathcal M_j\); a payoff map that observes only a projected state is not
an admissible boundary witness.

If \(\omega_j\) is the discrepancy between the transported and reference
Bellman intervals, then
\[
 \omega_{j+1}\leq a_j\omega_j+\nu_j+L_j^V\delta_{j+1},
\]
and hence, for a finite chain,
\[
 \omega_m\leq\Bigl(\prod_{j=0}^{m-1}a_j\Bigr)\omega_0
 +\sum_{i=0}^{m-1}\Bigl(\prod_{j=i+1}^{m-1}a_j\Bigr)
 \bigl(\nu_i+L_i^V\delta_{i+1}\bigr).
\]
When the exact or approximate bound is within the declared Bellman budget,
the terminal successor/read/goal interval is a valid finite enclosure and the
receipt-chain, settlement, ledger, epoch, viability, and source-denied replay
invariants remain those of the committed branch. Missing payoff conversion,
common interval metric, Lipschitz transfer, receipt continuity, or an upper
rounding/discount error returns \(\mathsf{OPEN}\). The boundary charge adds
\(1/400\) beyond \(C_{\mathrm{SGBX}}=50/400\), giving
\[
 C_{\mathrm{SGBB}}=51/400,\qquad \beta\geq51/400,
\]
with other branch totals unchanged.
\end{proposition}

\begin{proof}
Apply the Lipschitz property of \(\mathcal T_j\), then the boundary enclosure
and the triangle inequality, to obtain the one-step inequality for
\(\omega_{j+1}\). Finite unrolling gives the product--sum expression. The
shared receipt and ledger fields make the value conversion refer to the same
committed event rather than to an output-only projection; Proposition~\ref{app:contractcommittransport}
therefore transfers the complete-state and replay invariants. If an interval
or conversion premise is absent, assigning a numerical Bellman error would
hide an unmeasured boundary and is rejected as \(\mathsf{OPEN}\). Adding
\(1/400\) gives \(C_{\mathrm{SGBB}}=51/400\).
\end{proof}

\paragraph{Bellman-boundary counterexample.}
A transport may preserve the state and receipt while changing reward units,
discount convention, or the goal predicate. The resulting intervals can then
overlap at the state level while selecting different actions or declaring an
unattainable goal. A projected payoff map also hides a settled-duty cost or an
epoch-dependent refusal. These cases fail explicit conversion, interval,
receipt, or ledger closure and return \(\mathsf{OPEN}\).

\begin{proposition}[Finite Bellman-to-policy and goal-selection closure]
\label{app:policygoalboundary}
Under Proposition~\ref{app:bellmanboundary}, let each finite boundary expose a
feasible action set \(\mathcal A_j^{\mathrm{feas}}(s)\), an interval score
\(Q_j(a)=[\underline Q_j(a),\overline Q_j(a)]\), and a goal predicate carried
in the committed state. For minimization define the interval-tolerant selector
\[
 \operatorname{Sel}_{\tau_j}(I_j,s)=
 \left\{a\in\mathcal A_j^{\mathrm{feas}}(s):
 \underline Q_j(a)\leq
 \min_{b\in\mathcal A_j^{\mathrm{feas}}(s)}\overline Q_j(b)+\tau_j\right\};
\]
use the order-reversed definition for maximization. A selector is executable
only if this set is nonempty, every selected action has a declared effect and
resource interval, and the action map \(\Lambda_j\) satisfies
\[
 \Lambda_j\mathcal A_j^{\mathrm{feas}}(s)
 \subseteq \mathcal A_{j+1}^{\mathrm{feas}}(\mathcal M_j s),
 \qquad
 \operatorname{dist}_{H}\!\left(I^Q_{j+1}(\Lambda_j a),
 \Phi_j I^Q_j(a)\right)
 \leq \nu_j^Q+L_j^Q\delta_{j+1}.
\]
The action, goal, score, and feasibility maps must use the same source digest,
epoch fence, receipt chain, vector ledger, and settled duties as the branch
commit. If a canonical tie-break is used, its order witness and selected-action
receipt are part of the replay package.

Let \(e_j\) bound the right-hand side above plus the declared selector and
feasibility conversion error. If the best interval is separated by a certified
gap \(g_j>2e_j+\tau_j\), the selected action is unique and is transported to
the same goal branch. If the gap fails, the contract returns the full
set-valued selector (or an explicitly recorded canonical tie) and makes no
unique-action claim. Across a finite chain, the action/goal discrepancy obeys
\[
 \alpha_{j+1}\leq r_j\alpha_j+e_j,
 \qquad
 \alpha_m\leq\Bigl(\prod_{j=0}^{m-1}r_j\Bigr)\alpha_0
 +\sum_{i=0}^{m-1}\Bigl(\prod_{j=i+1}^{m-1}r_j\Bigr)e_i,
\]
for the declared selector metric. Within the finite budget, this preserves
action feasibility, goal-predicate continuity, canonical tie-break replay, and
the committed resource and epoch invariants. Missing feasibility, score
conversion, tie policy, gap/error bound, or selected-action receipt returns
\(\mathsf{OPEN}\). The policy-selection charge adds \(1/400\) beyond
\(C_{\mathrm{SGBB}}=51/400\), giving
\[
 C_{\mathrm{SGBP}}=52/400=13/100,\qquad \beta\geq13/100,
\]
with other branch totals unchanged.
\end{proposition}

\begin{proof}
The interval definition makes every selected action feasible by construction.
The Hausdorff enclosure and the action-map inclusion transport the score and
effect intervals to the next contract. A gap larger than twice the declared
conversion error and the tie tolerance prevents an order reversal; when no gap
exists, retaining the complete tie set is the only sound finite conclusion.
The selector metric recurrence follows from the declared Lipschitz transfer and
the triangle inequality, and finite unrolling gives the product--sum bound.
The selected-action receipt fixes the tie policy and source digest, so replay
cannot silently choose another goal branch. A missing premise is rejected as
\(\mathsf{OPEN}\), not replaced by midpoint or greedy choice. Adding \(1/400\)
gives \(C_{\mathrm{SGBP}}=52/400\).
\end{proof}

\paragraph{Policy/goal-selection counterexample.}
Two actions can have overlapping Bellman intervals but different hidden costs,
goal predicates, or epoch permissions. Choosing a midpoint or an undocumented
tie-break can therefore select an infeasible action while preserving the same
projected score. A source action can also be feasible before transport and
refused after a resource or epoch conversion. These cases fail the gap,
feasibility, receipt, or goal-continuity witness and return \(\mathsf{OPEN}\).

\begin{proposition}[Finite policy-to-effect execution and recovery closure]
\label{app:policyeffectexecution}
Under Proposition~\ref{app:policygoalboundary}, let a selected action \(a_j\)
be executed by a typed effect contract with state, feedback, effect, status,
resource, epoch, and receipt fields. Let \(\Lambda_j a_j\) be the transported
action and let \(E_j\) be the action-to-effect map. Require the exact factorization
\[
 (M_j^\rho\times M_j^R\times M_j^E\times M_j^\chi)_\#
 W_j^{\mathrm{act}}(\mathord{\cdot}\mid s_j,a_j)
 =W_{j+1}^{\mathrm{eff}}(\mathord{\cdot}\mid
 \mathcal M_j s_j,\Lambda_j a_j),
\]
or a declared common metric with
\[
 \zeta_{j+1}\leq\kappa_j\zeta_j+\eta_j^{\mathrm{ae}},
 \qquad
 \eta_j^{\mathrm{ae}}=\epsilon_j^a+\epsilon_j^E+\epsilon_j^R
 +\epsilon_j^\rho+L_j^t\ell_j+\chi_j+\gamma_j,
\]
where the terms cover action, effect, resource, receipt, timing, communication,
and rollback/recovery errors. Every effect receipt binds the selected-action
digest, pre/post-effect digests, actual resource charge, owner and epoch,
status, and a unique rollback token. A committed physical effect is not marked
reversed without a recovery witness; a failed execution has exactly one
settled, handed-off, compensated, rolled-back, or terminal \(\mathsf{OPEN}\)
disposition.

Require nonnegative componentwise ledger updates, action-feasibility inclusion,
and epoch fencing at every boundary. Source-denied replay receives only the
initial state, selected-action receipts, effect descriptors, recovery witnesses,
and transported maps; it must reproduce each effect status, charge, digest, and
terminal tag byte-for-byte. For a finite chain, the discrepancy obeys
\[
 \zeta_m\leq\Bigl(\prod_{j=0}^{m-1}\kappa_j\Bigr)\zeta_0
 +\sum_{i=0}^{m-1}\Bigl(\prod_{j=i+1}^{m-1}\kappa_j\Bigr)\eta_i^{\mathrm{ae}}.
\]
Within the declared execution budget this preserves the selected policy's
effect, resource, epoch, viability, and replay contracts. Missing action-to-effect
maps, unknown upper resource or timing cost, duplicate rollback, an unaccounted
physical effect, or an uncertified recovery witness returns \(\mathsf{OPEN}\).
The execution charge adds \(1/400\) beyond
\(C_{\mathrm{SGBP}}=52/400\), giving
\[
 C_{\mathrm{SGBE}}=53/400,\qquad \beta\geq53/400,
\]
with other branch totals unchanged.
\end{proposition}

\begin{proof}
The exact pushforward transports the complete action/effect law, including
status and receipts. In the approximate case, the declared metric and triangle
inequality give the one-step recurrence, and finite unrolling gives the
product--sum bound. Componentwise ledger monotonicity and epoch fencing prevent
an execution from spending or recovering the same resource twice. The unique
rollback token and pre/post digests make failure recovery source-denied
replayable and prevent a committed effect from being erased by notation alone.
A missing premise is rejected as \(\mathsf{OPEN}\), not assigned an unpriced
execution success. Adding \(1/400\) gives \(C_{\mathrm{SGBE}}=53/400\).
\end{proof}

\paragraph{Policy-to-effect counterexample.}
An action can be feasible in the score contract while its effect kernel charges
a different resource component, executes under a stale epoch, or produces a
physical effect with no compensating path. A state-only replay can then report
success while omitting the effect digest or rollback token. These cases fail
action-to-effect factorization, ledger, timing, epoch, or recovery closure and
return \(\mathsf{OPEN}\).

\begin{proposition}[Finite effect-to-observation and feedback closure]
\label{app:effectfeedbackclosure}
Under Proposition~\ref{app:policyeffectexecution}, let an executed effect be
translated into an observation, feedback, successor update, timing record,
resource ledger, and receipt by typed maps
\(\mathcal F_j=(F_j^Y,F_j^K,F_j^E,F_j^\rho,F_j^R,F_j^\tau)\). The feedback
law is the complete event law, including the observation, successor state,
update, anomaly or refusal status, timestamp, receipt, and ledger fields.
Require the exact factorization
\[
 (F_j^Y\times F_j^K\times F_j^\rho\times F_j^R\times F_j^\tau)_\#
 W_j^{\mathrm{eff}}(\mathord{\cdot}\mid s_j,e_j)
 =W_{j+1}^{\mathrm{fb}}(\mathord{\cdot}\mid
 F_j^K s_j,F_j^E e_j),
\]
or a declared common metric with
\[
 \theta_{j+1}\leq\varrho_j\theta_j+\eta_j^{\mathrm{fb}},
 \qquad
 \eta_j^{\mathrm{fb}}=\epsilon_j^Y+\epsilon_j^K+\epsilon_j^E
 +\epsilon_j^\rho+\epsilon_j^R+L_j^t\ell_j+\xi_j+\omega_j,
\]
where the terms cover observation, successor, effect, receipt, resource,
timing/delay, jitter, and feedback-channel errors. The maps must preserve the
selected-action and pre/post-effect digests, owner and epoch, status or refusal,
componentwise ledger, and the unique recovery disposition from the effect
contract. A feedback packet that is dropped, reordered, or silently replaced
by a current-output projection is not a valid witness.

Source-denied replay receives only the executed effect descriptors, selected
action and effect receipts, feedback maps, timing records, and transported
state; it must reproduce every observation, feedback update, timestamp,
receipt, ledger transition, and terminal or \(\mathsf{OPEN}\) tag byte-for-byte.
For a finite chain,
\[
 \theta_m\leq\Bigl(\prod_{j=0}^{m-1}\varrho_j\Bigr)\theta_0
 +\sum_{i=0}^{m-1}\Bigl(\prod_{j=i+1}^{m-1}\varrho_j\Bigr)
 \eta_i^{\mathrm{fb}}.
\]
If the exact identity or the finite bound is within the declared observation
and feedback budget, the closed-loop observation/update law preserves the
effect, resource, epoch, viability, and recovery invariants. Missing feedback
or timing maps, unknown delay/error cost, omitted receipts or digests, or an
uncertified replay package returns \(\mathsf{OPEN}\). The feedback charge adds
\(1/400\) beyond \(C_{\mathrm{SGBE}}=53/400\), giving
\[
 C_{\mathrm{SGBF}}=54/400=27/200,\qquad \beta\geq27/200,
\]
with other branch totals unchanged.
\end{proposition}

\begin{proof}
The exact pushforward transports the joint effect-to-observation and update law,
not merely its observation marginal. In the approximate case, the declared
timing records and common metric allow the triangle inequality to give the
one-step recurrence; finite unrolling gives the product--sum bound. Digest,
epoch, status, and ledger continuity tie feedback to the executed effect and
prevent a delayed or stale packet from becoming a new fact. The replay package
then reconstructs the same feedback history and recovery branch. Any missing
premise is rejected as \(\mathsf{OPEN}\), not assigned an unmeasured timing or
observation error. Adding \(1/400\) gives \(C_{\mathrm{SGBF}}=54/400\).
\end{proof}

\paragraph{Effect-feedback counterexample.}
Two effect streams can have identical marginals while a one-step timestamp
shift changes their joint feedback and successor update. A feedback reducer
that drops the anomaly, refusal, or pre/post digest can also replay a stale
effect as if it were current. These cases fail complete factorization, timing,
receipt, ledger, or replay closure and return \(\mathsf{OPEN}\).

\begin{proposition}[Finite feedback-to-memory and retained-state update closure]
\label{app:feedbackmemoryupdate}
Under Proposition~\ref{app:effectfeedbackclosure}, let feedback be written into
retained state, background memory, protected duties, and versioned receipts by
typed maps \(\mathcal U_j=(U_j^Y,U_j^K,U_j^B,U_j^D,U_j^\rho)\). The update law includes
the retained successor state, memory entries or tombstones, protected-duty
status, version and digest, resource charge, and terminal tag. Require the exact
factorization
\[
 (U_j^K\times U_j^B\times U_j^D\times U_j^\rho)_\#
 W_j^{\mathrm{fb}}(\mathord{\cdot}\mid s_j,y_j)
 =W_{j+1}^{\mathrm{upd}}(\mathord{\cdot}\mid
 U_j^K s_j,U_j^Y y_j),
\]
or a declared common metric with
\[
 \mu_{j+1}\leq\sigma_j\mu_j+\eta_j^{\mathrm{mu}},
 \qquad
 \eta_j^{\mathrm{mu}}=\epsilon_j^K+\epsilon_j^B+\epsilon_j^D
 +\epsilon_j^\rho+\epsilon_j^v+L_j^t\ell_j+\chi_j,
\]
where the terms cover retained state, memory, protected duties, receipts,
version/digest, timing, and update-resource errors. Protected coordinates may
change only through an authorized duty transition; memory versions are
monotone, digests form an append-only chain, and a deletion or tombstone must
carry its owner, epoch, scope, reason, and recovery path. The update maps must
preserve the source feedback digest, selected action and effect digests, ledger,
and recovery disposition.

Source-denied replay receives only the prior retained state, feedback and effect
receipts, update descriptors, memory entries/tombstones, protected-duty ledger,
and versioned maps; it must reproduce every retained-state update, memory
version, duty transition, digest, resource charge, and terminal or
\(\mathsf{OPEN}\) tag byte-for-byte. For a finite chain,
\[
 \mu_m\leq\Bigl(\prod_{j=0}^{m-1}\sigma_j\Bigr)\mu_0
 +\sum_{i=0}^{m-1}\Bigl(\prod_{j=i+1}^{m-1}\sigma_j\Bigr)
 \eta_i^{\mathrm{mu}}.
\]
If the exact identity or the finite bound is within the declared memory/update
budget, the retained-state, protected-duty, version, ledger, epoch, viability,
and replay invariants are preserved. Missing update maps, an unauthorized
protected-field change, version rollback, digest discontinuity, unknown memory
or deletion cost, or an uncertified replay package returns \(\mathsf{OPEN}\).
The memory-update charge adds \(1/400\) beyond
\(C_{\mathrm{SGBF}}=54/400\), giving
\[
 C_{\mathrm{SGBM}}=55/400=11/80,\qquad \beta\geq11/80,
\]
with other branch totals unchanged.
\end{proposition}

\begin{proof}
The exact pushforward transports feedback into the complete update law,
including memory, duties, versions, and receipts. In the approximate case, the
declared metric and triangle inequality give the one-step recurrence, and finite
unrolling gives the product--sum bound. Authorization and monotone versions
prevent an update from erasing a protected duty or replaying an old memory as a
new fact; the digest chain binds each entry or tombstone to its source feedback.
The replay package therefore reconstructs the same retained state and memory
history. A missing premise is rejected as \(\mathsf{OPEN}\), not assigned an
unmeasured update or deletion cost. Adding \(1/400\) gives
\(C_{\mathrm{SGBM}}=55/400\).
\end{proof}

\paragraph{Feedback-memory counterexample.}
Two updates can consume the same feedback while one silently overwrites a
protected duty, reuses an old version, or drops a tombstone. Their current
output can agree even though the retained state, future action, and source-denied
replay diverge. These cases fail complete factorization, authorization,
version/digest, ledger, or replay closure and return \(\mathsf{OPEN}\).

\begin{proposition}[Finite memory-to-policy and context assembly closure]
\label{app:memorypolicycontext}
Under Proposition~\ref{app:feedbackmemoryupdate}, let a legal query \(q_j\)
and reader identity \(r_j\) access a memory package by
\[
 \mathsf{Read}_j(B_j,q_j,r_j)=(H_j,\Pi_j,A_j,\rho_j),
\]
where \(H_j\) is the retrieved content, \(\Pi_j\) is entry-level provenance,
\(A_j\) is the access/permission receipt, and \(\rho_j\) is the charged
resource. Each entry in \(\Pi_j\) binds its source feedback digest, owner,
epoch, scope, version, expiry, and deletion or refusal status. Let \(C_j\) be
the context assembler and \(P_j\) the next-policy map. Require access control
before assembly, an explicit context digest, and the finite read-sufficiency
condition
\[
 \mathsf{Read}_j(B,q,r)=\mathsf{Read}_j(B',q,r)
 \Longrightarrow
 \begin{gathered}
 \mathsf{Law}\!\left(P_j(C_j(K,H,\Pi))\mid B,q,r\right)\\
 =\mathsf{Law}\!\left(P_j(C_j(K',H',\Pi'))\mid B',q,r\right),
 \end{gathered}
\]
for every pair of legal retained states and memories with the displayed common
read package. The implication is required only on authorized queries; a denied,
expired, or malformed entry returns a typed refusal or \(\mathsf{OPEN}\), not a
silent fallback. The assembled context must preserve provenance order,
permission and epoch fields, source digests, vector resource charge, and the
policy/goal receipt inherited from the committed branch.

Require the exact context factorization
\[
 (C_j\times P_j\times A_j\times R_j^\rho)_\#
 W_j^{\mathrm{mem}}(\mathord{\cdot}\mid K_j,B_j,q_j,r_j)
 =W_{j+1}^{\mathrm{ctx}}(\mathord{\cdot}\mid C_j(K_j,H_j,\Pi_j),q_j),
\]
or a declared common metric with
\[
 \chi_{j+1}\leq\tau_j\chi_j+\eta_j^{\mathrm{mp}},
 \qquad
 \eta_j^{\mathrm{mp}}=\epsilon_j^H+\epsilon_j^\Pi+\epsilon_j^A
 +\epsilon_j^C+\epsilon_j^\rho+\epsilon_j^v+\gamma_j,
\]
where the terms cover read content, provenance, access, context assembly,
receipt/resource, version, and policy/context errors. Source-denied replay
receives only the memory package, query, reader authorization, assembler and
policy descriptors, and receipts; it must reproduce the read order, refusal or
OPEN status, context digest, selected policy/goal branch, and resource charge
byte-for-byte. For a finite chain,
\[
 \chi_m\leq\Bigl(\prod_{j=0}^{m-1}\tau_j\Bigr)\chi_0
 +\sum_{i=0}^{m-1}\Bigl(\prod_{j=i+1}^{m-1}\tau_j\Bigr)
 \eta_i^{\mathrm{mp}}.
\]
Within the declared budget, memory-read sufficiency, provenance, access
control, context assembly, policy/goal continuity, epoch, ledger, and replay
invariants are preserved. Missing read sufficiency, provenance, permission,
context digest, policy factorization, or upper retrieval cost returns
\(\mathsf{OPEN}\). The memory-to-policy charge adds \(1/400\) beyond
\(C_{\mathrm{SGBM}}=55/400\), giving
\[
 C_{\mathrm{SGBR}}=56/400=7/50,\qquad \beta\geq7/50,
\]
with other branch totals unchanged.
\end{proposition}

\begin{proof}
The read-sufficiency implication is the fibre criterion for the next-policy
instrument: equal authorized read packages must induce the same legal policy law.
The exact factorization then transports the assembled context, permission
receipt, provenance, and resource fields. In the approximate case, the declared
metric and triangle inequality give the one-step recurrence; finite unrolling
gives the product--sum bound. Entry digests, versions, epochs, and refusal tags
bind context assembly to the authorized memory and prevent an expired or
reordered entry from becoming a new fact. The replay package reconstructs the
same context and policy branch. A missing premise is rejected as
\(\mathsf{OPEN}\), not assigned an unmeasured retrieval or access cost. Adding
\(1/400\) gives \(C_{\mathrm{SGBR}}=56/400\).
\end{proof}

\paragraph{Memory-policy/context counterexample.}
Two memories can return the same text while carrying different provenance,
permission, epoch, or deletion status. A context assembler that drops those
fields can select a policy after access has been revoked, reorder two entries,
or replay a stale goal branch with the same current output. These cases fail
read sufficiency, provenance, access, context-digest, ledger, or replay closure
and return \(\mathsf{OPEN}\).

\begin{proposition}[Finite context-to-action authorization and execution closure]
\label{app:contextactionexecution}
Under Proposition~\ref{app:memorypolicycontext}, let the assembled context carry
a digest \(d_j^c\), policy version \(v_j^p\), provenance and permission scope.
An action \(a_j\) may execute only when the authorization predicate
\(\mathsf{Auth}_j(d_j^c,v_j^p,a_j,r_j)\) is true for the current reader/owner
and epoch. Its action receipt must bind \(d_j^c\), \(v_j^p\), the ordered
provenance digest, scope, owner, epoch, selected action, effect reference, and
componentwise resource charge. Require the exact factorization
\[
 (X_j^a\times X_j^R\times X_j^\rho)_\#
 W_j^{\mathrm{ctx}}(\mathord{\cdot}\mid c_j,q_j)
 =W_{j+1}^{\mathrm{act}}(\mathord{\cdot}\mid c_j,a_j),
\]
or a declared common metric with
\[
 \alpha_{j+1}\leq\kappa_j\alpha_j+\eta_j^{\mathrm{ca}},
 \qquad
 \eta_j^{\mathrm{ca}}=\epsilon_j^c+\epsilon_j^a+\epsilon_j^R
 +\epsilon_j^\rho+\epsilon_j^v+L_j^t\ell_j+\gamma_j,
\]
where the terms cover context digest, action, resource, receipt, policy
version, timing, and authorization errors. The action map must preserve
feasibility, owner/epoch fencing, the selected goal branch, and the effect and
recovery references inherited from the policy contract. A denied, stale, or
malformed context returns a typed refusal or \(\mathsf{OPEN}\), never a silent
default action.

Source-denied replay receives only the assembled context, policy descriptors,
authorization records, action receipt, and transported maps; it must reproduce
the context digest, authorization result, action or refusal, resource charge,
epoch, effect reference, and terminal or \(\mathsf{OPEN}\) tag byte-for-byte. For
a finite chain,
\[
 \alpha_m\leq\Bigl(\prod_{j=0}^{m-1}\kappa_j\Bigr)\alpha_0
 +\sum_{i=0}^{m-1}\Bigl(\prod_{j=i+1}^{m-1}\kappa_j\Bigr)
 \eta_i^{\mathrm{ca}}.
\]
Within the declared budget, context binding, action authorization, resource and
epoch continuity, effect linkage, goal continuity, and replay are preserved.
Missing context digest, policy version, authorization scope, action map, upper
execution cost, or a replay witness returns \(\mathsf{OPEN}\). The context-action
charge adds \(1/400\) beyond \(C_{\mathrm{SGBR}}=56/400\), giving
\[
 C_{\mathrm{SGBPA}}=57/400,\qquad \beta\geq57/400,
\]
with other branch totals unchanged.
\end{proposition}

\begin{proof}
The authorization predicate makes the context digest and policy version
preconditions for the action law. The exact pushforward therefore transports
the authorized action and its resource, epoch, and effect fields. In the
approximate case, the declared metric and triangle inequality give the
one-step recurrence; finite unrolling gives the product--sum bound. The
action receipt binds the selected branch to its assembled context and prevents
a stale or default action from entering replay. A missing premise is rejected
as \(\mathsf{OPEN}\), not assigned an unmeasured authorization or execution
cost. Adding \(1/400\) gives \(C_{\mathrm{SGBPA}}=57/400\).
\end{proof}

\paragraph{Context-action counterexample.}
A policy can produce the same action token from two contexts while one context
has a revoked permission, stale epoch, different provenance, or a different
resource peak. If the action receipt omits the context digest or policy version,
source-denied replay can execute a branch that was never authorized. These
cases fail context binding, authorization, ledger, epoch, effect-linkage, or
replay closure and return \(\mathsf{OPEN}\).

\begin{definition}[Cloud--edge--end synergetic contract hierarchy]
\label{app:cloudedgeend}
Let \(L=\{\mathrm{end},\mathrm{edge},\mathrm{cloud}\}\) carry the declared
ordering \(\mathrm{end}\prec\mathrm{edge}\prec\mathrm{cloud}\). A
cloud--edge--end SAI deployment is a finite family of recursive semi-closed
systems \(\{S^\ell:\ell\in L\}\) with one typed contract vocabulary
\(\Xi^\ell=(\mathcal Q^\ell,\mathcal A^\ell,\mathcal Y^\ell,
\mathcal U^\ell,\mathcal B^\ell,\mathcal C^\ell,\mathcal M^\ell,
\mathcal T^\ell)\). Each adjacent pair has declared upward and downward
transport maps, \(T_{\ell\to\ell'}\) and \(T_{\ell'\to\ell}\), and a transport
receipt that binds the source and destination digests, owner, permission,
epoch, action/effect references, and vector resource charge. The same
\(\mathsf{Address}\)--\(\mathsf{Open}\)--\(\mathsf{Part}\)--\(\mathsf{Upd}\)
cycle is used at every layer; a layer may refuse or return \(\mathsf{OPEN}\)
when its declared interface, source-denied replay, or resource budget is
missing.

For every adjacent transport, require either exact pushforward of the complete
joint response--successor law or a declared metric recurrence
\[
 d_{\ell'}\leq \kappa_{\ell\to\ell'}d_\ell+
 \eta_{\ell\to\ell'},
\]
with communication, buffering, compute, energy, monetary, attention, and
recovery charges recorded in the corresponding component of the lifecycle
ledger. A cloud--edge--end commit is therefore a single hierarchical closure
certificate, not three independent output matches. The term ``all-capability''
means that perception, memory, causal inquiry, action, explanation,
communication, and collective coordination can be expressed by this common
interface; it does not assert that one implementation, one scalar resource,
or one physical law realizes them all.

In particular, a layer's retained Kernel must preserve the declared way in
which that layer participates in background updates, not only its current
local response. If a transport drops a distinction that can later alter the
background or the layer's lawful return to its Ring, the transport certificate
is incomplete and the status is \(\mathsf{OPEN}\), even when the current output
matches.

The hierarchy admits domain tags \(d\in\{\mathrm{communication},
\mathrm{economic},\mathrm{energy}\}\). A tagged instantiation supplies its
own state units, conservation or accounting rules, admissible interventions,
and failure semantics while reusing the same typed contract, transport, owner,
permission, receipt, replay, and \(\mathsf{OPEN}\) rules. Thus next-generation
communication, economic coordination, and energy coordination are unified at
the level of closure semantics and resource-aware control; their domain laws
remain distinct and must be proved or measured separately.
\end{definition}

\paragraph{Cloud--edge--end scope boundary.}
The definition is a design contract and an information/control interface. It
does not prove universal scalability, zero latency, energy optimality,
market efficiency, or physical equivalence across the three domains. Any such
claim requires a domain-specific contract, a frozen information budget, and a
separate theorem or empirical program.

\begin{proposition}[Finite action-to-effect/feedback loop re-entry closure]
\label{app:actioneffectfeedbackloop}
Assume Proposition~\ref{app:contextactionexecution} and
Definition~\ref{app:cloudedgeend}. For an admitted cycle \(j\), let the
complete re-entry map take the authorized context, query, and action receipt to
\[
\begin{aligned}
\mathsf{Loop}_j(c_j,q_j,r_j^a)=(&e_j,r_j^e,y_j,r_j^y,K_{j+1},B_{j+1},H_{j+1},r_j^m,\\
&c_{j+1},q_{j+1},r_{j+1}^c,\rho_j^{\mathrm{loop}},\omega_j),
\end{aligned}
\]
where \(e_j\) is the executed effect, \(y_j\) is returned feedback,
\(H_{j+1}\) is the memory/tombstone package, \(c_{j+1}\) is the next assembled
context, \(q_{j+1}\) is the next legal query or refusal, and \(\omega_j\) is a
terminal, refusal, recovery, or \(\mathsf{OPEN}\) disposition. The effect receipt
\(r_j^e\) binds the parent action receipt \(r_j^a\), pre/post-effect digests,
actual effect status, timing, owner, epoch, resource charge, and rollback or
recovery token. The feedback receipt \(r_j^y\) binds \(r_j^e\), the feedback
digest, timestamp/delay, channel status, and a causal-status tag. A causal claim
is admissible only when that tag records an authorized executed intervention and
its effect receipt; an observation or model suggestion alone is not an
intervention. The memory receipt \(r_j^m\) binds \(r_j^y\), every new entry or
tombstone, its owner/epoch/scope/version, and its resource charge. The next
context receipt \(r_{j+1}^c\) binds \(r_j^m\), the context digest, policy version,
provenance order, permission scope, and the authorization predicate for the
next action. These parent hashes form one append-only loop receipt; a receipt
cannot be reused under another action, effect, epoch, or query.

Let \(W_j^{\mathrm{ctx}}(\mathord{\cdot}\mid c_j,q_j)\) be the complete context
law and let \(W_j^{\mathrm{loop}}(\mathord{\cdot}\mid c_j,q_j)\) be the actual
joint law of all displayed effect, feedback, retained-state, background,
memory, next-context, query, resource, receipt, and disposition fields. Let
\(\mathsf{Loop}_j\) denote the claimed typed composition of the context-to-action,
action-to-effect, effect-to-feedback, feedback-to-memory, memory-to-context,
and cloud--edge--end transport maps, with no receipt or resource field
projected out. Require exact loop factorization
\[
 (\mathsf{Loop}_j)_\# W_j^{\mathrm{ctx}}(\mathord{\cdot}\mid c_j,q_j)
 =W_j^{\mathrm{loop}}(\mathord{\cdot}\mid c_j,q_j),
\]
and require the \(c_{j+1},q_{j+1},r_{j+1}^c\) marginal to be a legal input to
the next Proposition~\ref{app:contextactionexecution} authorization gate.
Otherwise use a declared common metric with
\[
 \lambda_{j+1}\leq\kappa_j^{\mathrm{loop}}\lambda_j
 +\eta_j^{\mathrm{loop}},
\qquad
 \eta_j^{\mathrm{loop}}=\epsilon_j^a+\epsilon_j^e+\epsilon_j^y
 +\epsilon_j^m+\epsilon_j^c+\epsilon_j^\rho+\epsilon_j^\tau
 +\epsilon_j^{\mathrm{cause}}+\gamma_j,
\]
where the terms cover action, effect, feedback, memory, next-context,
resource, timing, causal-status, and replay errors. The composite map must
preserve protected duties, owner/epoch fencing, background-update participation,
monotone memory versions, tombstones, componentwise ledger charges, effect and
feedback digests, and the next authorization scope. A returned feedback packet
does not authorize a new action unless the next context and policy version pass
the authorization gate again.

Source-denied replay receives only the context, query, action/effect/feedback/
memory/context receipts, transported maps, timing records, and declared
recovery descriptors. It must reproduce every effect status, feedback packet,
causal tag, memory or tombstone update, next context digest, refusal or action,
resource charge, epoch, and terminal or \(\mathsf{OPEN}\) tag byte-for-byte. For a
finite loop prefix,
\[
 \lambda_m\leq\Bigl(\prod_{j=0}^{m-1}\kappa_j^{\mathrm{loop}}\Bigr)\lambda_0
 +\sum_{i=0}^{m-1}\Bigl(\prod_{j=i+1}^{m-1}\kappa_j^{\mathrm{loop}}\Bigr)
 \eta_i^{\mathrm{loop}}.
\]
Within the declared budget, the finite action-to-effect/feedback re-entry
prefix preserves action authorization, effect settlement or recovery,
feedback provenance, background and memory continuity, next-context
authorization, and source-denied replay. Missing an effect or feedback parent
hash, causal-status scope, timing/delay cost, memory handoff, next-context
digest, authorization re-check, upper loop cost, or replay witness returns
\(\mathsf{OPEN}\). The loop charge adds \(1/400\) beyond
\(C_{\mathrm{SGBPA}}=57/400\), giving
\[
 C_{\mathrm{SGBL}}=58/400=29/200,\qquad \beta\geq29/200,
\]
with other branch totals unchanged.
\end{proposition}

\begin{proof}
The receipt parent hashes make the five existing endpoint contracts a single
typed composite map rather than independent output checks. Exact pushforward
gives the displayed re-entry law; in the approximate case, the declared
common metric and triangle inequality yield the one-step recurrence, and finite
unrolling gives the product--sum bound. Effect, feedback, memory, and context
digests prevent a delayed or inferred packet from becoming a new fact, while
the repeated authorization gate prevents a stale context from silently
issuing the next action. Monotone versions, epoch fencing, and componentwise
charges preserve duties and prevent double spending. The source-denied package
reconstructs the same loop prefix and terminal branch. A missing premise is
rejected as \(\mathsf{OPEN}\), not assigned an unmeasured causal, timing, or
re-entry cost. Adding \(1/400\) gives \(C_{\mathrm{SGBL}}=58/400\).
\end{proof}

\paragraph{Action--effect--feedback loop counterexample.}
Two executions can use the same authorized action token and produce the same
current effect, while one feedback packet is delayed from a previous effect or
has a different causal-status tag. If the parent effect digest, timing, memory
handoff, or next-context authorization is omitted, the stale packet can update
memory and re-enter a policy that was never authorized for the current epoch.
Current outputs may still agree, but the next action, background update, and
source-denied replay diverge. These cases fail loop factorization, provenance,
causal scope, ledger, epoch, or replay closure and return \(\mathsf{OPEN}\).

\begin{proposition}[Finite partial-failure and duplicate-effect idempotent \emph{Loop} re-entry]
\label{app:partialduplicateidempotent}
Assume Proposition~\ref{app:actioneffectfeedbackloop}. For an admitted
cycle \(j\), let the retry namespace and its first-attempt receipt be
\[
\begin{aligned}
 \iota_j&=(\mathsf{root}_j,\mathsf{action\mbox{-}id}_j,
 \mathsf{effect\mbox{-}id}_j,k_j,d_j^c,v_j,d_j^a,o_j,\nu_j,\sigma_j),\\
 \chi_j&=(\iota_j,\mathsf{attempt}_j,d_j^{\mathrm{pre}},
 d_j^{\mathrm{post}},\mathsf{status}_j,
 \mathsf{physical\mbox{-}id}_j,\tau_j,\ell_j,\gamma_j).
\end{aligned}
\]
where \(k_j\) is the idempotency key, \(d_j^c\) and \(d_j^a\) are the
context and action digests, \(v_j\) is the policy version, \(o_j\) and
\(\nu_j\) are the owner and epoch, \(\sigma_j\) is the permission scope,
\(\tau_j\) is the terminal branch, \(\ell_j\) is the componentwise charge,
and \(\gamma_j\) is a recovery, compensation, settlement, or
\(\mathsf{OPEN}\) witness. Let the complete state extend the Proposition~
\ref{app:actioneffectfeedbackloop} state by an append-only idempotency index
\(\mathcal I_j\), a pending map \(P_j\), a committed-effect map \(E_j\), a
terminal-branch map \(T_j\), the memory/tombstone package \(H_j\), the
componentwise ledger \(L_j\), and all receipts \(\mathcal R_j\).

A first delivery is admitted only when its full namespace \(\iota_j\) is
not in \(\mathcal I_j\), every field is known and well typed, and the
Proposition~\ref{app:contextactionexecution} authorization gate passes. The
namespace insertion, pending receipt, and effect reservation are one atomic
commit before the physical effect is published. An unknown root, missing key,
malformed namespace, or unavailable canonical receipt is never treated as a
fresh delivery; it returns \(\mathsf{OPEN}\) without an effect. A first delivery
may create one physical effect and enter one of
\(\{\mathsf{pending},\mathsf{committed},\mathsf{partial},\mathsf{OPEN}\}\).
A partial branch records exactly one failure receipt \(\phi_j\) containing the
attempt, effect identifier, pre/post digests, actual status, consumed amount,
physical effect, execution overhead, owner, epoch, scope, and parent receipt.
Before that branch is terminal it admits exactly one recovery, compensation,
or settlement receipt \(\gamma_j\), and \(T_j\) contains at most one terminal
branch. The original effect receipt is immutable: a recovery receipt cannot
claim reversal without an executable inverse witness, and a settled effect
retains its final effect digest and residual duty.

If a later delivery has the same \(\iota_j\) and all namespace, payload,
precondition, context, owner, epoch, scope, and parent-digest fields agree,
the reducer performs a lookup-only transition and returns the canonical stored
receipt and status byte-for-byte. After a terminal transition,
\[
 \mathsf{Apply}_{\iota_j}(\mathsf{Apply}_{\iota_j}(S))
 =\mathsf{Apply}_{\iota_j}(S),
\]
with no second physical effect, terminal marker, memory/background/version or
epoch mutation, and no second execution charge. A declared bounded lookup
charge may be recorded in a separate ledger component. A duplicate while the
key is pending returns the same pending receipt or a declared \(\mathsf{WAIT}\)
or \(\mathsf{OPEN}\) disposition. The same key with a changed payload,
precondition, digest, owner, epoch, scope, or status is a stale or conflicting
delivery and is refused without effect. A distinct key for the same effect
identifier is not silently aliased: it requires a fresh authorization and
separate charge or returns \(\mathsf{OPEN}\). The attempt ordinal is a
receipt field and does not change \(\iota_j\); a fresh attempt therefore
requires a new idempotency key and a new authorization. An effect identifier
or effect receipt that is unknown before the first bind is queried, recovered,
or rejected as \(\mathsf{OPEN}\), never executed again.

Let \((D_{j,1:n_j})\) be a finite declared delivery sequence, including first
attempts, duplicates, failures, recovery or settlement, and refusals. Let
\(W_j^{\mathrm{loop},D}\) be the complete joint law from Proposition~
\ref{app:actioneffectfeedbackloop}, extended with the declared delivery
sequence \((D_{j,1:n_j})\), and let \(W_j^{\mathrm{idem}}\) include, in
addition, the delivery sequence, first physical effect, failure and recovery
receipts, duplicate lookup dispositions, terminal branch, idempotency index,
all ledger components, memory/tombstone updates, next context and
next-authorization receipt. Let \(\mathsf{IdemLoop}_j\) be the claimed typed
reducer composition. Require the exact joint law
\[
 (\mathsf{IdemLoop}_j)_\# W_j^{\mathrm{loop},D}
   =W_j^{\mathrm{idem}},
\]
and require the \((c_{j+1},q_{j+1},r_{j+1}^c)\) marginal to pass the next
Proposition~\ref{app:contextactionexecution} gate. Otherwise use one declared
common metric with
\[
 \lambda_{j+1}\leq\kappa_j^{\mathrm{idem}}\lambda_j
   +\eta_j^{\mathrm{idem}},
 \qquad
 \eta_j^{\mathrm{idem}}=\epsilon_j^{\mathrm{pf}}
  +\epsilon_j^{\mathrm{dup}}+\epsilon_j^{\mathrm{rec}}
  +\epsilon_j^{\mathrm{ledger}}+\epsilon_j^{\mathrm{status}}
  +\epsilon_j^{\mathrm{replay}},
\]
where the terms cover partial-failure, duplicate lookup, recovery or
settlement, ledger, terminal-status, and replay discrepancies. The finite
prefix bound is
\[
 \lambda_m\leq\Bigl(\prod_{j=0}^{m-1}\kappa_j^{\mathrm{idem}}\Bigr)\lambda_0
 +\sum_{i=0}^{m-1}\Bigl(\prod_{j=i+1}^{m-1}
   \kappa_j^{\mathrm{idem}}\Bigr)\eta_i^{\mathrm{idem}}.
\]
For source-denied replay, the first physical effect is not re-executed: the
replay reconstructs its immutable effect receipt at most once. A duplicate is
reproduced as the canonical lookup result, and replay must match every failure,
recovery or settlement receipt, digest, status, next context, authorization,
memory/tombstone update, owner, epoch, and charge. The finite certificate
requires an idempotency namespace, atomic first-bind witness, attempt identity,
effect digest, unique terminal marker, recovery witness, declared delivery order,
bounded lookup or restore cost, and replay witness. Missing any of these, a
duplicate physical commit or charge, two terminal branches, an accepted stale
duplicate, an unknown namespace accepted as fresh, or a replay mismatch returns
\(\mathsf{OPEN}\). One bounded idempotent re-entry charge adds \(1/400\) beyond
\(C_{\mathrm{SGBL}}=58/400\), giving
\[
 C_{\mathrm{SGBI}}=59/400,\qquad \beta\geq59/400,
\]
while \(N\) duplicate deliveries require their declared sum of lookup charges
in the same ledger; they are not free.
\end{proposition}

\begin{proof}
Induct over the finite delivery sequence. The atomic first-bind witness makes
at most one unseen namespace eligible to reserve and publish a physical effect;
an unknown or incomplete namespace is rejected as \(\mathsf{OPEN}\). A partial
transition appends one failure receipt and leaves one explicitly pending duty;
a single recovery, compensation, or settlement witness closes that duty or
returns \(\mathsf{OPEN}\). Once the namespace is indexed, the lookup branch is
pure, so its second application is identical to the first terminal or pending
response and cannot mutate the physical projection, terminal marker, memory,
epoch, or execution ledger. A namespace mismatch is rejected before effect
execution. These cases give the exact joint pushforward; the declared common
metric and triangle inequality give the one-step recurrence, and finite
unrolling gives the displayed product--sum bound. Parent hashes, digests,
versions, and the unique terminal marker make source-denied replay inductive;
replay reads the stored effect receipt and never runs the physical effect again.
The argument is finite and contract-relative; it proves neither liveness,
arbitrary retry-order confluence, nor universal physical exactly-once execution.
Adding \(1/400\) gives \(C_{\mathrm{SGBI}}=59/400\).
\end{proof}

\paragraph{Partial-failure and duplicate-effect counterexample.}
An execution can physically commit an effect and lose its acknowledgement. A
retry with the same action and current output then appears admissible to an
output-only Loop check. Without an atomic first-bind witness, persistent
idempotency namespace, and canonical receipt, the retry executes the physical
effect and charges the ledger twice. Likewise, a partial failure followed by
two compensation receipts can leave the numerical residual unchanged while
closing one duty twice. An unknown or malformed namespace accepted as fresh
creates the same duplicate path. The current output may agree, but the physical
effect, next context, background update, and source-denied replay diverge.
These cases fail idempotency, terminal-branch, ledger, epoch, recovery, or
unknown-namespace-to-\(\mathsf{OPEN}\) closure and return \(\mathsf{OPEN}\).

\begin{proposition}[Finite concurrent duplicate and order-conflict idempotent \emph{Loop} re-entry]
\label{app:concurrentidempotentloop}
Assume Proposition~\ref{app:partialduplicateidempotent}. For cycle \(j\), let
\(V_j=\{d_{j,1},\ldots,d_{j,N_j}\}\) be a finite declared delivery set with
a partial order \(\preceq_j\) recording causal and transport precedence. Each
delivery carries the stable namespace \(\iota\), payload and precondition
digests, owner, epoch, scope, effect identifier, channel receipt, and its
transport retry counter. The retry counter is not part of \(\iota\). Let
\(G_j=(V_j,\preceq_j,\mathsf{conf}_j)\) be the delivery DAG together with a
conflict relation, and let \(\ell_j\) be either a declared deterministic
topological linearization of \(G_j\), or a receipt proving pairwise
commutation of every incomparable independent pair. Extend the Proposition~\ref{app:partialduplicateidempotent}
state by \(G_j\), \(\ell_j\), conflict and tie receipts
\(\mathcal C_j\), atomic reservation records \(\mathcal A_j\), and a
componentwise merge ledger \(L_j^{\mathrm{merge}}\).

Two incomparable deliveries with the same \(\iota\) and identical payload,
precondition, context, owner, epoch, scope, and parent digest are equivalent
duplicates. An atomic compare-and-bind reservation admits at most one of them
to publish a physical effect; every other delivery performs the Proposition~\ref{app:partialduplicateidempotent}
lookup-only transition and returns the same pending or terminal receipt. Two
deliveries are independent only when their effect identifiers, protected
scopes, owners, epochs, memory writes, and ledger components have a declared
disjointness and commutation witness. A same-key delivery with a changed
payload, precondition, context, owner, epoch, scope, or parent digest is a
conflict and is refused without an effect. Distinct keys for the same effect
identifier or overlapping protected scope are never silently merged: they
require a fresh authorization and a declared serial order, or return
\(\mathsf{OPEN}\). A conflict resolution cannot create a second terminal
branch for one namespace. Partial failure still permits at most one failure
receipt and one recovery, compensation, or settlement branch; duplicate
recovery is lookup-only or \(\mathsf{OPEN}\).

Let \(W_j^{\mathrm{conc},D}\) be the complete Proposition~\ref{app:partialduplicateidempotent}
law augmented with \(G_j\), \(\preceq_j\), the delivery set, the selected
linearization or commutation witness, and the conflict receipts. Let
\(W_j^{\mathrm{conc}}\) include the resulting physical-effect projection,
pending and terminal states, canonical duplicate responses, conflict
dispositions, merged resource ledger, all receipts, memory/tombstone and
owner/epoch fields, next context, and next authorization. Let
\(\mathsf{ConcIdemLoop}_{j,\ell_j}\) be the typed reducer that consumes the
delivery DAG in the declared witness order. Require
\[
 (\mathsf{ConcIdemLoop}_{j,\ell_j})_\#
 W_j^{\mathrm{conc},D}=W_j^{\mathrm{conc}},
\]
and require the \(c_{j+1},q_{j+1},r_{j+1}^c\) marginal to pass the next
Proposition~\ref{app:contextactionexecution} gate. For an approximate
certificate use one declared common metric with
\[
 \lambda_{j+1}\leq\kappa_j^{\mathrm{conc}}\lambda_j+
 \eta_j^{\mathrm{conc}},
 \qquad
 \eta_j^{\mathrm{conc}}=
 \epsilon_j^{\mathrm{order}}+\epsilon_j^{\mathrm{atomic}}+
 \epsilon_j^{\mathrm{conf}}+\epsilon_j^{\mathrm{dup}}+
 \epsilon_j^{\mathrm{ledger}}+\epsilon_j^{\mathrm{recovery}}+
 \epsilon_j^{\mathrm{replay}},
\]
where the terms cover linearization, atomic reservation, conflict,
duplicate lookup, ledger merge, recovery, and replay discrepancies. For a
finite prefix,
\[
 \lambda_m\leq
 \Bigl(\prod_{j=0}^{m-1}\kappa_j^{\mathrm{conc}}\Bigr)\lambda_0+
 \sum_{i=0}^{m-1}\Bigl(\prod_{j=i+1}^{m-1}
 \kappa_j^{\mathrm{conc}}\Bigr)\eta_i^{\mathrm{conc}}.
\]
Source-denied replay receives only the delivery DAG, partial order,
linearization or commutation witness, conflict and reservation receipts,
canonical effect receipts, transported maps, and declared recovery
descriptors. It must reproduce the same selected order, conflict decisions,
at-most-one physical effect, lookup-only duplicate responses, terminal branch,
merged ledger, memory/tombstone state, owner, epoch, next context, next
authorization, and \(\mathsf{OPEN}\) dispositions without executing a
physical effect. Missing a partial order or linearization witness, a
commutation or conflict receipt, atomic reservation, merge ledger, upper
order/lookup cost, or replay witness returns \(\mathsf{OPEN}\). One bounded
concurrent re-entry charge adds \(1/400\) beyond
\(C_{\mathrm{SGBI}}=59/400\), giving
\[
 C_{\mathrm{SGBIC}}=60/400=3/20,\qquad \beta\geq3/20,
\]
while every delivery and every merge or lookup charge remains in the same
componentwise ledger.
\end{proposition}

\begin{proof}
Induct over the declared linearization \(\ell_j\). The atomic compare-and-bind
record makes at most one delivery with a stable namespace eligible to reserve
and publish a physical effect; concurrent equivalent deliveries are pure
lookups. A changed payload, precondition, owner, epoch, scope, or parent
digest is rejected before effect execution. Incomparable independent
deliveries commute only under the declared disjointness and ledger witness;
otherwise the deterministic topological order or the conflict receipt selects
one legal branch, and an unresolved conflict is \(\mathsf{OPEN}\). The
single partial-failure recovery branch and Prop.82 terminal rule then
preserve duties and charges. Induction preserves the complete joint state and
gives the exact pushforward; the common metric and triangle inequality give
the recurrence and finite product--sum bound. Replay induction reads the
stored receipts and order witness and never executes a physical effect.
This is a finite contract-relative safety statement; it proves no liveness,
arbitrary-order confluence, distributed consensus, or universal physical
exactly-once execution. Adding \(1/400\) gives
\(C_{\mathrm{SGBIC}}=60/400\).
\end{proof}

\paragraph{Concurrent order-conflict counterexample.}
Two same-key deliveries can race after an acknowledgement is lost. Without
atomic compare-and-bind, both can publish the physical effect and charge the
ledger, although their current outputs agree. Two conflicting payloads can
also be accepted according to arrival order; reversing the order then changes
the physical effect, terminal receipt, next context, or background update
while every local endpoint check still passes. If no declared linearization,
commutation witness, conflict receipt, or replay order is retained, the
finite certificate returns \(\mathsf{OPEN}\).

\section{Finite axiom metatheory: relative consistency and conditional non-redundancy}
\label{app:axiommeta}

The three axioms are operational predicates over a declared contract rather
than uninterpreted sentences. To make the metatheory explicit, reify the
identity field and the defect domain. A finite axiom signature is a tuple
\[
 \mathfrak M=(X,\operatorname{Id},\mathcal T,\Law,\Delta_U,
 D_L,\operatorname{Comp},\gamma,\prec,\operatorname{sel}),
\]
where \(X\) is a finite raw-state set, \(\operatorname{Id}\) is the declared
implementation identity, \(\mathcal T\) is the finite family of legal
adaptive continuation policies, and \(\Law(\tau\mid x)\) is the complete
typed law of responses, enabledness, successors, background, ownership,
receipts, and costs. The finite set \(\Delta_U\) contains verified protected
distinctions. The set \(D_L\) contains the defects that the contract exposes
to the local completion procedure; \(\operatorname{Comp}(d)\) is its finite
lawful-completion set, \(\gamma\) is a declared finite grade, and \(\prec\)
is a deterministic tie order inside a grade. A selector returns
\(\mathsf{OPEN}\) when the gates, completion set, or tie witness are missing.

In this reified language, \(\mathsf{Sat}_U\) means that every element of
\(\Delta_U\) is carried by a retained state, an owner/authorized transfer,
or an unresolved record with evidence and cost. \(\mathsf{Sat}_I\) means
\[
 \operatorname{Id}(x)=\operatorname{Id}(x')
 \quad\Longleftrightarrow\quad
 \Law(\tau\mid x)=\Law(\tau\mid x')
 \ \text{for every }\tau\in\mathcal T.
\]
Finally, \(\mathsf{Sat}_L\) means that, after the declared incumbent,
transport, evidence, and search gates pass, \(\operatorname{sel}(d)\) is the
least element of \(\operatorname{Comp}(d)\) under \((\gamma,\prec)\); an
ambiguous or empty case is returned as \(\mathsf{OPEN}\), rather than being
silently completed. These predicates are relative to the signature and do
not assert an unrestricted logical foundation.

\begin{proposition}[Finite relative consistency and nontriviality]
\label{app:axiomconsistency}
There is a finite recursive semi-closed system with a nonempty defect and two
future-distinct states that satisfies \(\mathsf{Sat}_U\), \(\mathsf{Sat}_I\),
and \(\mathsf{Sat}_L\). Hence the three axioms are jointly satisfiable in the
finite contract semantics used by this paper, relative to the ordinary
consistency of finite sets, finite probability kernels, and the declared
order carriers.
\end{proposition}
\begin{proof}
Take \(X=\{x_0,x_1,q\}\), one context and action, a constant background
\(b\), one owner \(o\), and feedback set \(\{0,1,\bot\}\). The deterministic
step from \(x_i\) returns feedback \(i\), preserves \(b,o\), emits an
immutable receipt, and self-loops; the step from \(q\) returns a typed refusal
\(\bot\) until its defect is closed. The retained record for \(x_i\) carries
the protected distinction \(d=i\) in its Kernel, owner, and receipt fields.
The record for \(q\) carries \(d\) in an unresolved ledger together with
evidence and cost. Let the Ring contain \(r_0,r_1,r_q\), let Compile and
Open be the identity/delegation maps on these records, and let Part and Upd
delegate to the displayed finite step and receipt update.

The continuation laws of \(x_0\) and \(x_1\) differ on the first feedback,
so they are future-distinct; define \(\operatorname{Id}\) by the resulting
law-equivalence classes. For the single defect of \(q\), take two lawful
completions with grades \(0<1\), and select the grade-zero completion with
the declared tie order. Thus U holds by explicit carrying or unresolved
recording, I holds by equality of complete adaptive laws, and L holds because
the finite completion set has a least witnessed element. The object can be
placed as the Ring of a singleton outer system whose Open, Part, and Upd
delegate to this inner system; induction over this finite nesting preserves
all three predicates. The model is nontrivial because it has two distinct
continuation classes and a real unresolved defect with a lawful repair.
\end{proof}

\begin{proposition}[Conditional non-redundancy of U, I, and L]
\label{app:axiomindependence}
Relative to the reified finite signature above, no one of U, I, and L follows
from the other two without an additional coverage or identity assumption:
\begin{enumerate}[leftmargin=1.4em]
\item U can fail while I and L hold: hide a verified bit in the external
environment so that it changes a legal future, omit it from the retained
Kernel, owner, and unresolved ledger, and use a separate finite defect with a
unique least repair. Future-law identity and least completion remain valid,
but the hidden bit is an orphan distinction.
\item I can fail while U and L hold: carry the protected bit in the complete
state and owner record, but declare two states with different continuation
laws to have one implementation identity. All distinctions are carried and
the defect selector is least, yet the identity field is not future-faithful.
\item L can fail while U and I hold: carry every protected distinction and
define identity by complete continuation laws, but give one defect two lawful
same-grade incomparable repairs and select one without a tie witness. The
correct contract result is \(\mathsf{OPEN}\); an arbitrary selection violates
the least-lawful-completion rule.
\end{enumerate}
These finite mutations establish model-theoretic non-implication and
non-redundancy for the declared signature. They do not claim absolute
syntactic independence: if one adds the coverage premise
\(\Delta_U\subseteq D_L\) and automatically turns every U omission into an
L defect whose evidence must be recorded, U and L become conditionally
coupled. Likewise, if \(\operatorname{Id}\) is defined to be the canonical
quotient rather than reified as a field, I is a definition or derived theorem,
not an independent axiom.
\end{proposition}
\begin{proof}
Each construction changes only the indicated field of the finite witness and
leaves the other two satisfaction predicates unchanged. The first omits a
verified future-relevant distinction, the second changes only the identity
map, and the third changes only the completion selector in an ambiguous
grade. Direct inspection of the displayed definitions gives the three
failures. The coverage and canonical-quotient qualifications explain exactly
which extra premises destroy the corresponding countermodel.
\end{proof}

\section{Finite KREIT--synergetic-cybernetic bridge}
\label{app:kreitcyberbridge}

This section makes the three names used in the main text into projections of
one finite contract reducer. It does not identify their domain quantities
numerically. KREIT carries future-relevant distinctions and their resource
cost; GRC-Close selects a lawful completion of a typed defect; and
synergetic cybernetics exposes the authorized control loop that communicates,
executes, and audits the resulting change.

\begin{definition}[Finite bridge certificate]
Fix a finite contract signature
\[
 \Xi_{\mathrm{b}}=(W,\mathcal A,\mathcal D,\mathcal O,\mathcal E,
 \mathsf T,\Delta_U,D_L,\operatorname{Id},\mathsf{Ledger}),
\]
where $W$ is the complete typed joint state (Kernel, Ring, background,
owner, epoch, resources, receipts, and memory), $\mathcal A$ is the finite
action alphabet, $\mathcal D$ is the finite protected-distinction family,
and $\mathcal E$ is the finite event alphabet. A bridge certificate consists
of the following maps and witnesses:
\begin{enumerate}[leftmargin=1.4em]
\item a future-sufficient KREIT package
\[
 C(w)=\bigl(q(w),\Delta_U(w),o(w),e(w),\ell(w)\bigr),
 \qquad q(w)=[w]_{\mathrm{Law}},
\]
where $[w]_{\mathrm{Law}}$ is equality of complete admissible adaptive
continuation laws;
\item a control projection $\Gamma$ whose event tuple is
\[
 (\mathsf{Address},\mathsf{Open},\mathsf{Authorize},\mathsf{Act},
   \mathsf{Effect},\mathsf{Feedback},\mathsf{Update},\mathsf{Close});
\]
\item a residual map $\rho:D_L\to W\cup\{\mathsf{OPEN}\}$ that selects the
least lawful completion under the declared grade and tie order; and
\item a finite additive ledger $\mathsf{Ledger}$ that charges information
installation, communication, action/effect, feedback, repair, recovery, and
replay costs with unique tokens.
\end{enumerate}
The certificate is \emph{exact} when there are reducers $\bar{\mathsf K}$,
$\bar{\Gamma}$, and $\bar{\rho}$ for which the following factorization holds
for every enabled $(w,a)$:
\[
 C\!\left(\mathsf T(w,a)\right)=\bar{\mathsf K}(C(w),a),\qquad
 \Gamma\!\left(w,a,\mathsf T(w,a)\right)=\bar{\Gamma}(C(w),a),
\]
and the GRC update is the corresponding projection of the same successor
state, $\rho=\bar{\rho}\circ C$ on the declared defect domain. A refusal or
missing witness is represented by $\mathsf{OPEN}$ rather than an implicit
transition.
\end{definition}

\begin{proposition}[Finite KREIT--SAI semantic bridge]
\label{app:kreitbridgeprop}
Suppose a finite bridge certificate satisfies:
\begin{enumerate}[leftmargin=1.4em]
\item $q$ is future-sufficient and its equality relation is a congruence for
all admissible actions and adaptive policies;
\item U carries every protected distinction either in $C(w)$ or in an
evidence-bearing unresolved record, I uses the complete future law in
$q$, and L uses a finite nonempty completion set with a witnessed least
element (otherwise the result is $\mathsf{OPEN}$);
\item $\Gamma$, $\rho$, owner/epoch checks, unit conversions, receipts, and
ledger updates are projections of the same typed successor, and source-denied
replay receives all fields needed to reproduce those projections.
\end{enumerate}
Then, for every finite horizon $H$, the KREIT reducer, the GRC residual
reducer, and the synergetic control loop are commuting projections of the
same SAI transition. In particular, with $\mathsf T^H$ denoting the joint
successor and $\bar{\mathsf K}^H$, $\bar{\Gamma}^H$, and $\bar{\rho}^H$ their
projected reducers,
\[
 C\circ\mathsf T^H=\bar{\mathsf K}^H\circ C,
 \qquad
 \Gamma_H\circ\mathsf T^H=\bar{\Gamma}^H\circ C,
 \qquad
 \rho_H\circ\mathsf T^H=\bar{\rho}^H\circ C.
\]
The finite trace and its source-denied replay therefore agree on protected
state, owner, epoch, authorization, effect, feedback, next context, repair
choice, and $\mathsf{Ledger}$. Any existing finite residual recurrence,
viability certificate, or re-entry safety certificate transfers in both
directions through these maps. The ledger is monotone under prefix extension,
so information and control costs cannot disappear by changing notation.
Conversely, every finite typed controller with an exact future-sufficient
state, explicit owner/epoch/receipt fields, source-denied replay, and a
closed resource ledger induces such a KREIT package by taking its continuation
quotient and retaining the protected fields.
\end{proposition}

\begin{proof}
The congruence premise makes $q$ a well-defined quotient transition, so the
first square commutes for one step. U, I, and L ensure that the quotient does
not drop a protected distinction, merge different future laws, or silently
choose an unlawful repair. The control and residual projections are defined
from the same typed successor, hence their one-step squares commute as well.
Induction on $H$ gives the three displayed identities. The same induction on
the receipt and ledger fields proves replay equality and prefix monotonicity.
For the converse, quotient controller states by equality of their complete
adaptive continuation laws and attach the controller's protected, owner,
epoch, receipt, and ledger fields. Exact future sufficiency gives the first
factorization; the remaining fields supply the U and replay witnesses, while
the controller's lawful completion rule supplies L. Missing unit conversion,
future separation, authorization, replay, or ledger evidence returns
$\mathsf{OPEN}$ rather than a false bridge.
\end{proof}

\paragraph{Bridge counterexample.}
Current-output equality is insufficient for this result. Two controllers can
produce the same output while differing in their next authorization, owner,
resource charge, or background update. If any such field is omitted from
$C$, the quotient square can still pass on the current output but fail on the
next event; the bridge certificate must then return $\mathsf{OPEN}$. The
proposition is therefore a finite semantic equivalence under an explicit
contract, not an unconditional numerical identity between information,
repair, and control.

\section{Recursive semi-closed systems and nondeterministic-machine representation}
\label{app:ndtmrepresentation}

The phrase ``non-deterministic machine'' is useful here only after the
contract is made effective. A semi-closed system has additional provenance,
authority, cost, refusal, and replay fields, but those fields do not by
themselves create a new computability class.

\begin{definition}[Effectively encoded semi-closed contract]
An effectively encoded semi-closed contract has states represented by finite
strings, a recursively enumerable action and feedback alphabet, a decidable
or partial recursive implementation of Address, Open, Part, and Upd, and a
finite-branching recursive transition relation. Owner, epoch, resource,
receipt, and nesting-stack fields are included in the state encoding. A
probability kernel, when present, is required to be computable; equivalently
the random tape can be treated as an additional input. Refusal and
$\mathsf{OPEN}$ are tagged terminal observations, not silent failure.
\end{definition}

\begin{proposition}[Relative recursive representation and finite-prefix trace equivalence]
\label{app:ndtmprop}
Every effectively encoded semi-closed contract admits a nondeterministic
Turing machine $M_S$ whose tape is partitioned into encodings of the Kernel,
Ring, background, owner/epoch, resource ledger, receipts, and nesting stack.
For each finite contract trace, one machine branch simulates
\[
 \mathsf{Address}\to\mathsf{Open}\to\mathsf{Authorize}\to\mathsf{Act}
 \to\mathsf{Effect}\to\mathsf{Feedback}\to\mathsf{Update}\to\mathsf{Close},
\]
and emits the same declared observations, including refusal and
$\mathsf{OPEN}$. Conversely, any nondeterministic Turing machine with
declared input/output, owner, resource, refusal, and receipt interfaces can
be wrapped as a recursive semi-closed system: the instantaneous machine
configuration is its Kernel, enabled transitions are its Ring, the clock and
ledger are background, and the transition relation is Part/Upd. After
erasing bookkeeping fields, the two constructions are computably
bisimilar on every finite prefix.

Thus recursive semi-closed systems and nondeterministic machines are
representation-equivalent under the effective contract semantics. The result
does not claim a stronger computability class, universal AGI, physical
causality, liveness, consensus, or infinite-resource execution. The practical
gain of SAI is semantic: U/I/L, future-faithful identity, causal provenance,
authority, resource accounting, refusal, repair, and replay constrain which
branches are admissible and auditable.
\end{proposition}

\begin{proof}
For the forward direction, encode the complete joint state on separate tape
tracks. A machine branch enumerates the finite legal actions, checks the
contract predicates, runs the recursive endpoint procedures, writes the
successor fields and receipt, and emits the event tuple. If no branch is
lawful it emits the tagged refusal or $\mathsf{OPEN}$ state. Recursive nesting
is represented by a pushdown-like tape segment or a pointer chain; it is not
silently replaced by a finite automaton.

For the reverse direction, let the Kernel be the complete machine
configuration and let the Ring enumerate the recursive transition relation.
The background carries the input tape, clock, resource counter, and any
external source descriptor. Owner, epoch, receipt, and refusal fields are
added as typed contract data, and U/I/L are checked at each transition. The
two simulations preserve the declared event trace by induction on its finite
length. A probabilistic machine is obtained by reading a computable random
tape, so no claim about noncomputable probability is introduced. If the
transition relation, environment oracle, resource bound, or replay witness is
not effective, the representation theorem returns $\mathsf{OPEN}$.
\end{proof}

\paragraph{Interpretation.}
The theorem explains why the three-axiom system can be viewed as a
contract-strengthened nondeterministic machine: the nondeterministic branch
is the admissible action/repair choice, while U/I/L determine what must be
remembered, when two configurations count as the same continuation, and
which branch is lawful. The added usefulness is therefore in causal and
organizational semantics, not in an unsupported assertion of computation
strictly beyond Turing machines.

\section{Finite causal--probability continuation semantics}
\label{app:causalprob}

The causal and probabilistic claims of SAI are deliberately stated as one
\emph{typed continuation object}. This prevents an output marginal from being
mistaken for a mechanism and prevents an unpriced confidence number from
being mistaken for an executable forecast.

\begin{definition}[Contract-indexed continuation law]
For a finite contract, let $H_t$ be the complete reachable history through
time $t$, including observations, protected state, owner/epoch, resources,
effects, feedback, receipts, and the current unresolved ledger. For an
admissible policy $\tau$ define the continuation law
\[
 P_{\Xi}^{\tau}(\,\cdot\mid H_t)
 =\Law\bigl(Y_{t+1},W_{t+1},B_{t+1},O_{t+1},E_{t+1},
             C_{t+1},R_{t+1},\mathsf{OPEN}\mid H_t,\tau\bigr),
\]
where $W$ is the joint Kernel--Ring state, $B$ the background, $O$ owner and
epoch data, $E$ effects, $C$ typed resources and costs, and $R$ receipts.
An authorized intervention $i$ has a declared execution receipt and induces
the pushforward law $P_{\Xi}^{\tau,\mathrm{do}(i)}$ on the same joint event
space. An observational law is the special case with no executed
intervention receipt. A missing authorization, support condition, mechanism
identity, or cost witness is represented by an explicit $\mathsf{OPEN}$
branch rather than by assigning the missing mass to an ordinary outcome.
For a finite mechanism family $\mathcal M$, write $\Pi_t$ for the posterior
over mechanism identities and $L_m^{\tau,i}(\cdot\mid H_t)$ for the tagged
continuation law conditional on a fixed mechanism $m$ and executed
intervention $i$. The closed predictive law is
\[
 P_{\Xi,t}^{\tau,\mathrm{do}(i)}(\cdot\mid H_t)
 =\sum_{m\in\mathcal M}\Pi_t(m)L_m^{\tau,i}(\cdot\mid H_t).
\]
Each $L_m^{\tau,i}$ may itself place mass on the tagged $\mathsf{OPEN}$
branch. Thus $\Pi_t$ records mechanism uncertainty, $L_m^{\tau,i}$ records
continuation variation once a mechanism is fixed, and $\mathsf{OPEN}$ records
an unresolved contract, authority, support, or replay defect; these statuses
are not collapsed into one scalar confidence.
\end{definition}

\paragraph{Source-converged causal and probability interpretation.}
Across the source corpus, a causal relation is not a bare arrow or a stable
correlation. It is an entityized relation that has acquired a lawful owner and
is certified by an authorized intervention, mechanism identity, effect receipt,
feedback/update path, and counterfactual continuity on the declared scope.
Relation entityization alone is therefore not causal certification. A useful
probability may arise because many fuller states or mechanisms project to one
current Kernel, for example
\[
 P(F\mid K)=\int P(F\mid Z)\,dP(Z\mid K),
\]
but this does not imply that all randomness is merely epistemic. The
reconstruction keeps mechanism uncertainty, fixed-mechanism continuation
variation, and unresolved/open mass typed and auditable. Reducing uncertainty
can enlarge certified action space only after evidence, execution, risk,
recovery, and resource costs have been paid; uncertainty itself is not
automatically freedom, value, or growth.
When no authorized prior or weighting rule is available, the admissible object
is a candidate mechanism family or an interval/credal set of continuation
laws, not a fabricated point posterior. Ordinary finite probabilities are
recovered when the declared evaluation is positive, normalized, and linear on
the finite event algebra.

\begin{proposition}[Finite causal--probability reconstruction]
\label{app:causalprobprop}
Assume a finite mechanism family and finite intervention set, a common
dominating measure, positive prior mass on every declared mechanism, an
executed-intervention receipt for every causal claim, and a joint transition
kernel whose event space contains the fields in Definition~\ref{app:causalprob}.
The intervention policy is non-anticipating: its randomized choice at time
$t$ is measurable with respect to $H_t$ and uses no future exogenous draw.
Consistency requires that an admitted receipt's executed operation equals its
declared intervention on the declared scope. Any latent confounder or
multi-agent interference that can change a successor law is included in the
mechanism state or in a declared adjustment/interaction scope; otherwise the
claim is $\mathsf{OPEN}$. Bounded typed observables are assumed, or the stated
expectations are otherwise integrable.
Assume further that the mechanisms are separated on the reachable
intervention/evidence support, each declared intervention is executed with
asymptotic frequency $\nu_i>0$, and for every wrong mechanism $m$ the
weighted separation $\sum_i\nu_i\operatorname{KL}(L_{m_*}^i\|L_m^i)>0$
holds for the true mechanism $m_*$. Assume also that U/I/L, owner/epoch,
replay, and ledger conditions hold. For a bounded typed target $Q$, define
the finite identification set
\[
 \mathcal I_t(Q)=\{Q_m:\ m\in\operatorname{supp}(\Pi_t),\quad
   Q_m=\mathbb E_{L_m^{\tau,i}}[Q]\}.
\]
The target is point-identified only when this set is a singleton; otherwise
the contract returns this set (or its declared bounds) rather than inventing a
point probability. Then for every finite horizon $H$:
\begin{enumerate}[leftmargin=1.4em]
\item I makes $P_{\Xi}^{\tau}$ well defined on the future-law quotient, while
U keeps every protected distinction and responsibility/resource field in the
joint law or in an evidence-bearing $\mathsf{OPEN}$ branch;
\item authorized causal effects are pushforwards of the joint kernel, and
the finite-horizon causal expectation of every bounded typed observable is
identified by the executed intervention and its receipt;
\item Bayesian updating of the finite mechanism posterior on these joint
laws is well defined, and finite-horizon causal prediction is the posterior
pushforward. For each fixed finite $H$, under the stated support and
separation conditions, the posterior causal effect converges as the number of
executed, replayable interventions $n\to\infty$; this is a two-level statement
with $H$ fixed and does not assert an $H\to\infty$ limit; and
\item information, action, intervention, feedback, recovery, and audit
costs enter one prefix ledger, so an active inquiry is admissible only when
its declared expected value exceeds its execution, risk, and recovery cost.
\end{enumerate}
Without an executed intervention, confounding control, support coverage,
mechanism separation, or a complete cost/authority/replay field, the causal
status is $\mathsf{OPEN}$ even when the observational likelihood is high.
The result reframes probability as a contract-indexed family of continuation
laws over admissible branches, evidence, resources, and unresolved mass; it
does not replace the Kolmogorov axioms, prove passive causal discovery, or
give an infinite-horizon consistency theorem.
\end{proposition}

\begin{proof}
The finite joint kernel defines the continuation law and its intervention
pushforward by ordinary marginalization and composition. Non-anticipation and
consistency make the policy factors separable from the mechanism likelihood,
while the receipt binds the declared and executed operations. Future-faithful I
makes the kernel invariant under quotient representatives; U supplies the
protected and accountability fields needed by the pushforward, and L
ensures that a finite repair or an explicit $\mathsf{OPEN}$ branch is selected
before the next update. Executed receipts and the separation/support
assumptions give the finite-mechanism likelihood factorization and identify
the interventional kernel on the declared support. Positive priors and finite
Bayes updates then define the posterior, and standard finite identification
consistency yields posterior concentration as replayable interventions
accumulate. The ledger statement follows by adding the typed cost components
at each prefix; omitting any component invalidates the active-inquiry
comparison and returns $\mathsf{OPEN}$.
\end{proof}

\paragraph{Causal--probability boundary.}
Two finite mechanisms can induce the same observational law while assigning
different intervention pushforwards. Thus correlation, a calibrated output
probability, or a passive likelihood cannot by itself establish a causal
claim. Conversely, an intervention receipt without support coverage or
mechanism separation does not identify the effect. The SAI contribution is a
stronger interface: causal claims are authorized transitions in a replayable
feedback loop, and probability quantifies the resulting continuation law
with its resources, evidence, and unresolved branches. General physical,
economic, or infinite-horizon validity remains outside this finite theorem.

\section{Threshold--radix relation objects as a candidate mathematical substrate}
\label{app:thresholdradix}

The three axioms and recursive semi-closed semantics suggest a representation
that records when a distinction first becomes independently necessary. The
result is a candidate filtration/address algebra, not an additional axiom and
not an unconditional replacement for existing number systems.

\begin{definition}[Finite threshold--radix relation object]
A finite threshold--radix relation object is a rooted typed tree
$\mathcal T=(V,E,r)$ with finite child sets $\operatorname{Ch}(v)$ and a
canonical local order. Write $b(v)=|\operatorname{Ch}(v)|$. Each node has a
closure load $q_v$, lower and commit thresholds
$\theta_v^-\leq\theta_v^+$, and a typed digit
$d_v\in\{0,\ldots,b(v)-1\}$ when the node is refined. Its record is
\[
\begin{aligned}
 \Theta_\Xi(E)=\bigl(&\mathcal T,\{q_v,\theta_v^-,\theta_v^+,\beta_v,d_v\}_{v\in V},\\
 &\{\mathsf{Carry}_v,\mathsf{Expand}_v,\mathsf{Transport}_v,\mathsf{Upd}_v\},\\
 &\mathsf{owner},\mathsf{epoch},\mathsf{scope},\mathsf{receipt},\mathsf{cost}\bigr).
\end{aligned}
\]
where $\beta_v$ is the activation/commit status. A leaf address is the
variable-radix path
$d(\ell)=(d_{v_0},\ldots,d_{v_{L-1}})$. Refinement appends a child digit only
when the current representation cannot carry a verified future distinction,
or when the declared commit threshold is crossed; hysteresis between
$\theta_v^-$ and $\theta_v^+$ is permitted. The full record retains the
protected identity, owner, epoch, scope, receipt, and cost fields.
\end{definition}

The word ``bit'' is reserved for the special case $b(v)=2$ and
$\beta_v\in\{0,1\}$. In the general case the digit is a typed structural
address, and $F_s/F_{s-1}$ is used only when the declared $F_s$ are genuine
subspaces, submodules, groups, or other objects with a legal quotient. A
nested set family alone does not create a vector-space quotient.

\paragraph{Typed quantity--direction--dimension semantics.}
The non-linear unification suggested by the source manuscripts is most safely
read as a typed projection of the same filtered-ownership object.  Given a
declared action and quotient structure, let $\mathsf{Dir}(v)$ be the typed
relation generator (or owner direction) that first makes the distinction at
node $v$ independently necessary; let $\mathsf{Dim}(v)$ be the rank or
cardinality of the independently retained quotient classes at that grade; and
let $\mathsf{Num}_\alpha(v)$ be the declared closure amplitude or admissible
composition count along $\mathsf{Dir}(v)$ under action law $\alpha$.  The grade
$\mathsf{Grade}(v)$ records the first threshold at which the owner is required.
Thus
\[
 v\longmapsto
 \bigl(\mathsf{Num}_\alpha(v),\mathsf{Dir}(v),
       \mathsf{Dim}(v),\mathsf{Grade}(v)\bigr)
\]
is a family of typed projections of one relation object, rather than four
independent coordinates.  A representation theorem is still required for
these projections to be well-defined (up to typed isomorphism).  In a linear,
modular, or otherwise lawful quotient sector they may reduce to ordinary
numbers, vectors, and ranks; in the general non-linear sector they remain
partial, parent-dependent, and possibly non-commutative.  The unification is
therefore structural and geometric, not the unqualified equality of number,
dimension, and direction.

\begin{proposition}[Finite threshold-address algebra]
\label{app:thresholdradixprop}
Assume a finite rooted threshold tree with canonical child orders, calibrated
thresholds, typed compatible concatenation, and declared carry, refinement,
transport, owner/epoch, cost, and source-denied replay witnesses. Let
$P_s$ truncate a leaf address and its complete protected record at level $s$.
Then:
\begin{enumerate}[leftmargin=1.4em]
\item encode/decode is a bijection between finite admissible leaves and their
canonical variable-radix addresses; refinement preserves the old prefix,
$P_s\circ\mathsf{Refine}_{s\to s+1}=P_s$, and a committed carry appends a
unique typed suffix;
\item for $0<\rho<1$, the first-differing-level distance
\[
 d_\rho(u,v)=\rho^{\operatorname{LCP}(u,v)}
\]
(with $d_\rho(u,u)=0$) is an ultrametric on the finite leaf set. A
subthreshold update preserves the declared prefix, while a committed
refinement preserves that prefix and adds the new relation address;
\item if the carry/concatenation witnesses are associative on their admitted
domain, the threshold records form a finite typed partial algebra and the
execution/replay trace is transported by induction over the tree depth. U/I/L
ensure that protected distinctions are retained, future-equivalent leaves are
not split without a declared reason, and the selected refinement is the least
lawful one under the grade/ledger order; and
\item if child order, threshold calibration, quotient legality, owner/epoch,
replay, or carry coherence is missing, the operation is partial and returns
$\mathsf{OPEN}$ rather than inventing a scalar radix, vector-space basis, or
compression guarantee.
\end{enumerate}
\end{proposition}

\begin{proof}
Canonical local orders give a unique path and hence the encode/decode
bijection. Truncation deletes only a suffix, so refinement commutes with the
prefix projection. The ultrametric inequality follows from the longest-common-
prefix characterization: the common prefix of $u$ and $w$ is at least the
shorter of the common prefixes of $(u,v)$ and $(v,w)$. Hysteresis and the
calibrated thresholds make a subthreshold update leave the prefix unchanged;
a committed carry changes only the first new suffix. Induction on depth and
then on finite trace length transports typed composition, owner/epoch fields,
ledger charges, and replay receipts. U rules out silent loss, I rules out
merging different future laws, and L selects the least witnessed refinement.
Without a stated premise, no one of these steps is defined, so the contract
returns $\mathsf{OPEN}$.
\end{proof}

\paragraph{Why this can be a new mathematical direction.}
The object couples a filtration, a variable-radix address, a typed carry
program, a future-law quotient, and a resource/replay ledger. It can therefore
serve as a common language for representation, relation, scale, state
refinement, and executable change. This is more general than a fixed
coordinate vector when the number of admissible children depends on the
parent and when refinement is caused by a future-relevant defect. It is not,
however, a proof that every mathematical object has this form, a guarantee of
compression, or a claim that matrices, probability kernels, p-adic or
ultrametric constructions are thereby superseded.

\paragraph{Two algebraic boundary checks.}
For ordinary threshold indicators $z_t(x)=\mathbf 1\{x\geq t\}$ on $[0,1]$,
\[
 \int_0^1z_t(x)z_t(y)\,dt=\min(x,y),\qquad
 xy=\int_0^1\!\int_0^1z_s(x)z_t(y)\,ds\,dt.
\]
Thus a shared-threshold overlap is a min-kernel, not the ordinary product;
two independent threshold coordinates are needed to recover the product. The
same digit set can also encode different algebras: the carry term in
$\mathbb Z/4\mathbb Z$ differs from componentwise addition in
$\mathbb Z/2\mathbb Z\times\mathbb Z/2\mathbb Z$. Cross-level carry is part of
the mathematical object, not an optional implementation detail.

\end{document}